# Multi-visual-inertial system: Analysis, calibration, and estimation

**Yulin Yang, Patrick Geneva and Guoquan Huang**

**Abstract**

*In this paper, we study state estimation of multi-visual-inertial systems (MVIS) and develop sensor fusion algorithms to optimally fuse an arbitrary number of asynchronous inertial measurement units (IMUs) or gyroscopes and global and/or rolling shutter cameras. We are especially interested in the full calibration of the associated visual-inertial sensors, including the IMU/camera intrinsics and the IMU-IMU/camera spatiotemporal extrinsics as well as the image readout time of rolling-shutter cameras (if used). To this end, we develop a new analytic combined IMU integration with inertial intrinsics—termed $ACI^3$—to pre-integrate IMU measurements, which is leveraged to fuse auxiliary IMUs and/or gyroscopes alongside a base IMU. We model the multi-inertial measurements to include all the necessary inertial intrinsic and IMU-IMU spatiotemporal extrinsic parameters, while leveraging IMU-IMU rigid-body constraints to eliminate the necessity of auxiliary inertial poses and thus reducing computational complexity. By performing observability analysis of MVIS, we prove that the standard four unobservable directions remain—no matter how many inertial sensors are used, and also identify, for the first time, degenerate motions for IMU-IMU spatiotemporal extrinsics and auxiliary inertial intrinsics. In addition to extensive simulations that validate our analysis and algorithms, we have built our own MVIS sensor rig and collected over 25 real-world datasets to experimentally verify the proposed calibration against the state-of-the-art calibration method Kalibr. We show that the proposed MVIS calibration is able to achieve competing accuracy with improved convergence and repeatability, which is open sourced to better benefit the community.*

## 1. Introduction

A camera image provides two degree-of-freedom (DoF) bearing observations to environmental features, while a 6-axis MEMS inertial measurement unit (IMU), which consists of a gyroscope (or gyrometer) and an accelerometer, can measure high-rate angular velocity and linear acceleration of a rigid sensor body. The fusion of cameras and IMUs has become prevalent in autonomous vehicles and mobile devices in the recent decade due to their decrease in cost and complementary sensing characteristics. This has led to a significant progress of developing visual-inertial navigation system (VINS) algorithms focusing on efficient and accurate pose estimation (Huang, 2019). While many works have shown accurate estimation for the minimal sensing case of a single camera and IMU (Bloesch et al., 2015; Forster et al., 2016; Geneva et al., 2020; Mourikis and Roumeliotis, 2007; Qin et al., 2018), it is known that the inclusion of additional sensors can provide improved accuracy due to additional information and robustness to single sensor failure cases (Eckenhoff et al., 2021; Paul et al., 2017). Recently, multi-visual-inertial systems (MVIS)—which uses multiple IMUs and multiple cameras for 6 DoF pose tracking and 3D mapping—have been deployed to micro aerial vehicles (MAVs), AR/VR devices, and autonomous vehicles, thus the need for accurate sensor calibration and state estimation algorithms continues to grow. We have previously investigated the *single* IMU-camera calibration case (Yang et al., 2023b), showing that even a small perturbation to calibration parameters may cause significant trajectory accuracy loss, which calls for accurate calibration of MVIS.

Regarding state estimation, many works have explored to use multiple vision sensors for better VINS performance (Campos et al., 2021; Fu et al., 2021; Kuo et al., 2020; Leutenegger et al., 2015; Paul et al., 2017; Sun et al., 2018;

Robot Perception and Navigation Group, University of Delaware, Newark, DE, USA

**Corresponding author:**
Yulin Yang, Department of Mechanical Engineering, University of Delaware, 130 Academy Street, Newark, DE 19716, USA.
Email: yuyang@udel.edu

Usenko et al., 2016). In particular, Leutenegger et al. (2015), Usenko et al. (2016), and Fu et al. (2021) have shown that stereo camera or multiple cameras can achieve better pose accuracy or lower the uncertainties of IMU-Camera calibration. Only a few works recently investigate multiple inertial sensor fusion for VINS (Eckenhoff et al., 2019b; Faizullin and Ferrer, 2023; Kim et al., 2017; Wu et al., 2023; Zhang et al., 2020), showing that the system robustness and pose accuracy can be improved by fusing additional IMUs. For optimal fusion of multiple asynchronous visual and inertial sensors for MVIS, it is crucial to provide accurate full-parameter calibration for these sensors, which include (i) IMU-IMU/camera rigid transformation, (ii) IMU-IMU/camera time offset, (iii) time-varying IMU biases, (iv) constant IMU intrinsics, including scale/skewness correction for gyroscope/accelerometer, g-sensitivity, and the rotation between the gyroscope and accelerometer, (v) camera projection/distortion model parameters, and (vi) image readout time of rolling shutter (RS) cameras.

While there exists literature regarding to multi-camera and multi-IMU navigation systems (Eckenhoff et al., 2021; Fu et al., 2021; Furgale et al., 2013; Geneva et al., 2020; Kim et al., 2017; Rehder et al., 2016; Zhi et al., 2022), most of these works do not support full-parameter calibration. For example, only synchronized multiple global shutter (GS) cameras are supported in Fu et al. (2021) and Geneva et al. (2020), which cannot handle measurements from multiple asynchronous inertial sensors or RS cameras. Only rigid transformations for multiple IMUs can be calibrated in the work of Kim et al. (2017). Although the work (Zhi et al., 2022) can handle the spatiotemporal calibration for asynchronous cameras and IMUs, rolling shutter (RS) calibration and IMU/camera intrinsic calibration are missing. The work (Eckenhoff et al., 2021) can calibrate multiple asynchronous RS cameras and IMUs but the IMU intrinsic parameters (including scale/skewness correction and g-sensitivity) were not estimated. Although the work (Furgale et al., 2013) and its extension (Rehder et al., 2016) support the spatiotemporal calibration for multiple IMUs and cameras with their intrinsics, they do not support hybrid calibration of GS/RS cameras, nor the joint optimization of IMU-IMU or IMU-Camera time offsets. To the best of our knowledge, no work can perform joint optimization of all these calibration parameters, which are critical for multi-sensor fusion.

In this paper, we thus aim to perform full-parameter joint calibration of MVIS, including IMU-IMU and IMU-Camera spatiotemporal calibration, IMU/camera intrinsics, and RS readout time. Note that IMU intrinsics include both time-varying biases and constant IMU intrinsics terms, such as scale/skewness correction for gyroscope/accelerometer, g-sensitivity, and the rotation between gyroscope and accelerometer. All the IMU intrinsics are jointly estimated in the proposed MVIS. Joint calibration is often necessary due to its removal of specialized IMU calibration fixtures, e.g., rate tables, since the aiding camera sensor is able to provide exteroceptive information concurrently. Additionally, key parameters such as IMU scale and camera focal lengths are sensitive to environmental humidity and temperature, which can cause unmodeled errors if sequential data collections are used. Many works have shown the benefits of concurrent estimation and calibration on trajectory and parameter accuracy. For example, Rehder et al. (2016) showed that estimating IMU intrinsics improves IMU-Camera extrinsic calibration, Fu et al. (2021) showed that joint calibration in multi-camera systems reduced parameter uncertainty. Li et al. (2014) and Huai et al. (2022) gained improvements in system performance (including reductions in reprojection errors) by performing concurrent full-parameter estimation. Moreover, no observability analysis focusing on MVIS is available in the literature. We specifically focus on a MVIS which contains multiple IMUs (IMU-IMU) or additional gyroscopes (IMU-Gyroscope), as the fusion of multiple low-cost noisy gyroscopes holds great potential to improve downstream orientation estimation (Eckenhoff et al., 2021; Zhang et al., 2020). In particular, the degenerate motion study of the spatiotemporal calibration for IMU-IMU is missing from the literature, which greatly limits our understanding of such systems.

To fill this gap, we first leverage our previous work on analytic combined IMU integration (Yang et al., 2020a) to derive a new IMU integrator for IMU intrinsic calibration (i.e., $ACI^3$). As compared to the conventional IMU preintegration (Forster et al., 2016; Lupton and Sukkarieh, 2012) and general pre-integration (Fourmy et al., 2021) with only mean correction from bias terms, we support inertial intrinsics calibration and have both the mean and covariance corrections when linearization points change for biases and constant IMU intrinsics. In addition, we also model cross correlations between IMU navigation state and bias, as done in Brossard et al. (2021) and Eckenhoff et al. (2019a), which are missing from Forster et al. (2016) and Fourmy et al. (2021). Additionally, $ACI^3$ analytically computes IMU intrinsics Jacobians which have not been seen in the literature. Based on this, we design a novel algorithm to fuse multiple IMU/gyroscope measurements by using the rigid body constraints between these inertial sensors. A complete MVIS algorithm is developed, which can truly jointly estimate all the calibration parameters (spatiotemporal parameters for IMU-IMU and IMU-Camera, IMU/camera intrinsics, and readout time) within a batch nonlinear least squares (NLS) optimization framework. Based on the linearized system models, the observability analysis of MVIS with full-calibration is performed. We show that all these calibration parameters are observable given fully excited motions, and also, for the first time, identify the degenerate motions for IMU-IMU and IMU-Gyroscope spatiotemporal calibration. By building our own MVIS sensor rig with multiple IMUs and GS/RS cameras for data collection, we validate the proposed system against the state-of-art Kalibr (Furgale et al., 2013; Rehder et al., 2016).

In particular, the main contributions of this work are the following:

- We propose an optimization-based multi-visual-inertial (IMU and/or gyroscope) sensor calibration algorithm, which jointly estimates *all* spatiotemporal (including RS readout time) and intrinsic parameters for an arbitrary number of visual and inertial sensors.
- Building upon our prior work (Yang et al., 2020a), we develop a new analytic combined IMU integrator with inertial intrinsics (i.e., $ACI^3$), which corrects both mean and covariance of pre-integrated IMU measurements when IMU bias and intrinsic linearization changes. We also propose an auxiliary IMU fusion algorithm that allows for both the extrinsic and intrinsic calibration for multiple IMU sensors.
- We perform comprehensive observability analysis for the MVIS with full-parameter calibration, and, for the first time, identify the degenerate motions related to IMU-IMU calibration. We show that under one-axis rotation motion, the rotation between IMU and gyroscope is unobservable along rotation axis. We also show that under constant local angular and linear velocity, the time offset between IMUs is observable, which counters our intuitions.
- We conduct extensive simulations with three typical motion profiles. The simulation results confirm that we are able to recover all visual and inertial calibration parameters with the proposed MVIS in the fully excited motion case. Specifically, 25 datasets collected by our self-made sensor rig are also used for evaluating the proposed MVIS against state-of-art calibration methods, and the results prove that the proposed approach achieves comparable accuracy and better repeatability. The identified degenerate motions for the pertinent calibration parameters are also verified through both simulation and real-world experiments.

The paper is organized as follows: after briefly reviewing the related works in the next section, we present the proposed MVIS estimation framework in Section 3. The visual-inertial measurement constraints used by the system are explained in Sections 4–7, while in Sections 8–9, we present the observability analysis and identified degenerate motions. We validate our analysis and algorithms in Sections 10–11 and conclude the paper in Section 12.

## 2. Related work

While there exists rich literature in VINS (Huang, 2019), in the following, we only review the works closely related to MVIS and the calibration of MVIS, which can be categorized as (i) multiple inertial sensors aided navigation systems, (ii) multiple cameras aided inertial navigation systems, and (iii) multi-camera and multi-IMU navigation systems.

### 2.1. Multi-inertial navigation

There are a few works using multiple inertial sensors to improve the navigation system. Wu et al. (2023) proposed to use triple IMUs with wheel encoders to improve dead reckoning and showed that the drifting rates continue dropping as the number of used IMU increases. Faizullin and Ferrer (2023) proposed to use the best axes composition (BAC) algorithm to select the best fitting data from multiple inertial sensors to avoid systematic errors when fusing multiple customer grade IMUs. They showed that the inertial navigation system performances can be improved by increasing the number of used inertial sensors. Kim et al. (2017) fused multiple IMUs through reformulating preintegration (Forster et al., 2016) by transforming the auxiliary inertial readings into the base inertial frame. However, they relied on the numerical computation of angular accelerations to perform this transformation. They did not estimate the IMU-IMU related calibration parameters, either.

Zhang et al. (2020) proposed to convert the readings from multiple IMUs into a single “virtual” synthetic IMU measurement, which is expected to be less noisy. While offering computational savings compared to other multi-inertial fusion algorithms, it relies on having perfectly known spatiotemporal calibration for these inertial sensors. It is clear that the above mentioned works all assume the high-accuracy IMU-IMU calibration is provided and they leverage multiple IMUs but with only one camera for pose estimation. Jadid et al. (2019) showed that the fusion of three low-cost calibrated IMUs can be used to achieve similar pose tracking performances as a single high-end IMU in the application of tracking head mounted device (HMD). They also proposed to use static IMU measurements to calibrate accelerometer intrinsics and non-static IMU measurements to calibrate the gyroscope intrinsics. Lee et al. (2022) proposed an extrinsic calibration algorithm for multiple IMUs when these IMUs are rigidly connected and moving arbitrarily. Only measurements from these IMUs are needed for the calibration. However, the time offsets between these IMUs and the IMU intrinsic calibration are all missing from this work.

Although all the above mentioned works have shown that fusion of multiple IMUs to wheel-INS or VINS can improve pose tracking accuracy, most of them rely entirely or partially on high-accuracy prior calibration of these IMUs: including IMU-IMU spatiotemporal calibration and the inertial intrinsics of these IMUs. Instead, this paper aims to solve the full-parameter calibration for multiple IMUs, including intrinsics, extrinsics, and time offsets, especially for the application of multiple inertial sensors in the VINS domain.

### 2.2. Multi-camera aided inertial navigation

There have been quite a few works investigating fusing observations from multiple cameras for visual-inertial navigation. Processing all the measurements from multiple cameras will significantly slow down the system. Hence, Kuo et al. (2020) proposed an information-based keyframe selection algorithm for efficient multi-camera fusion. Zhang et al. (2022) proposed an efficient feature selection and tracking algorithm to speed up the measurement processing for multiple cameras. These two works only fuse visual observations from multiple cameras but without considering the sensor calibration. Fu et al. (2021) proposed to use multiple synchronized cameras to improve IMU-Camera calibration. They proved that the extrinsic covariance bound will be smaller when more cameras are used. This indicates that the IMU-Camera calibration can converge faster with more confidences. Our previous work, OpenVINS (Geneva et al., 2020) supports synchronized multi-camera aided VINS with extrinsic, intrinsic, and temporal calibration between IMU and camera. However, rolling shutter cameras are not supported by either of the above works.

In this paper, the proposed MVIS support the extrinsic and intrinsic calibration for multiple asynchronous global shutter or rolling shutter cameras. We provide quantitative analysis for how the calibration estimates are improved when 1, 2, or 3 cameras are used simultaneously. In addition, the proposed MVIS supports simultaneous calibration of global shutter and rolling shutter cameras with image readout time.

### 2.3. Multi-camera and multi-IMU navigation

There are only a few works focusing on joint calibration for multiple cameras and multiple IMUs. Zhi et al. (2022) proposed MultiCal, which exploits continuous-time curves to represent pose states and supports the spatial and temporal calibration for multiple IMUs and cameras with planar targets. However, rolling shutter camera calibration and IMU/camera intrinsic calibration are not supported. Eckenhoff et al. (2021) proposed a filter-based framework for fusing multiple IMUs and multiple cameras by estimating each auxiliary IMU with full state (containing orientation, position, velocity, and biases), and enforced relative pose constraints between sensors at fixed rates. It also showed robustness to inertial sensor failures. This work does not take into account the inertial intrinsic parameters and only includes the IMU-IMU and IMU-Camera spatiotemporal calibration. Additionally, their multi-IMU constraints required an additional 6 DoF pose for each auxiliary IMU since each IMU is propagated independently forward. Rehder et al. (2016) extended the continuous-time Kalibr framework (Furgale et al., 2013) to calibrate the extrinsic and intrinsic parameters of auxiliary inertial sensors by formulating the angular velocity and linear accelerations as functions of the trajectory spline derivatives. However, when performing IMU-Camera and IMU-IMU calibration, the camera intrinsics and IMU-IMU time offset are fixed instead of jointly optimized with other parameters. It does not support visual-inertial rolling shutter calibration. In addition, no theoretical consistency analysis or 3 sigma plots for IMU intrinsic calibration are provided for validation.

Unlike the above mentioned works, the proposed MVIS supports multiple IMU/gyroscope calibrations with both global shutter and rolling shutter cameras. All related parameters, including IMU/camera intrinsics, spatial and temporal parameters between IMUs and cameras, can be calibrated. In addition, we also, for the first time, provide MVIS observability analysis, which shows that all these calibration parameters are observable given fully excited motions. We also investigate the degenerate motions that might cause certain calibration parameter to fail, especially for IMU-IMU and IMU-Gyroscope spatiotemporal calibrations.

## 3. Multi-visual-inertial system

In this section, we first present the full IMU model (containing scales, axis-misalignment, and g-sensitivity) and camera model (containing camera intrinsics, lens distortion, and RS readout time) in the MVIS. We then introduce the state vector containing all the calibration parameters for an arbitrary number of IMUs and cameras, followed by the nonlinear least-squares (NLS) formulation.

### 3.1. IMU model

Since an IMU can read angular velocity and linear acceleration, it is assumed to consist of two distinguishable frames (Yang et al., 2023b): gyroscope frame $\{w\}$ and accelerometer frame $\{a\}$. Similar to the IMU models in Li et al. (2014), Schneider et al. (2019), and Yang et al. (2023b), the raw angular velocity ${}^{w}\boldsymbol{\omega}_m$ from the gyroscope and linear acceleration ${}^{a}\mathbf{a}_m$ from accelerometer are written as:

$$ {}^{w}\boldsymbol{\omega}_m = \mathbf{T}_w {}^{w}_{I}\mathbf{R}^{I}\boldsymbol{\omega} + \mathbf{T}_g {}^{I}\mathbf{a} + \mathbf{b}_g + \mathbf{n}_g \tag{1} $$

$$ {}^{a}\mathbf{a}_m = \mathbf{T}_a {}^{a}_{I}\mathbf{R}^{I}\mathbf{a} + \mathbf{b}_a + \mathbf{n}_a \tag{2} $$

where $\mathbf{T}_w$ denotes the scale and axis misalignment for $\{w\}$, while $\mathbf{T}_a$ represents the scale and axis misalignment for $\{a\}$. $\mathbf{T}_g$ represents the g-sensitivity. ${}^{w}_{I}\mathbf{R}$ and ${}^{a}_{I}\mathbf{R}$ denote the rotation from base IMU $\{I\}$ to the gyroscope and acceleration frame, respectively. We use 6 parameters (indexed columnwise upper/lower triangular matrix) to describe the $\mathbf{T}_w$ and $\mathbf{T}_a$ while $\mathbf{T}_g$ remains a $3 \times 3$ full matrix. $\mathbf{b}_g$ and $\mathbf{b}_a$ are the gyroscope and accelerometer biases, which are both modeled as random walks. $\mathbf{n}_g$ and $\mathbf{n}_a$ are the zero-mean Gaussian noises contaminating the measurements. The local acceleration ${}^{I}\mathbf{a}$ is defined as:

$$ {}^{I}\mathbf{a} = {}^{I}\mathbf{a}_b - {}^{I}_{G}\mathbf{R}^{G}\mathbf{g} \tag{3} $$

where ${}^{G}\mathbf{g} = [0\ 0\ -9.81]^{\top}$; ${}^{I}\mathbf{a}_b$ refer to the local acceleration of the rigid sensor body, which reflects the motion of the sensor.

The corrected angular velocity ${}^{I}\boldsymbol{\omega}$ and linear acceleration ${}^{I}\mathbf{a}$ are thus defined as:

$$ {}^{I}\boldsymbol{\omega} = {}^{I}_{w}\mathbf{R}\mathbf{D}_w \left( {}^{w}\boldsymbol{\omega}_m - \mathbf{b}_g - \mathbf{n}_g - \mathbf{T}_g {}^{I}\mathbf{a} \right) \tag{4} $$

$$ {}^I\mathbf{a} = {}^I_a\mathbf{R}\mathbf{D}_a({}^a\mathbf{a}_m - \mathbf{b}_a - \mathbf{n}_a) \tag{5} $$

where $\mathbf{D}_w = \mathbf{T}_w^{-1}$ and $\mathbf{D}_a = \mathbf{T}_a^{-1}$. Instead of $\mathbf{T}_w$ and $\mathbf{T}_a$, $\mathbf{D}_w$ and $\mathbf{D}_a$ are estimated, since they can be used directly for computing the corrected angular velocity ${}^I\boldsymbol{\omega}$ and linear acceleration ${}^I\mathbf{a}$. In the following, we refer $\mathbf{D}_w$ as the gyroscope correction and $\mathbf{D}_a$ as accelerometer correction.

To simplify the transformation among these three frames: $\{I\}$, $\{a\}$, and $\{w\}$, the IMU frame $\{I\}$ is chosen to coincide with either $\{w\}$ or $\{a\}$ (see Figure 1).

In practice, we consider two models in this paper: `RPNG` model and `kalibr` model[1].

- `RPNG` model: IMU frame $\{I\}$ coincides with $\{w\}$ and ${}^I_w\mathbf{R} = \mathbf{I}_3$. $\mathbf{T}_g$, ${}^I_a\mathbf{R}$, $\mathbf{D}_w$, and $\mathbf{D}_a$ are estimated. $\mathbf{D}_*$ is in upper triangular matrix form as:

$$ \mathbf{D}_* = \begin{bmatrix} d_{*1} & d_{*2} & d_{*4} \\ 0 & d_{*3} & d_{*5} \\ 0 & 0 & d_{*6} \end{bmatrix} \tag{6} $$

- `Kalibr` model: IMU frame $\{I\}$ coincides with $\{a\}$ and ${}^I_a\mathbf{R} = \mathbf{I}_3$. $\mathbf{T}_g$, ${}^I_w\mathbf{R}$, $\mathbf{D}'_w$ and $\mathbf{D}'_a$ are estimated. $\mathbf{D}'_*$ is in lower triangular matrix form as:

$$ \mathbf{D}'_* = \begin{bmatrix} d_{*1} & 0 & 0 \\ d_{*2} & d_{*4} & 0 \\ d_{*3} & d_{*5} & d_{*6} \end{bmatrix} \tag{7} $$

where the subscript $*$ denotes either $w$ or $a$.

Since the translation between the gyroscope and the accelerometer ${}^a\mathbf{p}_w$ has been safely ignored in most VINS algorithms (Li et al., 2014; Schneider et al., 2019; Schubert et al., 2018; Yang et al., 2020b) (e.g., they assume ${}^a\mathbf{p}_w = \mathbf{0}$), in the remaining of our analysis the `RPNG` model will be used to be compatible with existing literatures. For most MEMS IMUs, this translation should be sufficiently small and in the case that they are not, the `Kalibr` model is sufficient to ensure proper modeling. The `Kalibr` model is used to compare with Kalibr (Furgale et al., 2013) and benefits from assuming the IMU frame to be aligned with the accelerometer.

## *3.2. Camera model*

Cameras follow a pinhole model as in Eckenhoff et al. (2021) and Geneva et al. (2020). A 3D point feature, ${}^G\mathbf{p}_f$, is captured by a camera with visual measurement function:

$$ \mathbf{z}_C = [u \quad v]^\top + \mathbf{n}_C \triangleq \mathbf{h}_d(\mathbf{z}_n, \mathbf{x}_{C_{in}}) + \mathbf{n}_C \tag{8} $$

where $\{u, v\}$ is the distorted image pixel coordinate; $\mathbf{z}_n = [u_n \; v_n]^\top$ represents the normalized image pixel; $\mathbf{n}_C$ denotes the measurement noise; $\mathbf{h}_d(\cdot)$ maps the normalized image pixel onto the image plane based on the lens distortion models and camera intrinsic parameters $\mathbf{x}_{C_{in}}$:

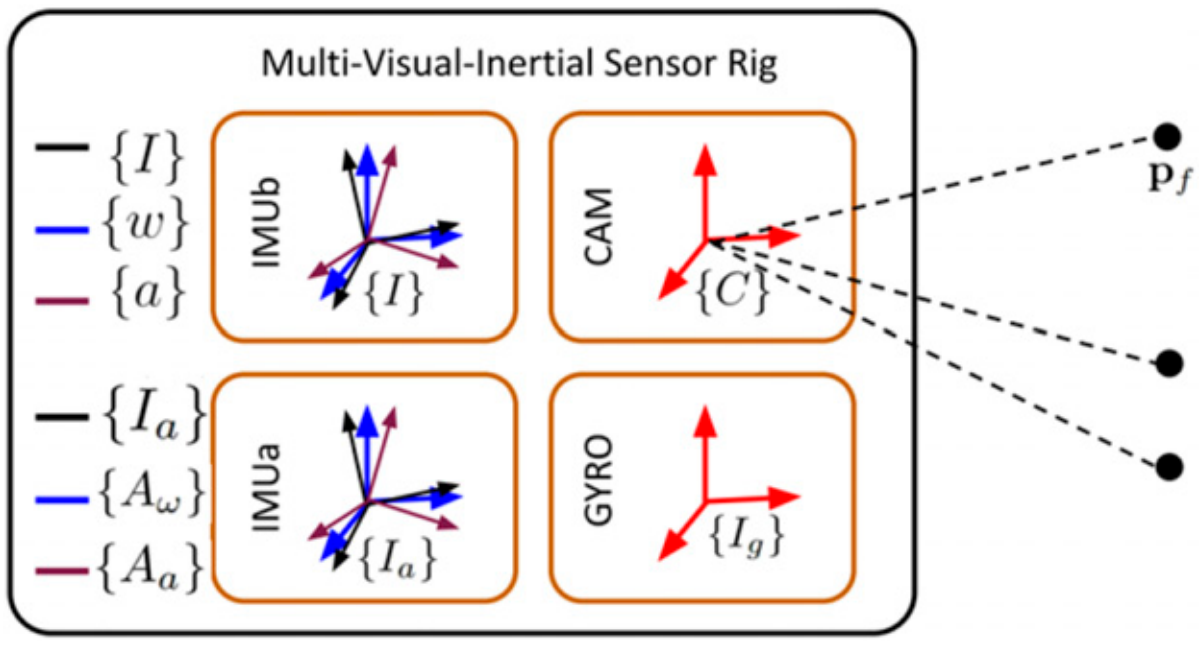

**Figure 1.** The multi-visual-inertial system contains: base IMU (`IMUb`) composed of an accelerometer $\{a\}$ and a gyroscope $\{w\}$, auxiliary IMUs (`IMUa`) $\{I_a\}$, auxiliary gyroscopes (`GYRO`) $\{I_g\}$, and cameras (`CAM`) $\{C\}$. Base IMU frame $\{I\}$ is determined to coincide with gyroscope frame $\{w\}$ or accelerometer frame $\{a\}$, $\{A_w\}$ and $\{A_a\}$ represent the gyroscope and accelerometer frames of the auxiliary IMU. While an IMU can read both angular velocities and linear acceleration, a gyroscope (`GYRO`) only reads angular velocities. The system observes environmental landmarks $\mathbf{p}_f$ through its cameras. The system can contain arbitrary amounts of sensors.

$$ \mathbf{x}_{C_{in}} = [f_u \; f_v \; c_u \; c_v \; k_1 \; k_2 \; p_1 \; p_2]^\top \tag{9} $$

Specifically, $\mathbf{x}_{C_{in}}$ can represent a pinhole model ($\{f_u, f_v\}$ denotes focal length and $\{c_u, c_v\}$ the center point) with radial-tangential (`radtan`) or equivalent-distant (`equidist`) distortion.

For `radtan` distortion model, $k_1$ and $k_2$ represent the radial distortion coefficients while $p_1$ and $p_2$ are tangential distortion coefficients. We refer the reader to Geneva et al. (2020) and OpenCV Developers Team (2021) for details on the `equidist` distortion model. Note that the `radtan` model is used in the following derivations and analysis. With $\mathbf{x}_{C_{in}}$, $\mathbf{h}_d(\cdot)$ with the `radtan` model is given by:

$$ \begin{bmatrix} u \\ v \end{bmatrix} = \begin{bmatrix} f_u & 0 \\ 0 & f_v \end{bmatrix} \begin{bmatrix} u_d \\ v_d \end{bmatrix} + \begin{bmatrix} c_u \\ c_v \end{bmatrix} \tag{10} $$

$$ \begin{bmatrix} u_d \\ v_d \end{bmatrix} = \begin{bmatrix} du_n + 2p_1u_nv_n + p_2(r^2 + 2u_n^2) \\ dv_n + p_1(r^2 + 2v_n^2) + 2p_2u_nv_n \end{bmatrix} \tag{11} $$

where $r^2 = u_n^2 + v_n^2$; $d = 1 + k_1r^2 + k_2r^4$;

GS cameras expose all pixels at a single time instance, while RS cameras expose each row sequentially. As shown by Guo et al. (2014), it may lead to large estimation errors if RS effects are not taken into account. Additionally, the camera and IMU measurement timestamps can be incorrect due to processing delays, or different clock references. To address this, we model both the time offset and camera readout time to ensure all measurements are processed in a common clock frame of reference and at the correct corresponding poses. Specifically, $t_d$ denotes the time offset between IMU and camera timeline while $t_r$ denotes the constant RS readout time for the whole image. If $t$ denotes the time when the pixel is captured, the RS measurement function for a normalized image pixel $\mathbf{z}_n$ is given by:

$$\mathbf{z}_n = \mathbf{h}_p\left({}^C\mathbf{p}_f\right) \triangleq \frac{1}{{}^Cz_f}\begin{bmatrix} {}^Cx_f \\ {}^Cy_f \end{bmatrix} \tag{12}$$

$$\begin{aligned} {}^C\mathbf{p}_f &= \mathbf{h}_t\left({}^{I(t)}_G\mathbf{R}, {}^G\mathbf{p}_{I(t)}, {}^C_I\mathbf{R}, {}^C\mathbf{p}_I, {}^G\mathbf{p}_f\right) \\ &\triangleq {}^C_I\mathbf{R}^{I(t)}_G\mathbf{R}\left({}^G\mathbf{p}_f - {}^G\mathbf{p}_{I(t)}\right) + {}^C\mathbf{p}_I \end{aligned} \tag{13}$$

where $\{{}^C_I\mathbf{R}, {}^C\mathbf{p}_I\}$ represents the rigid transformation between the IMU and camera frame. As usual, $\{{}^G_{I(t)}\mathbf{R}, {}^G\mathbf{p}_{I(t)}\}$ is the IMU global pose corresponding to the camera measurement time $t$.

If the image pixel $\mathbf{z}_C$, equation (8), is captured in the $m$-th row (out of total $M$ rows), and $t_I$ is the IMU state time corresponding to the captured image time $t_C$ when the first row of the image is collected, the relationship between $t_I$, $t_C$, $t_d$, and $t_r$ are expressed as:

$$t_C = t_I + t_d \tag{14}$$

$$t = t_I + \frac{m}{M}t_r \tag{15}$$

If the readout time $t_r = 0$, then the camera is actually a GS camera and all rows are a function of the same pose.

## 3.3. State vector

The proposed MVIS can support any number of IMUs and cameras. For simplicity of presentation, we only consider one representative sensor (one base IMU, one auxiliary IMU, one auxiliary gyroscope, and one RS camera) of each type in the state vector. Both simulation and real world experiments use multiple auxiliary IMUs/gyroscopes and cameras.

The state vector of MVIS contains the base IMU states $\mathcal{X}_I$, the auxiliary IMU states $\mathcal{X}_{I_a}$, and the auxiliary gyroscope states $\mathcal{X}_{I_g}$ from time stamp $t_1$ to $t_k$. Additionally, it also contains all the constant sensor intrinsics $\mathcal{X}_{In}$, spatiotemporal extrinsics $\mathcal{X}_{Ex}$, and environmental features $\mathcal{X}_F$.

$$\mathcal{X} = \begin{bmatrix} \mathcal{X}_I^\top & \mathcal{X}_{I_a}^\top & \mathcal{X}_{I_g}^\top & \mathcal{X}_{In}^\top & \mathcal{X}_{Ex}^\top & \mathcal{X}_F^\top \end{bmatrix}^\top \tag{16}$$

$$\mathcal{X}_I = \begin{bmatrix} \mathbf{x}_{I_1}^\top & \mathbf{x}_{I_2}^\top & \cdots & \mathbf{x}_{I_k}^\top \end{bmatrix}^\top \tag{17}$$

$$\mathcal{X}_{I_a} = \begin{bmatrix} \mathbf{x}_{I_{a_1}}{}^\top & \mathbf{x}_{I_{a_2}}{}^\top & \cdots & \mathbf{x}_{I_{a_k}}{}^\top \end{bmatrix}^\top \tag{18}$$

$$\mathcal{X}_{I_g} = \begin{bmatrix} \mathbf{x}_{I_{g_1}}{}^\top & \mathbf{x}_{I_{g_2}}{}^\top & \cdots & \mathbf{x}_{I_{g_k}}{}^\top \end{bmatrix}^\top \tag{19}$$

where $\mathcal{X}_F$ contains 3D feature positions ${}^G\mathbf{p}_{f_i}$ $(i = 1, \ldots, l)$:

$$\mathcal{X}_F = \begin{bmatrix} {}^G\mathbf{p}_{f_1}^\top \cdots {}^G\mathbf{p}_{f_l}^\top \end{bmatrix}^\top \tag{20}$$

The sub-states $\mathbf{x}_I$, $\mathbf{x}_{I_a}$, and $\mathbf{x}_{I_g}$ denote the base IMU state, auxiliary IMU state, and auxiliary gyroscope state, respectively. They contain the following:

$$\mathbf{x}_I = \begin{bmatrix} \mathbf{x}_{nav}^\top \mid \mathbf{x}_b^\top \end{bmatrix}^\top \triangleq \begin{bmatrix} {}^G_I\boldsymbol{\theta}^\top & {}^G\mathbf{p}_I^\top & {}^G\mathbf{v}_I^\top & \mid & \mathbf{b}_g^\top & \mathbf{b}_a^\top \end{bmatrix}^\top \tag{21}$$

$$\mathbf{x}_{I_a} = \begin{bmatrix} {}^G\mathbf{v}_{I_a}^\top \mid \mathbf{x}_{A_b}^\top \end{bmatrix}^\top \triangleq \begin{bmatrix} {}^G\mathbf{v}_{I_a}^\top & \mid & \mathbf{b}_{A_g}^\top & \mathbf{b}_{A_a}^\top \end{bmatrix}^\top \tag{22}$$

$$\mathbf{x}_{I_g} = \mathbf{b}_{G_g} \tag{23}$$

where $\mathbf{x}_I$, representing the time varying IMU state, contains IMU navigation state $\mathbf{x}_{nav}$ and bias state $\mathbf{x}_b$; ${}^G_I\boldsymbol{\theta}$ (${}^G_{I_a}\boldsymbol{\theta}$) is 3D angle-axis vector corresponding to the rotation ${}^G_I\mathbf{R}$ (${}^G_{I_a}\mathbf{R}$) from the base (auxiliary) IMU frame to global frame $\{G\}$. Note that ${}^G_I\boldsymbol{\theta} = \log({}^G_I\mathbf{R})$ with $\log(\cdot)$ defined as the log of $\mathcal{SO}(3)$ (Barfoot, 2017). ${}^G\mathbf{p}_I$ (${}^G\mathbf{p}_{I_a}$) and ${}^G\mathbf{v}_I$ (${}^G\mathbf{v}_{I_a}$) represent the global position and velocity of base (auxiliary) IMU in $\{G\}$. $\mathbf{b}_g$ and $\mathbf{b}_a$ denote the gyroscope and accelerometer bias for base IMU, respectively. $\mathbf{b}_{A_g}$ and $\mathbf{b}_{A_a}$ denote the gyroscope and accelerometer bias for auxiliary IMU, respectively. $\mathbf{b}_{G_g}$ denotes the gyroscope bias for auxiliary gyroscope. We did not keep a full navigation state for auxiliary IMU/gyroscope since the poses can be recovered by the rigid body transform from the base IMU. Note that IMU state $\mathbf{x}_I$ is created based on IMU frame corresponding to the camera image time. All the IMU readings between two image timestamps will be used for state estimation.

$\mathcal{X}_{In}$ contains constant base IMU intrinsics $\mathbf{x}_{in}$, constant auxiliary IMU intrinsics $\mathbf{x}_{A_{in}}$, constant auxiliary gyroscope intrinsics $\mathbf{x}_{G_{in}}$, and camera intrinsics $\mathbf{x}_{C_{in}}$:

$$\mathcal{X}_{In} = \begin{bmatrix} \mathbf{x}_{in}^\top & \mathbf{x}_{A_{in}}^\top & \mathbf{x}_{G_{in}}^\top & \mathbf{x}_{C_{in}}^\top \end{bmatrix}^\top \tag{24}$$

$$\mathbf{x}_{in} = \begin{bmatrix} \mathbf{x}_w^\top & \mathbf{x}_a^\top & \mathbf{x}_{T_g}^\top & {}^I_a\boldsymbol{\theta}^\top \end{bmatrix}^\top \tag{25}$$

$$\mathbf{x}_{A_{in}} = \begin{bmatrix} \mathbf{x}_{A_w}^\top & \mathbf{x}_{A_a}^\top & \mathbf{x}_{A_g}^\top & {}^{I_a}_{A_a}\boldsymbol{\theta}^\top \end{bmatrix}^\top \tag{26}$$

$$\mathbf{x}_{G_{in}} = \mathbf{x}_{G_w} \tag{27}$$

where $\mathbf{x}_w$ ($\mathbf{x}_{A_w}$), $\mathbf{x}_a$ ($\mathbf{x}_{A_a}$) contains all the 6 column-wise parameters from gyroscope correction $\mathbf{D}_w$ and accelerometer correction $\mathbf{D}_a$ for base (auxiliary) IMU, respectively. $\mathbf{x}_{T_g}$ ($\mathbf{x}_{A_g}$) contains all the 9 parameters for g-sensitivity $\mathbf{T}_g$ and $\mathbf{T}_{A_g}$ from base (auxiliary) IMU, respectively. ${}^I_a\boldsymbol{\theta}$ (${}^{I_a}_{A_a}\boldsymbol{\theta}$) denotes the rotation from the accelerometer frame to the base (auxiliary) IMU frame. $\mathbf{x}_{G_w}$ denotes the 6 column-wise parameters from gyroscope correction $\mathbf{D}_w$ for auxiliary gyroscope.

$\mathcal{X}_{Ex}$ contains the spatiotemporal calibrations for base IMU to auxiliary IMU $\mathbf{x}_{I_A}$, base IMU to auxiliary gyroscope $\mathbf{x}_{I_G}$, and base IMU to camera $\mathbf{x}_{I_C}$:

$$\mathcal{X}_{Ex} = \begin{bmatrix} \mathbf{x}_{I_A}^\top & \mathbf{x}_{I_G}^\top & \mathbf{x}_{I_C}^\top \end{bmatrix}^\top \tag{28}$$

$$\mathbf{x}_{I_A} = \begin{bmatrix} {}^I_{I_a}\boldsymbol{\theta}^\top & {}^I\mathbf{p}_{I_a}^\top & t_{d_a} \end{bmatrix}^\top \tag{29}$$

$$\mathbf{x}_{I_G} = \begin{bmatrix} {}^{I}_{I_g}\boldsymbol{\theta}^{\top} & t_{d_g} \end{bmatrix}^{\top} \tag{30}$$

$$\mathbf{x}_{I_C} = \begin{bmatrix} {}^{I}_{C}\boldsymbol{\theta}^{\top} & {}^{I}\mathbf{p}_C^{\top} & t_d & t_r \end{bmatrix}^{\top} \tag{31}$$

Note that time offset between auxiliary IMU and base IMU is defined as: $t_{d_a} = t_a - t_I$, where $t_a$ and $t_I$ represent the auxiliary and base IMU measurement time, respectively. The IMU-Gyroscope time offset $t_{d_g}$ and IMU-Camera time offset $t_d$ are defined in a similar way as $t_{d_a}$. $t_r$ denotes the whole image reading out time for the RS camera. Note that all the calibration parameters are summarized in Table 1.

### 3.4. NLS formulation

Given measurements $\mathbf{z}_S$ from a sensor $S$, with additive white Gaussian noise $\mathbf{n}_S$, we have:

$$\mathbf{z}_S = \mathbf{h}_S(\mathbf{x}) + \mathbf{n}_S, \mathbf{n}_S \sim \mathcal{N}(\mathbf{0}, \mathbf{R}_S) \tag{32}$$

where $\mathbf{h}_S(\cdot)$ denotes the nonlinear observation function. Then, we can formulate the NLS problem with state $\mathbf{x}$ as:

$$\min_{\mathbf{x}} \sum \left\| \mathbf{z}_S - \mathbf{h}_S(\mathbf{x}) \right\|^2_{\mathbf{R}_S} \tag{33}$$

An initial guess $\hat{\mathbf{x}}^{\ominus}$ is needed to start the optimization. After computing the incremental state correction $\delta\mathbf{x}$, we can refine the state estimates by $\hat{\mathbf{x}}^{\oplus} = \hat{\mathbf{x}}^{\ominus} \boxplus \delta\mathbf{x}$, where $\boxplus$ represents the state manifold update (Barfoot, 2017). In summary, we have the following NLS equivalent to maximum likelihood estimation (MLE) under some common assumptions:

$$\min_{\mathbf{x}} \sum \mathbb{C}_I + \sum \mathbb{C}_{I_a} + \sum \mathbb{C}_{I_g} + \sum \mathbb{C}_C \tag{34}$$

where $\mathbb{C}_I$, $\mathbb{C}_{I_a}$, $\mathbb{C}_{I_g}$, and $\mathbb{C}_C$ denotes the cost for base IMU, auxiliary IMUs, auxiliary gyroscopes, and cameras, respectively, and will be built explicitly later. The NLS from equation (34) can be solved through various nonlinear least squares solvers (e.g., IPOPT (Wächter and Biegler, 2006), g2o (Kümmerle et al., 2011), GTSAM (Dellaert, 2012), and Google Ceres (Agarwal et al., 2023)) and yields the optimal IMU states, point features, and calibration parameters [see eq. (16)]. In this paper, we choose the well-known GTSAM as our NLS solver for evaluations since it achieves comparable accuracy to other solvers (Jurić et al., 2021), but this should not stop interested readers from using other solvers.

## 4. ACI$^3$: Pre-integration with intrinsics

In this section, we extend our analytic combined IMU integration (ACI$^2$) (Yang et al., 2020a) to incorporate constant IMU intrinsics into pre-integration, and propose an efficient IMU integrator that can be leveraged for constant IMU intrinsic calibration.

The IMU dynamic model is given by Sola (2017) and Trawny and Roumeliotis (2005):

**Table 1.** The full-calibration parameters in the proposed MVIS.

| Sensor | Extrinsics | Temporal | Intrinsics | Qty |
|---|---|---|---|---|
| Base IMU | – | – | $\mathbf{x}_{in}$ | 1 |
| Aux IMU | ${}^{I}_{I_a}\mathbf{R}$, ${}^{I}\mathbf{p}_{I_a}$ | $t_{d_a}$ | $\mathbf{x}_{A_{in}}$ | ≥1 |
| Aux Gyro | ${}^{I}_{I_g}\mathbf{R}$ | $t_{d_g}$ | $\mathbf{x}_{G_{in}}$ | ≥1 |
| Camera | ${}^{I}_{C}\mathbf{R}$, ${}^{I}\mathbf{p}_C$ | $t_d$, $t_r$ | $\mathbf{x}_{C_{in}}$ | ≥1 |

$$\begin{aligned} {}^{G}_{I}\dot{\mathbf{R}} &= {}^{G}_{I}\mathbf{R} \cdot \lfloor {}^{I}\boldsymbol{\omega} \rfloor, \quad {}^{G}\dot{\mathbf{p}}_I = {}^{G}\mathbf{v}_I \\ {}^{G}\dot{\mathbf{v}}_I &= {}^{G}_{I}\mathbf{R}\,{}^{I}\mathbf{a} + {}^{G}\mathbf{g}, \\ {}^{I}\dot{\mathbf{b}}_g &= \mathbf{n}_{wg}, \quad {}^{I}\dot{\mathbf{b}}_a = \mathbf{n}_{wa} \end{aligned} \tag{35}$$

where ${}^{G}\mathbf{g}$ is defined in equation (3); $\mathbf{n}_{wg}$ and $\mathbf{n}_{wa}$ are the white Gaussian noises driving the gyroscope and accelerometer biases. ${}^{I}\boldsymbol{\omega} = [\omega_x \ \omega_y \ \omega_z]^{\top}$ and $\lfloor {}^{I}\boldsymbol{\omega} \rfloor = \begin{bmatrix} 0 & -\omega_z & \omega_y \\ \omega_z & 0 & -\omega_x \\ -\omega_y & \omega_x & 0 \end{bmatrix}$ represents a skew symmetric matrix (Trawny and Roumeliotis, 2005). We also denote $\hat{\mathbf{x}}$ as estimate of $\mathbf{x}$, while $\tilde{\mathbf{x}}$ as error states between $\mathbf{x}$ and $\hat{\mathbf{x}}$, i.e., $\tilde{\mathbf{x}} = \mathbf{x} \boxminus \hat{\mathbf{x}}$. The IMU pose is represented on $\mathcal{SO}(3) \times \mathbb{R}^3$ with error states:

$$\mathbf{R} = \hat{\mathbf{R}} \exp(\delta\boldsymbol{\theta}) \tag{36}$$

$$\mathbf{p} = \hat{\mathbf{p}} + \tilde{\mathbf{p}} \tag{37}$$

where $\exp(\delta\boldsymbol{\theta}) \simeq \mathbf{I}_3 + \lfloor \delta\boldsymbol{\theta} \rfloor$ for small angle $\delta\boldsymbol{\theta} \in \mathbb{R}^3$; $\exp(\cdot)$ denotes the exponential operation of $\mathcal{SO}$ (3) (Barfoot, 2017).

### 4.1. Pre-integration terms

Between two sampling times $t_k$ and $t_j$, we integrate the IMU dynamic model as follows:

$${}^{G}_{I_j}\mathbf{R} = {}^{G}_{I_k}\mathbf{R} \cdot \Delta\mathbf{R} \tag{38}$$

$${}^{G}\mathbf{p}_{I_j} = {}^{G}\mathbf{p}_{I_k} + {}^{G}\mathbf{v}_{I_k}\delta t + {}^{G}_{I_k}\mathbf{R}\Delta\mathbf{p} + \frac{1}{2}{}^{G}\mathbf{g}\delta t^2 \tag{39}$$

$${}^{G}\mathbf{v}_{I_j} = {}^{G}\mathbf{v}_{I_k} + {}^{G}_{I_k}\mathbf{R}\Delta\mathbf{v} + {}^{G}\mathbf{g}\delta t \tag{40}$$

$$\mathbf{b}_{g_j} = \mathbf{b}_{g_k} + \Delta\mathbf{b}_g \tag{41}$$

$$\mathbf{b}_{a_j} = \mathbf{b}_{a_k} + \Delta\mathbf{b}_a \tag{42}$$

where $\delta t = t_j - t_k$, while $\Delta\mathbf{R}$, $\Delta\mathbf{p}$, $\Delta\mathbf{v}$, $\Delta\mathbf{b}_g$, and $\Delta\mathbf{b}_a$, which are the IMU pre-integration terms from $t_k$ to $t_j$, are described by:

$$\Delta\mathbf{R} \triangleq \exp\left( \int_{t_k}^{t_j} {}^{I_\tau}\boldsymbol{\omega} d\tau \right) = \mathbf{h}_R\left(\mathbf{x}_{I_k}, \mathbf{x}_{I_j}\right) \tag{43}$$

$$\Delta\mathbf{p} \triangleq \int_{t_k}^{t_j} \int_{t_k}^{s} {}^{I_k}_{I_\tau}\mathbf{R}\,{}^{I_\tau}\mathbf{a} d\tau ds = \mathbf{h}_p\left(\mathbf{x}_{I_k}, \mathbf{x}_{I_j}\right) \tag{44}$$

$$\Delta \mathbf{v} \triangleq \int_{t_k}^{t_j} {}^{I_k}_{I_\tau}\mathbf{R}^{I_\tau}\mathbf{a} d\tau = \mathbf{h}_v(\mathbf{x}_{I_k}, \mathbf{x}_{I_j}) \tag{45}$$

$$\Delta \mathbf{b}_g \triangleq \int_{t_k}^{t_j} \mathbf{n}_{wg} d\tau = \mathbf{b}_{g_j} - \mathbf{b}_{g_k} \tag{46}$$

$$\Delta \mathbf{b}_a \triangleq \int_{t_k}^{t_j} \mathbf{n}_{wa} d\tau = \mathbf{b}_{a_j} - \mathbf{b}_{a_k} \tag{47}$$

where we have defined:

$$\mathbf{h}_R(\mathbf{x}_{I_k}, \mathbf{x}_{I_j}) \triangleq {}^{G}_{I_k}\mathbf{R}^\top {}^{G}_{I_j}\mathbf{R} \tag{48}$$

$$\mathbf{h}_p(\mathbf{x}_{I_k}, \mathbf{x}_{I_j}) \triangleq {}^{G}_{I_k}\mathbf{R}^\top \left( {}^{G}\mathbf{p}_{I_j} - {}^{G}\mathbf{p}_{I_k} - {}^{G}\mathbf{v}_{I_k}\delta t - \frac{1}{2}{}^{G}\mathbf{g}\delta t^2 \right) \tag{49}$$

$$\mathbf{h}_v(\mathbf{x}_{I_k}, \mathbf{x}_{I_j}) \triangleq {}^{G}_{I_k}\mathbf{R}^\top \left( {}^{G}\mathbf{v}_{I_j} - {}^{G}\mathbf{v}_{I_k} - {}^{G}\mathbf{g}\delta t \right) \tag{50}$$

In the following, the proposed ACI$^3$ recursively computes the mean and covariance of these pre-integrated terms (i.e., $\Delta\mathbf{R}$, $\Delta\mathbf{p}$, $\Delta\mathbf{v}$, $\Delta\mathbf{b}_g$, and $\Delta\mathbf{b}_a$) with intrinsics $\mathbf{x}_{in}$.

## 4.2. Recursive formulation

Assuming there are $j - k + 1$ IMU readings between the timestamps $k$ and $j$, there exits an integer $i$ such that $k \le k + i < k + i + 1 \le j$. $\Delta\mathbf{R}_i$, $\Delta\mathbf{p}_i$, $\Delta\mathbf{v}_i$, $\Delta\mathbf{b}_{gi}$, and $\Delta\mathbf{b}_{ai}$ denote the integration components using all IMU readings from time $t_k$ to $t_{k+i}$:

$$\Delta \mathbf{R}_i = \exp\left( \int_{t_k}^{t_{k+i}} {}^{I_\tau}\boldsymbol{\omega} d\tau \right) \tag{51}$$

$$\Delta \mathbf{p}_i = \int_{t_k}^{t_{k+i}} \int_{t_k}^{s} {}^{I_k}_{I_\tau}\mathbf{R}^{I_\tau}\mathbf{a} d\tau ds \tag{52}$$

$$\Delta \mathbf{v}_i = \int_{t_k}^{t_{k+i}} {}^{I_k}_{I_\tau}\mathbf{R}^{I_\tau}\mathbf{a} d\tau \tag{53}$$

$$\Delta \mathbf{b}_{gi} = \mathbf{b}_{g_{k+i}} - \mathbf{b}_{g_k} \tag{54}$$

$$\Delta \mathbf{b}_{ai} = \mathbf{b}_{a_{k+i}} - \mathbf{b}_{a_k} \tag{55}$$

With that, we compute the pre-integration in the following recursive form:

$$\Delta \mathbf{R}_{i+1} = \Delta \mathbf{R}_i \cdot \mathbf{R}_{i,i+1} \tag{56}$$

$$\Delta \mathbf{p}_{i+1} = \Delta \mathbf{p}_i + \Delta \mathbf{v}_i \delta t_i + \Delta \mathbf{R}_i \cdot \mathbf{p}_{i,i+1} \tag{57}$$

$$\Delta \mathbf{v}_{i+1} = \Delta \mathbf{v}_i + \Delta \mathbf{R}_i \cdot \mathbf{v}_{i,i+1} \tag{58}$$

$$\Delta \mathbf{b}_{gi+1} = \Delta \mathbf{b}_{gi} + \mathbf{b}_{gi,i+1} \tag{59}$$

$$\Delta \mathbf{b}_{ai+1} = \Delta \mathbf{b}_{ai} + \mathbf{b}_{ai,i+1} \tag{60}$$

where the increments are defined as:

$$\mathbf{R}_{i,i+1} \triangleq {}^{I_{k+i}}_{I_{k+i+1}}\mathbf{R} = \exp\left( \int_{t_{k+i}}^{t_{k+i+1}} {}^{I_\tau}\boldsymbol{\omega} d\tau \right) \tag{61}$$

$$\mathbf{p}_{i,i+1} \triangleq \int_{t_{k+i}}^{t_{k+i+1}} \int_{t_{k+i}}^{s} {}^{I_{k+i}}_{I_\tau}\mathbf{R}^{I_\tau}\mathbf{a} d\tau ds \tag{62}$$

$$\mathbf{v}_{i,i+1} \triangleq \int_{t_{k+i}}^{t_{k+i+1}} {}^{I_{k+i}}_{I_\tau}\mathbf{R}^{I_\tau}\mathbf{a} d\tau \tag{63}$$

$$\mathbf{b}_{gi,i+1} \triangleq \int_{t_{k+i}}^{t_{k+i+1}} \mathbf{n}_{wg} d\tau \tag{64}$$

$$\mathbf{b}_{ai,i+1} \triangleq \int_{t_{k+i}}^{t_{k+i+1}} \mathbf{n}_{wa} d\tau \tag{65}$$

By applying equations (56)–(60) to all the IMU readings from $t_k$ to $t_j$, we can compute both the mean and covariance of the IMU pre-integration terms (i.e., $\Delta\mathbf{R}$, $\Delta\mathbf{p}$, $\Delta\mathbf{v}$, $\Delta\mathbf{b}_g$, and $\Delta\mathbf{b}_a$), as shown in the following sections.

## 4.3. Mean prediction

To simplify the ensuring derivations, we rewrite the IMU readings $\boldsymbol{\omega}_{k+i}$ and $\mathbf{a}_{k+i}$ as:

$$\boldsymbol{\omega}_{k+i} = \widehat{\boldsymbol{\omega}}_{k+i} + \tilde{\boldsymbol{\omega}}_{k+i} \tag{66}$$

$$\mathbf{a}_{k+i} = \widehat{\mathbf{a}}_{k+i} + \tilde{\mathbf{a}}_{k+i} \tag{67}$$

where $\tilde{\boldsymbol{\omega}}_{k+i}$ and $\tilde{\mathbf{a}}_{k+i}$ are the error terms that contain IMU noises as defined in Appendix A. With bias terms defined in equations (54) and (55), $\widehat{\boldsymbol{\omega}}_{k+i}$ and $\widehat{\mathbf{a}}_{k+i}$ are computed as [see Eqs. (4) and (5)]:

$$\widehat{\boldsymbol{\omega}}_{k+i} = {}^{I}_{w}\widehat{\mathbf{R}}\widehat{\mathbf{D}}_w \left( {}^{w}\boldsymbol{\omega}_{m_{k+i}} - \Delta\widehat{\mathbf{b}}_{gi} - \widehat{\mathbf{b}}_{g_k} - \widehat{\mathbf{T}}_g \widehat{\mathbf{a}}_{k+i} \right) \tag{68}$$

$$\widehat{\mathbf{a}}_{k+i} = {}^{I}_{a}\widehat{\mathbf{R}}\widehat{\mathbf{D}}_a \left( {}^{a}\mathbf{a}_{m_{k+i}} - \Delta\widehat{\mathbf{b}}_{ai} - \widehat{\mathbf{b}}_{a_k} \right) \tag{69}$$

Assuming that $\widehat{\boldsymbol{\omega}}_{k+i}$ and $\widehat{\mathbf{a}}_{k+i}$ are constant during the IMU sampling interval $[t_{k+i}, t_{k+i+1}]$, we have:

$$\widehat{\mathbf{R}}_{i,i+1} = \exp(\widehat{\boldsymbol{\omega}}_{k+i}\delta t_i) \tag{70}$$

$$\widehat{\mathbf{p}}_{i,i+1} = \int_{t_{k+i}}^{t_{k+i+1}} \int_{t_{k+i}}^{s} {}^{I_{k+i}}_{I_\tau}\widehat{\mathbf{R}} d\tau ds \cdot \widehat{\mathbf{a}}_{k+i} \triangleq \boldsymbol{\Xi}_2 \cdot \widehat{\mathbf{a}}_{k+i} \tag{71}$$

$$\widehat{\mathbf{v}}_{i,i+1} = \int_{t_{k+i}}^{t_{k+i+1}} {}^{I_{k+i}}_{I_\tau}\widehat{\mathbf{R}} d\tau \cdot \widehat{\mathbf{a}}_{k+i} \triangleq \boldsymbol{\Xi}_1 \cdot \widehat{\mathbf{a}}_{k+i} \tag{72}$$

$$\widehat{\mathbf{b}}_{gi,i+1} = \mathbf{0}_{3\times 1}, \widehat{\mathbf{b}}_{ai,i+1} = \mathbf{0}_{3\times 1} \tag{73}$$

where $\boldsymbol{\Xi}_1$ and $\boldsymbol{\Xi}_2$ are defined below.

$$\boldsymbol{\Xi}_1 = \int_{t_{k+i}}^{t_{k+i+1}} {}^{I_{k+i}}_{I_\tau}\widehat{\mathbf{R}} d\tau \tag{74}$$

$$\boldsymbol{\Xi}_2 = \int_{t_{k+i}}^{t_{k+i+1}} \int_{t_{k+i}}^{s} {}^{I_{k+i}}_{I_\tau}\widehat{\mathbf{R}} d\tau ds \tag{75}$$

We thus recursively compute the IMU pre-integration mean:

$$\Delta\widehat{\mathbf{R}}_{i+1} = \Delta\widehat{\mathbf{R}}_i \cdot \widehat{\mathbf{R}}_{i,i+1} \tag{76}$$

$$\Delta\widehat{\mathbf{p}}_{i+1} = \Delta\widehat{\mathbf{p}}_i + \Delta\widehat{\mathbf{v}}_i \delta t_i + \Delta\widehat{\mathbf{R}}_i \cdot \widehat{\mathbf{p}}_{i,i+1} \tag{77}$$

$$\Delta\mathbf{v}_{i+1} = \Delta\widehat{\mathbf{v}}_i + \Delta\widehat{\mathbf{R}}_i \cdot \widehat{\mathbf{v}}_{i,i+1} \tag{78}$$

$$\Delta\widehat{\mathbf{b}}_{gi+1} = \Delta\widehat{\mathbf{b}}_{gi} + \widehat{\mathbf{b}}_{gi,i+1} \tag{79}$$

$$\Delta\widehat{\mathbf{b}}_{ai+1} = \Delta\widehat{\mathbf{b}}_{ai} + \widehat{\mathbf{b}}_{ai,i+1} \tag{80}$$

### *4.4. Covariance prediction*

To compute the covariance of the pre-integration measurements, we need to obtain the state transition matrix and noise Jacobians of the recursive formulation by linearizing the three pre-integration terms [see Eqs. (56)–(58)]:

$$\mathbf{R}_{i,i+1} = \widehat{\mathbf{R}}_{i,i+1}\tilde{\mathbf{R}}_{i,i+1} \tag{81}$$

$$= \widehat{\mathbf{R}}_{i,i+1} \exp(\mathbf{J}_r(\widehat{\boldsymbol{\theta}}_{i,i+1})\tilde{\boldsymbol{\omega}}_{k+i}\delta t_i) \tag{82}$$

$$\mathbf{p}_{i,i+1} = \widehat{\mathbf{p}}_{i,i+1} + \tilde{\mathbf{p}}_{i,i+1} \tag{83}$$

$$= \widehat{\mathbf{p}}_{i,i+1} - \boldsymbol{\Xi}_4\tilde{\boldsymbol{\omega}}_{k+i} + \boldsymbol{\Xi}_2\tilde{\mathbf{a}}_{k+i} \tag{84}$$

$$\mathbf{v}_{i,i+1} = \widehat{\mathbf{v}}_{i,i+1} + \tilde{\mathbf{v}}_{i,i+1} \tag{85}$$

$$= \widehat{\mathbf{v}}_{i,i+1} - \boldsymbol{\Xi}_3\tilde{\boldsymbol{\omega}}_{k+i} + \boldsymbol{\Xi}_1\tilde{\mathbf{a}}_{k+i} \tag{86}$$

where $\widehat{\boldsymbol{\theta}}_{i,i+1} = \widehat{\boldsymbol{\omega}}_{k+i}\delta t_i$ and $\mathbf{J}_r(\widehat{\boldsymbol{\theta}}_{i,i+1}) \triangleq \mathbf{J}_r(\widehat{\boldsymbol{\omega}}_{k+i}\delta t_i)$ denotes the right Jacobian of $\mathcal{SO}(3)$ (Chirikjian, 2011). The integrated components $\boldsymbol{\Xi}_3$ and $\boldsymbol{\Xi}_4$ are defined as:

$$\boldsymbol{\Xi}_3 \triangleq \int_{t_{k+i}}^{t_{k+i+1}} {}^{I_{k+i}}_{I_\tau}\mathbf{R}\lfloor\widehat{\mathbf{a}}_{k+i}\rfloor\mathbf{J}_r(\widehat{\boldsymbol{\omega}}_{k+i}\delta\tau)\delta\tau \, d\tau \tag{87}$$

$$\boldsymbol{\Xi}_4 \triangleq \int_{t_{k+i}}^{t_{k+i+1}} \int_{t_{k+i}}^{s} {}^{I_{k+i}}_{I_\tau}\mathbf{R}\lfloor\widehat{\mathbf{a}}_{k+i}\rfloor\mathbf{J}_r(\widehat{\boldsymbol{\omega}}_{k+i}\delta\tau)\delta\tau \, d\tau ds \tag{88}$$

Note that $\boldsymbol{\Xi}_i$, $i = \{1\ldots4\}$ can be evaluated analytically with detailed derivations in Yang et al. (2020b; 2023b) or numerically using Runge–Kutta fourth-order (RK4) method. The IMU pre-integration error states are given by:

$$\tilde{\mathbf{z}}_{I_i} = \begin{bmatrix} \delta\Delta\boldsymbol{\theta}_i^\top & \Delta\tilde{\mathbf{p}}_i^\top & \Delta\tilde{\mathbf{v}}_i^\top & \Delta\tilde{\mathbf{b}}_{gi}^\top & \Delta\tilde{\mathbf{b}}_{ai}^\top \end{bmatrix}^\top \tag{89}$$

with these errors states defined from equations (51)–(55):

$$\Delta\mathbf{R}_i = \Delta\widehat{\mathbf{R}}_i \exp(\delta\Delta\boldsymbol{\theta}_i) \tag{90}$$

$$\Delta\mathbf{p}_i = \Delta\widehat{\mathbf{p}}_i + \Delta\tilde{\mathbf{p}}_i \tag{91}$$

$$\Delta\mathbf{v}_i = \Delta\widehat{\mathbf{v}}_i + \Delta\tilde{\mathbf{v}}_i \tag{92}$$

$$\Delta\mathbf{b}_{gi} = \Delta\widehat{\mathbf{b}}_{gi} + \Delta\tilde{\mathbf{b}}_{gi} \tag{93}$$

$$\Delta\mathbf{b}_{ai} = \Delta\widehat{\mathbf{b}}_{ai} + \Delta\tilde{\mathbf{b}}_{ai} \tag{94}$$

Given that, the linearized IMU pre-integration model becomes:

$$\tilde{\mathbf{z}}_{I_{i+1}} = \boldsymbol{\Phi}_{i+1,i}\ \tilde{\mathbf{z}}_{I_i} + \boldsymbol{\Phi}_b\tilde{\mathbf{x}}_{b_k} + \boldsymbol{\Phi}_{in}\tilde{\mathbf{x}}_{in} + \mathbf{G}_i\mathbf{n}_{dI} \tag{95}$$

where $\boldsymbol{\Phi}_{i+1,i}$, $\boldsymbol{\Phi}_b$, $\boldsymbol{\Phi}_{in}$ and $\mathbf{G}_i$ are given in Appendix B. Finally, the measurement covariance $\mathbf{Q}_I$ follows the recursive form:

$$\mathbf{Q}_{I_{i+1}} = \boldsymbol{\Phi}_{i+1,i}\mathbf{Q}_{I_i}\boldsymbol{\Phi}_{i+1,i}^\top + \mathbf{G}_i\mathbf{Q}_d\mathbf{G}_i^\top \tag{96}$$

where $\mathbf{Q}_d$ denotes the discretized noises of ($\mathbf{n}_g$, $\mathbf{n}_a$, $\mathbf{n}_{wg}$, and $\mathbf{n}_{wa}$) from IMU readings. Through recursive evaluation of the above equation, we can recover the pre-integrated IMU measurement covariance between $t_k$ and $t_j$.

## 5. Base inertial costs

Since constant IMU intrinsics and biases are needed for IMU integration, $\Delta\mathbf{R}$, $\Delta\mathbf{p}$, and $\Delta\mathbf{v}$ are also functions of IMU biases $\mathbf{x}_{b_k}$ and constant IMU intrinsics $\mathbf{x}_{in}$. In order to avoid re-integration and re-linearization in iterative solvers when the IMU intrinsics and bias estimates are refined, the IMU pre-integration needs to fix the linearization points not only for IMU biases $\mathbf{x}_{b_k}$ as in Forster et al. (2016), *but* for constant IMU intrinsics $\mathbf{x}_{in}$.

To this end, we model the pre-integrated IMU measurements between time $t_k$ and $t_j$ as $\mathbf{z}_I \sim \mathcal{N}(\widehat{\mathbf{z}}_I, \mathbf{Q}_I)$:

$$\mathbf{z}_I = \begin{bmatrix} \log(\Delta\mathbf{R}) \\ \Delta\mathbf{p} \\ \Delta\mathbf{v} \\ \Delta\mathbf{x}_b \end{bmatrix} = \begin{bmatrix} \log(\Delta\mathbf{R}(\mathbf{x}_{b_k}, \mathbf{x}_{in}, \mathbf{n}_\theta)) \\ \Delta\mathbf{p}(\mathbf{x}_{b_k}, \mathbf{x}_{in}, \mathbf{n}_p) \\ \Delta\mathbf{v}(\mathbf{x}_{b_k}, \mathbf{x}_{in}, \mathbf{n}_v) \\ \mathbf{n}_b \end{bmatrix} \tag{97}$$

where the accumulated noises of IMU measurements are denoted with $\mathbf{n}_I = [\mathbf{n}_\theta^\top \ \mathbf{n}_p^\top \ \mathbf{n}_v^\top \ \mathbf{n}_b^\top]^\top$ and $\mathbf{n}_I \sim \mathcal{N}(\mathbf{0}, \mathbf{Q}_I)$ from Section 4.4. We linearize the above measurements at the current state estimate $\widehat{\mathbf{x}}$ as:

$$\Delta\mathbf{R} = \Delta\widehat{\mathbf{R}}\exp\left(\frac{\partial\delta\Delta\boldsymbol{\theta}}{\partial\tilde{\mathbf{x}}_{b_k}}\tilde{\mathbf{x}}_{b_k} + \frac{\partial\delta\Delta\boldsymbol{\theta}}{\partial\tilde{\mathbf{x}}_{in}}\tilde{\mathbf{x}}_{in} + \mathbf{n}_\theta\right) \tag{98}$$

$$\Delta\mathbf{p} = \Delta\widehat{\mathbf{p}} + \frac{\partial\Delta\tilde{\mathbf{p}}}{\partial\tilde{\mathbf{x}}_{b_k}}\tilde{\mathbf{x}}_{b_k} + \frac{\partial\Delta\tilde{\mathbf{p}}}{\partial\tilde{\mathbf{x}}_{in}}\tilde{\mathbf{x}}_{in} + \mathbf{n}_p \tag{99}$$

$$\Delta\mathbf{v} = \Delta\widehat{\mathbf{v}} + \frac{\partial\Delta\tilde{\mathbf{v}}}{\partial\tilde{\mathbf{x}}_{b_k}}\tilde{\mathbf{x}}_{b_k} + \frac{\partial\Delta\tilde{\mathbf{v}}}{\partial\tilde{\mathbf{x}}_{in}}\tilde{\mathbf{x}}_{in} + \mathbf{n}_v \tag{100}$$

Appendix C contains details of the Jacobian computations.

We now wish to fix the linearization points for the bias and IMU intrinsics states with their initial guesses to avoid costly re-integration during iterative solving. If we use $\widehat{\mathbf{x}}^{(0)}$ denote the initial estimates while $\tilde{\mathbf{x}}^{(0)}$ denote the corresponding initial error states, then we have:

$$\mathbf{x} = \hat{\mathbf{x}} + \tilde{\mathbf{x}} = \hat{\mathbf{x}}^{(0)} + \tilde{\mathbf{x}}^{(0)} \tag{101}$$

$$\Rightarrow \tilde{\mathbf{x}}^{(0)} = \hat{\mathbf{x}} - \hat{\mathbf{x}}^{(0)} + \tilde{\mathbf{x}} \triangleq \Delta\hat{\mathbf{x}} + \tilde{\mathbf{x}} \tag{102}$$

The IMU measurements can be linearized with the initial estimates $\hat{\mathbf{x}}^{(0)}$ as:

$$\begin{aligned}\Delta\mathbf{R} &= \Delta\hat{\mathbf{R}}^{(0)} \exp\left(\frac{\partial \delta\Delta\boldsymbol{\theta}}{\partial \tilde{\mathbf{x}}_{b_k}}\tilde{\mathbf{x}}_{b_k}^{(0)} + \frac{\partial \delta\Delta\boldsymbol{\theta}}{\partial \tilde{\mathbf{x}}_{in}}\tilde{\mathbf{x}}_{in}^{(0)} + \mathbf{n}_\theta\right) \\ &= \mathbf{h}_R\left(\mathbf{x}_{I_k}, \mathbf{x}_{I_j}\right)\end{aligned} \tag{103}$$

$$\begin{aligned}\Delta\mathbf{p} &= \Delta\hat{\mathbf{p}}^{(0)} + \frac{\partial \Delta\tilde{\mathbf{p}}}{\partial \tilde{\mathbf{x}}_{b_k}}\tilde{\mathbf{x}}_{b_k}^{(0)} + \frac{\partial \Delta\tilde{\mathbf{p}}}{\partial \tilde{\mathbf{x}}_{in}}\tilde{\mathbf{x}}_{in}^{(0)} + \mathbf{n}_p \\ &= \mathbf{h}_p\left(\mathbf{x}_{I_k}, \mathbf{x}_{I_j}\right)\end{aligned} \tag{104}$$

$$\begin{aligned}\Delta\mathbf{v} &= \Delta\hat{\mathbf{v}}^{(0)} + \frac{\partial \Delta\tilde{\mathbf{v}}}{\partial \tilde{\mathbf{x}}_{b_k}}\tilde{\mathbf{x}}_{b_k}^{(0)} + \frac{\partial \Delta\tilde{\mathbf{v}}}{\partial \tilde{\mathbf{x}}_{in}}\tilde{\mathbf{x}}_{in}^{(0)} + \mathbf{n}_v \\ &= \mathbf{h}_v\left(\mathbf{x}_{I_k}, \mathbf{x}_{I_j}\right)\end{aligned} \tag{105}$$

By applying equation (102) to equations (103)–(105), the pre-integrated IMU measurement with initial biases and initial IMU intrinsic estimates can be rewritten as:

$$\begin{aligned}\Delta\hat{\mathbf{R}}^{(0)} &= \mathbf{h}_R\left(\mathbf{x}_{I_k}, \mathbf{x}_{I_j}\right) \\ &\cdot \exp\left(-\frac{\partial \delta\Delta\boldsymbol{\theta}}{\partial \tilde{\mathbf{x}}_{b_k}}(\Delta\hat{\mathbf{x}}_{b_k} + \tilde{\mathbf{x}}_{b_k}) - \frac{\partial \delta\Delta\boldsymbol{\theta}}{\partial \tilde{\mathbf{x}}_{in}}(\Delta\hat{\mathbf{x}}_{in} + \tilde{\mathbf{x}}_{in}) - \mathbf{n}_\theta\right) \\ \Delta\hat{\mathbf{p}}^{(0)} &= \mathbf{h}_p\left(\mathbf{x}_{I_k}, \mathbf{x}_{I_j}\right) - \frac{\partial \Delta\tilde{\mathbf{p}}}{\partial \tilde{\mathbf{x}}_{b_k}}(\Delta\hat{\mathbf{x}}_{b_k} + \tilde{\mathbf{x}}_{b_k}) \\ &- \frac{\partial \Delta\tilde{\mathbf{p}}}{\partial \tilde{\mathbf{x}}_{in}}(\Delta\hat{\mathbf{x}}_{in} + \tilde{\mathbf{x}}_{in}) - \mathbf{n}_p \\ \Delta\hat{\mathbf{v}}^{(0)} &= \mathbf{h}_v\left(\mathbf{x}_{I_k}, \mathbf{x}_{I_j}\right) - \frac{\partial \Delta\tilde{\mathbf{v}}}{\partial \tilde{\mathbf{x}}_{b_k}}(\Delta\hat{\mathbf{x}}_{b_k} + \tilde{\mathbf{x}}_{b_k}) \\ &- \frac{\partial \Delta\tilde{\mathbf{v}}}{\partial \tilde{\mathbf{x}}_{in}}(\Delta\hat{\mathbf{x}}_{in} + \tilde{\mathbf{x}}_{in}) - \mathbf{n}_v\end{aligned}$$

We define $\boldsymbol{\theta}_{corr}$, $\mathbf{p}_{corr}$ and $\mathbf{v}_{corr}$ as the orientation, position and velocity correction terms due to the linearization point changes of IMU biases $\mathbf{x}_{b_k}$ and constant IMU intrinsics $\mathbf{x}_{in}$:

$$\boldsymbol{\theta}_{corr} \triangleq \frac{\partial \delta\Delta\boldsymbol{\theta}}{\partial \tilde{\mathbf{x}}_{b_k}}\Delta\hat{\mathbf{x}}_{b_k} + \frac{\partial \delta\Delta\boldsymbol{\theta}}{\partial \tilde{\mathbf{x}}_{in}}\Delta\hat{\mathbf{x}}_{in} \tag{106}$$

$$\mathbf{p}_{corr} \triangleq \frac{\partial \Delta\tilde{\mathbf{p}}}{\partial \tilde{\mathbf{x}}_{b_k}}\Delta\hat{\mathbf{x}}_{b_k} + \frac{\partial \Delta\tilde{\mathbf{p}}}{\partial \tilde{\mathbf{x}}_{in}}\Delta\hat{\mathbf{x}}_{in} \tag{107}$$

$$\mathbf{v}_{corr} \triangleq \frac{\partial \Delta\tilde{\mathbf{v}}}{\partial \tilde{\mathbf{x}}_{b_k}}\Delta\hat{\mathbf{x}}_{b_k} + \frac{\partial \Delta\tilde{\mathbf{v}}}{\partial \tilde{\mathbf{x}}_{in}}\Delta\hat{\mathbf{x}}_{in} \tag{108}$$

Finally, the base IMU pre-integration measurements are formulated in equation (109).The new IMU measurement noise $\mathbf{n}_I'$ is computed as:

$$\mathbf{n}_I' \triangleq \begin{bmatrix}\mathbf{n}_\theta' \\ \mathbf{n}_p \\ \mathbf{n}_v \\ \mathbf{n}_b\end{bmatrix} = \underbrace{\begin{bmatrix}\mathbf{J}_r(-\boldsymbol{\theta}_{corr}) & \mathbf{0}_{3\times 12} \\ \mathbf{0}_{12\times 3} & \mathbf{I}_{12}\end{bmatrix}}_{\mathbf{H}_{n'}} \begin{bmatrix}\mathbf{n}_\theta \\ \mathbf{n}_p \\ \mathbf{n}_v \\ \mathbf{n}_b\end{bmatrix} \triangleq \mathbf{H}_{\mathbf{n}'}\mathbf{n}_I \tag{110}$$

with covariance $\mathbf{n}_I' \sim \mathcal{N}(\mathbf{0}, \mathbf{Q}_I')$ and $\mathbf{Q}_I' = \mathbf{H}_{n'}\mathbf{Q}_I\mathbf{H}_{n'}^\top$. Finally, the corresponding base IMU pre-integration cost is:

$$\mathbb{C}_I \triangleq \left\|\mathbf{z}_I' \boxminus \mathbf{h}_I\left(\mathbf{x}_{I_k}, \mathbf{x}_{I_j}, \mathbf{x}_{in}\right)\right\|^2_{\mathbf{Q}_I'} \tag{111}$$

$$\underbrace{\begin{bmatrix}\log\left(\Delta\hat{\mathbf{R}}^{(0)}\right) \\ \Delta\hat{\mathbf{p}}^{(0)} \\ \Delta\hat{\mathbf{v}}^{(0)} \\ \mathbf{0}_{6\times 1}\end{bmatrix}}_{\mathbf{z}_I'} = \underbrace{\begin{bmatrix}\log\left(\mathbf{h}_R\left(\mathbf{x}_{I_k}, \mathbf{x}_{I_j}\right)\exp\left(-\frac{\partial \delta\Delta\boldsymbol{\theta}}{\partial \tilde{\mathbf{x}}_{b_k}}\tilde{\mathbf{x}}_{b_k} - \frac{\partial \delta\Delta\boldsymbol{\theta}}{\partial \tilde{\mathbf{x}}_{in}}\tilde{\mathbf{x}}_{in} - \boldsymbol{\theta}_{corr}\right)\exp\left(-\mathbf{n}_\theta'\right)\right) \\ \mathbf{h}_p\left(\mathbf{x}_{I_k}, \mathbf{x}_{I_j}\right) - \frac{\partial \Delta\tilde{\mathbf{p}}}{\partial \tilde{\mathbf{x}}_{b_k}}\tilde{\mathbf{x}}_{b_k} - \frac{\partial \Delta\tilde{\mathbf{p}}}{\partial \tilde{\mathbf{x}}_{in}}\tilde{\mathbf{x}}_{in} - \mathbf{p}_{corr} - \mathbf{n}_p \\ \mathbf{h}_v\left(\mathbf{x}_{I_k}, \mathbf{x}_{I_j}\right) - \frac{\partial \Delta\tilde{\mathbf{v}}}{\partial \tilde{\mathbf{x}}_{b_k}}\tilde{\mathbf{x}}_{b_k} - \frac{\partial \Delta\tilde{\mathbf{v}}}{\partial \tilde{\mathbf{x}}_{in}}\tilde{\mathbf{x}}_{in} - \mathbf{v}_{corr} - \mathbf{n}_v \\ \mathbf{x}_{b_j} - \mathbf{x}_{b_k} - \mathbf{n}_b\end{bmatrix}}_{\mathbf{h}_I\left(\mathbf{x}_{I_k}, \mathbf{x}_{I_j}, \mathbf{x}_{in}\right) \boxplus \mathbf{n}_I'} \tag{109}$$

As compared to other IMU pre-integration (Forster et al., 2016; Fourmy et al., 2021; Lupton and Sukkarieh, 2012) with only mean correction from bias terms, we support IMU intrinsics calibration and have both the mean correction [see Eqs. (106)–(108)] and covariance correction [see Eq. (110)] when linearization points change for time-varying biases and constant IMU intrinsics. In addition, the cross correlations between IMU navigation state and bias are modeled in $\mathbf{Q}_I'$, which are missing from (Forster et al., 2016; Fourmy et al., 2021).

## 6. Auxiliary inertial costs

Leveraging the base IMU pre-integration measurements [see Eq. (109)], we now show how to derive the costs for the auxiliary IMU and gyroscope by using the rigid body constraints between the base and auxiliary IMUs.

### 6.1. Auxiliary IMU cost

As the auxiliary IMUs are considered to be temporally asynchronous with the base IMU, we employ pose interpolation to associate the base IMU state with the auxiliary IMU state. The rigid body constraint between the auxiliary and base IMU with interpolation terms is given by:

$$\begin{bmatrix} {}^G_{I_a}\mathbf{R} & {}^G\mathbf{p}_{I_a} \\ \mathbf{0}_{1\times3} & 1 \end{bmatrix} \triangleq \begin{bmatrix} {}^G_{I_{in}}\mathbf{R} & {}^G\mathbf{p}_{I_{in}} \\ \mathbf{0}_{1\times3} & 1 \end{bmatrix} \begin{bmatrix} {}^I_{I_a}\mathbf{R} & {}^I\mathbf{p}_{I_a} \\ \mathbf{0}_{1\times3} & 1 \end{bmatrix} \tag{112}$$

where $\{{}^G_{I_{in}}\mathbf{R}, {}^G\mathbf{p}_{I_{in}}\}$ is the interpolated pose and represented in $\mathcal{SO}(3)\times\mathbb{R}^3$ space. It can be computed with constant linear velocity ${}^G\mathbf{v}_I$ and constant angular velocity ${}^I\boldsymbol{\omega}$ assumption:

$$\begin{bmatrix} {}^G_{I_{in}}\mathbf{R} & {}^G\mathbf{p}_{I_{in}} \\ \mathbf{0}_{1\times3} & 1 \end{bmatrix} = \begin{bmatrix} {}^G_{I}\mathbf{R} & {}^G\mathbf{p}_{I} \\ \mathbf{0}_{1\times3} & 1 \end{bmatrix} \begin{bmatrix} \exp\left({}^I\boldsymbol{\omega}t_{d_a}\right) & {}^G\mathbf{v}_I t_{d_a} \\ \mathbf{0}_{1\times3} & 1 \end{bmatrix}$$

The auxiliary IMU pose $\{{}^G_{I_a}\mathbf{R}, {}^G\mathbf{p}_{I_a}\}$ can be found with the base IMU pose $\{{}^G_I\mathbf{R}, {}^G\mathbf{p}_I\}$ as:

$${}^G_{I_a}\mathbf{R} = {}^G_I\mathbf{R}\exp\left({}^I\boldsymbol{\omega}t_{d_a}\right){}^I_{I_a}\mathbf{R} \tag{113}$$

$${}^G\mathbf{p}_{I_a} = {}^G_I\mathbf{R}\,{}^G\mathbf{v}_I t_{d_a} + {}^G\mathbf{p}_I + {}^G_I\mathbf{R}\exp\left({}^I\boldsymbol{\omega}t_{d_a}\right){}^I\mathbf{p}_{I_a} \tag{114}$$

Note that ${}^I\boldsymbol{\omega}$ denotes the angular velocity from the base IMU. Since ${}^I\boldsymbol{\omega}$ is not in the state vector, we need to use the current best estimate of the ${}^I\hat{\boldsymbol{\omega}}$.

There is no need to keep auxiliary IMU pose in the state vector, because the auxiliary IMU pose can be expressed by the base IMU state $\mathbf{x}_I$ and extrinsics $\mathbf{x}_{I_A}$. The auxiliary IMU state, equation (22), only contains the auxiliary IMU velocity and biases. We need to reformulate the pre-integration equations (48)–(50) for the auxiliary IMU cost with the equations (113) and (114) rigid body constraints. Following equation (109), with some abuse of the notations for the auxiliary IMU pre-integrated measurements $\mathbf{z}'_{I_a}$ and noises $\mathbf{n}'_{I_a}$, we can define auxiliary IMU residual as equation (115).

We have defined $\mathbf{h}_R(\cdot)$, $\mathbf{h}_p(\cdot)$, and $\mathbf{h}_v(\cdot)$ for the auxiliary IMU:

$$\begin{aligned}
\mathbf{h}_R(\,\cdot\,) &\triangleq {}^G_{I_{a_k}}\mathbf{R}^\top\,{}^G_{I_{a_j}}\mathbf{R} \\
&\triangleq \left({}^G_{I_k}\mathbf{R}\exp\left({}^{I_k}\boldsymbol{\omega}t_{d_a}\right){}^I_{I_a}\mathbf{R}\right)^\top {}^G_{I_j}\mathbf{R}\exp\left({}^{I_j}\boldsymbol{\omega}t_{d_a}\right){}^I_{I_a}\mathbf{R} \\
&\triangleq \mathbf{h}_R\left(\mathbf{x}_{I_k}, \mathbf{x}_{I_j}, \mathbf{x}_{I_A}\right) \\
\mathbf{h}_p(\,\cdot\,) &\triangleq {}^G_{I_{a_k}}\mathbf{R}^\top\left({}^G\mathbf{p}_{I_{a_j}} - {}^G\mathbf{p}_{I_{a_k}} - {}^G\mathbf{v}_{I_{a_k}}\delta t - \frac{1}{2}\,{}^G\mathbf{g}\delta t^2\right) \\
&\triangleq \mathbf{h}_p\left(\mathbf{x}_{I_k}, \mathbf{x}_{I_j}, \mathbf{x}_{I_{a_k}}, \mathbf{x}_{I_A}\right) \\
\mathbf{h}_v(\,\cdot\,) &\triangleq {}^G_{I_{a_k}}\mathbf{R}^\top\left({}^G\mathbf{v}_{I_{a_j}} - {}^G\mathbf{v}_{I_{a_k}} - {}^G\mathbf{g}\delta t\right) \\
&\triangleq \mathbf{h}_v\left(\mathbf{x}_{I_k}, \mathbf{x}_{I_{a_k}}, \mathbf{x}_{I_{a_j}}, \mathbf{x}_{I_A}\right)
\end{aligned}$$

Following equations (106)–(108), the linearization correction terms of orientation $\boldsymbol{\theta}_{A_{corr}}$, position $\mathbf{p}_{A_{corr}}$, and velocity $\mathbf{v}_{A_{corr}}$ for the auxiliary IMU are given by:

$$\boldsymbol{\theta}_{A_{corr}} \triangleq \frac{\partial\delta\Delta\boldsymbol{\theta}}{\partial\tilde{\mathbf{x}}_{A_{b_k}}}\Delta\widehat{\mathbf{x}}_{A_{b_k}} + \frac{\partial\delta\Delta\boldsymbol{\theta}}{\partial\tilde{\mathbf{x}}_{A_{in}}}\Delta\widehat{\mathbf{x}}_{A_{in}} \tag{116}$$

$$\mathbf{p}_{A_{corr}} \triangleq \frac{\partial\Delta\tilde{\mathbf{p}}}{\partial\tilde{\mathbf{x}}_{A_{b_k}}}\Delta\widehat{\mathbf{x}}_{A_{b_k}} + \frac{\partial\Delta\tilde{\mathbf{p}}}{\partial\tilde{\mathbf{x}}_{A_{in}}}\Delta\widehat{\mathbf{x}}_{A_{in}} \tag{117}$$

$$\mathbf{v}_{A_{corr}} \triangleq \frac{\partial\Delta\tilde{\mathbf{v}}}{\partial\tilde{\mathbf{x}}_{A_{b_k}}}\Delta\widehat{\mathbf{x}}_{A_{b_k}} + \frac{\partial\Delta\tilde{\mathbf{v}}}{\partial\tilde{\mathbf{x}}_{A_{in}}}\Delta\widehat{\mathbf{x}}_{A_{in}} \tag{118}$$

Finally, the corresponding auxiliary IMU cost is given by:

$$\mathbb{C}_{I_a} \triangleq \left\|\mathbf{z}'_{I_a} \boxminus \mathbf{h}_A(\mathbf{x})\right\|^2_{\mathbf{Q}'_{I_a}} \tag{119}$$

### 6.2. Auxiliary gyroscope cost

Similarly, the auxiliary gyroscope cost can be derived as the integration of angular velocity and gyroscope biases. The gyroscope state is defined as gyroscope biases, the intrinsics $\mathbf{x}_{G_w}$ and the extrinsics $\mathbf{x}_{I_G}$. The rotation constraint when considering a time offset is written as:

$$\underbrace{\begin{bmatrix} \log\left(\Delta\widehat{\mathbf{R}}^{(0)}\right) \\ \Delta\widehat{\mathbf{p}}^{(0)} \\ \Delta\widehat{\mathbf{v}}^{(0)} \\ \mathbf{0}_{6\times1} \end{bmatrix}}_{\mathbf{z}'_{I_a}} = \underbrace{\begin{bmatrix} \log\left(\mathbf{h}_R(\cdot)\exp\left(-\frac{\partial\delta\Delta\boldsymbol{\theta}}{\partial\tilde{\mathbf{x}}_{A_{b_k}}}\tilde{\mathbf{x}}_{A_{b_k}} - \frac{\partial\delta\Delta\boldsymbol{\theta}}{\partial\tilde{\mathbf{x}}_{A_{in}}}\tilde{\mathbf{x}}_{A_{in}} - \boldsymbol{\theta}_{A_{corr}}\right)\exp\left(-\mathbf{n}'_\theta\right)\right) \\ \mathbf{h}_p(\cdot) - \frac{\partial\Delta\tilde{\mathbf{p}}}{\partial\tilde{\mathbf{x}}_{A_{b_k}}}\tilde{\mathbf{x}}_{A_{b_k}} - \frac{\partial\Delta\tilde{\mathbf{p}}}{\partial\tilde{\mathbf{x}}_{A_{in}}}\tilde{\mathbf{x}}_{A_{in}} - \mathbf{p}_{A_{corr}} - \mathbf{n}_p \\ \mathbf{h}_v(\cdot) - \frac{\partial\Delta\tilde{\mathbf{v}}}{\partial\tilde{\mathbf{x}}_{A_{b_k}}}\tilde{\mathbf{x}}_{A_{b_k}} - \frac{\partial\Delta\tilde{\mathbf{v}}}{\partial\tilde{\mathbf{x}}_{A_{in}}}\tilde{\mathbf{x}}_{A_{in}} - \mathbf{v}_{A_{corr}} - \mathbf{n}_v \\ \mathbf{x}_{A_{b_j}} - \mathbf{x}_{A_{b_k}} - \mathbf{n}_b \end{bmatrix}}_{\mathbf{h}_A(\mathrm{x})\boxplus\mathbf{n}'_{I_a}} \tag{115}$$

$$ {}^{G}_{I_g}\mathbf{R} = {}^{G}_{I}\mathbf{R}\exp\left({}^{I}\boldsymbol{\omega} t_{d_g}\right){}^{I}_{I_g}\mathbf{R} \tag{120} $$

Reusing the notation $\mathbf{h}_R(\cdot)$, see equation (48), we get the gyroscope rotation function as:

$$ \begin{aligned} \mathbf{h}_R(\cdot) &\triangleq {}^{G}_{I_{gk}}\mathbf{R}^{\top}\,{}^{G}_{I_{gj}}\mathbf{R} \\ &\triangleq \left({}^{G}_{I_k}\mathbf{R}\exp\left({}^{I_k}\boldsymbol{\omega} t_{d_g}\right){}^{I}_{I_g}\mathbf{R}\right)^{\top}\,{}^{G}_{I_j}\mathbf{R}\exp\left({}^{I_j}\boldsymbol{\omega} t_{d_g}\right){}^{I}_{I_g}\mathbf{R} \\ &\triangleq \mathbf{h}_R\left(\mathbf{x}_{I_k}, \mathbf{x}_{I_j}, \mathbf{x}_{I_G}\right) \end{aligned} $$

where we still use the current best estimate for the ${}^{I}\boldsymbol{\omega}$. The pre-integrated auxiliary gyroscope measurements and noises are defined in equation (121). The linearization correction term is defined as:

$$ \underbrace{\begin{bmatrix} \log\left(\Delta\widehat{\mathbf{R}}^{(0)}\right) \\ \mathbf{0}_{3\times 1} \end{bmatrix}}_{\mathbf{z}'_{I_g}} = \underbrace{\begin{bmatrix} \log\left(\mathbf{h}_R(\cdot)\exp\left(-\frac{\partial\delta\Delta\boldsymbol{\theta}}{\partial\tilde{\mathbf{b}}_{G_{gk}}}\tilde{\mathbf{b}}_{G_{gk}} - \frac{\partial\delta\Delta\boldsymbol{\theta}}{\partial\tilde{\mathbf{x}}_{G_{in}}}\tilde{\mathbf{x}}_{G_{in}} - \boldsymbol{\theta}_{G_{corr}}\right)\exp\left(-\mathbf{n}'_{\theta}\right)\right) \\ \mathbf{b}_{G_{gj}} - \mathbf{b}_{G_{gk}} - \mathbf{n}_g \end{bmatrix}}_{\mathbf{h}_a(\mathbf{x})\boxplus\mathbf{n}'_{I_g}} \tag{121} $$

$$ \boldsymbol{\theta}_{G_{corr}} = \frac{\partial\delta\Delta\boldsymbol{\theta}}{\partial\tilde{\mathbf{b}}_{G_{gk}}}\Delta\widehat{\mathbf{b}}_{G_{gk}} + \frac{\partial\delta\Delta\boldsymbol{\theta}}{\partial\tilde{\mathbf{x}}_{G_{in}}}\Delta\widehat{\mathbf{x}}_{G_{in}} \tag{122} $$

Finally, the auxiliary gyroscope cost is given by:

$$ \mathbb{C}_{I_g} \triangleq \left\|\mathbf{z}'_{I_g} \boxminus \mathbf{h}_a(\mathbf{x})\right\|^2_{\mathbf{Q}'_{I_g}} \tag{123} $$

## 6.3. *Auxiliary inertial sensor initialization*

An initial linearization point of the auxiliary inertial states is required to perform optimization. This can be done by leveraging the initial linearization point of the base IMU. Specifically, we initialize the auxiliary IMU state $\mathbf{x}_{I_a}$, which contains the velocity ${}^{G}\mathbf{v}_{I_a}$ and biases ($\mathbf{b}_{A_g}$ and $\mathbf{b}_{A_a}$) of the auxiliary IMU, as follows:

- The initial velocity of the auxiliary IMU is computed based on the rigid body constraints:

$$ {}^{G}\mathbf{v}_{I_a} = {}^{G}\mathbf{v}_{I} + {}^{G}_{I}\mathbf{R}\left\lfloor {}^{I}_{I_a}\mathbf{R}\,{}^{I_a}\boldsymbol{\omega}\right\rfloor {}^{I}\mathbf{p}_{I_a} \tag{124} $$

- For $\widehat{\mathbf{b}}_{A_{gk}}$, we integrate the angular velocity measurements with zero bias for auxiliary IMU to get $\Delta\mathbf{R}_a$. Then the following linear system can be solved:

$$ \frac{\partial\delta\Delta\boldsymbol{\theta}}{\partial\tilde{\mathbf{b}}_{A_{gk}}}\cdot\widehat{\mathbf{b}}_{A_{gk}} = \log\left(\Delta\mathbf{R}_a^{\top}\,{}^{G}_{I_{a_k}}\mathbf{R}^{\top}\,{}^{G}_{I_{a_{k+1}}}\mathbf{R}\right) \tag{125} $$

where ${}^{G}_{I_{a_k}}\mathbf{R}$ and ${}^{G}_{I_{a_{k+1}}}\mathbf{R}$ denote the orientation of the auxiliary IMU which can be computed from base IMU orientation with IMU-IMU extrinsics.

- For $\widehat{\mathbf{b}}_{A_{a_k}}$, the accelerometer bias is initialized to $\mathbf{0}_{3\times 1}$.

## 7. Visual costs

We build the complete camera measurement function $\mathbf{h}_C(\cdot)$ by incorporating the distortion function $\mathbf{h}_d(\cdot)$ [see Eq. (8)], the projection function $\mathbf{h}_p(\cdot)$ [see Eq. (12)] and the transformation function $\mathbf{h}_t(\cdot)$ [see Eq. (13)] (Eckenhoff et al., 2021; Geneva et al., 2020):

$$ \begin{aligned} \mathbf{z}_C &= \mathbf{h}_C(\mathbf{x}) + \mathbf{n}_C & (126) \\ &= \mathbf{h}_d(\mathbf{z}_n, \mathbf{x}_{C_{in}}) + \mathbf{n}_C & (127) \\ &= \mathbf{h}_d\left(\mathbf{h}_p\left({}^{C}\mathbf{p}_f\right), \mathbf{x}_{C_{in}}\right) + \mathbf{n}_C & (128) \\ &= \mathbf{h}_d\left(\mathbf{h}_p\left(\mathbf{h}_t\left({}^{C}_{G}\mathbf{R}, {}^{G}\mathbf{p}_C, {}^{G}\mathbf{p}_f\right)\right), \mathbf{x}_{C_{in}}\right) + \mathbf{n}_C & (129) \end{aligned} $$

We need to linearize the camera model for update, which is given by:

$$ \tilde{\mathbf{z}}_C \simeq \mathbf{H}_C\tilde{\mathbf{x}} + \mathbf{n}_C \tag{130} $$

where $\tilde{\mathbf{z}}_C \triangleq \mathbf{z}_C - \mathbf{h}_C(\widehat{\mathbf{x}})$ and $\mathbf{H}_C \triangleq \partial\tilde{\mathbf{z}}_C/\partial\tilde{\mathbf{x}}$. Using the chain rule, we get the following Jacobian matrix:

$$ \begin{aligned} \mathbf{H}_C &= \begin{bmatrix} \frac{\partial\tilde{\mathbf{z}}_C}{\partial\tilde{\mathbf{x}}_I} & \frac{\partial\tilde{\mathbf{z}}_C}{\partial\tilde{\mathbf{x}}_{I_C}} & \frac{\partial\tilde{\mathbf{z}}_C}{\partial\tilde{\mathbf{x}}_{C_{in}}} & \frac{\partial\tilde{\mathbf{z}}_C}{\partial\tilde{\mathbf{x}}_f} \end{bmatrix} \\ &= \begin{bmatrix} \mathbf{H}_{\mathbf{p}_f}\frac{\partial{}^{C}\tilde{\mathbf{p}}_f}{\partial\tilde{\mathbf{x}}_I} & \mathbf{H}_{\mathbf{p}_f}\frac{\partial{}^{C}\tilde{\mathbf{p}}_f}{\partial\tilde{\mathbf{x}}_{I_C}} & \frac{\partial\tilde{\mathbf{z}}_C}{\partial\tilde{\mathbf{x}}_{C_{in}}} & \mathbf{H}_{\mathbf{p}_f}\frac{\partial{}^{C}\tilde{\mathbf{p}}_f}{\partial\tilde{\mathbf{x}}_f} \end{bmatrix} \end{aligned} \tag{131} $$

where $\mathbf{H}_{\mathbf{p}_f} = (\partial\tilde{\mathbf{z}}_C/\partial\tilde{\mathbf{z}}_n)\cdot(\partial\tilde{\mathbf{z}}_n/\partial{}^{C}\tilde{\mathbf{p}}_f)$. We refer the reader to the technical report for how to compute $\partial{}^{C}\tilde{\mathbf{p}}_f/\partial\tilde{\mathbf{x}}_I$, $\partial{}^{C}\tilde{\mathbf{p}}_f/\partial\tilde{\mathbf{x}}_{I_C}$, $\partial{}^{C}\tilde{\mathbf{p}}_f/\partial\tilde{\mathbf{x}}_f$, and $\mathbf{H}_{\mathbf{p}_f}$ (Yang et al., 2023a). Hence, the visual cost can be formulated:

$$ \mathbb{C}_C \triangleq \left\|\mathbf{z}_C - \mathbf{h}_C(\mathbf{x})\right\|^2_{\mathbf{Q}_C^{-1}} \tag{132} $$

Pose interpolation, which has been verified for accurate temporal calibration (Eckenhoff et al., 2021; Guo et al.,

2014; Lee et al., 2020, 2021), is leveraged to model the time offset and RS calibration in this work. Note that the pose interpolation follows manifold in $\mathcal{SO}(3) \times \mathbb{R}^3$ space. For example, if the feature measurement is in the $m$-th row with total $M$ rows in an image, we can find two bounding poses $k$ and $k+1$ based on the measurement time $t$. The corresponding time $t$ is between two IMU poses, $t_k \leq t \leq t_{k+1}$. We can then find the *virtual* IMU pose $\{{}^{G}_{I(t)}\mathbf{R}, {}^{G}\mathbf{p}_{I(t)}\}$ between poses at $k$ and $k+1$:

$$\lambda = \left(t_I + \frac{m}{M}t_r - t_k\right) \Big/ (t_{k+1} - t_k) \tag{133}$$

$${}^{G}_{I(t)}\mathbf{R} = {}^{G}_{I_k}\mathbf{R} \exp\left(\lambda \log\left({}^{G}_{I_k}\mathbf{R}^{\top}\, {}^{G}_{I_{k+1}}\mathbf{R}\right)\right) \tag{134}$$

$${}^{G}\mathbf{p}_{I(t)} = (1-\lambda)\,{}^{G}\mathbf{p}_{I_k} + \lambda\,{}^{G}\mathbf{p}_{I_{k+1}} \tag{135}$$

## 8. Observability analysis

Observability analysis plays an important role in state estimation for VINS (Huang, 2012; Martinelli, 2013). This analysis allows for determining the minimum measurements needed to determine the state and identify degenerate motions which may degrade system performance by introducing additional unobservable directions for certain parameters (Lee et al., 2020; Wu et al., 2017; Yang et al., 2019, 2020b). As MVIS continues to gain popularity, the observability analysis for such a system with full calibration parameters, especially IMU-IMU spatiotemporal calibration, is needed to better understand the fundamental properties of the underlying system.

### 8.1. Reduced state vector

Although the proposed MVIS supports arbitrary number of auxiliary inertial sensors and cameras, for simplicity and without loss of generality, we use a typical system consisting of only one base IMU, one auxiliary IMU, one auxiliary gyroscope, and one RS camera as unique sensors for the following observability analysis (Hesch et al., 2014; Yang et al., 2023b; Yang and Huang, 2019).

To simplify the ensuing derivation, we re-order the state vector and assume that the base IMU, auxiliary inertial sensors are all kept as full states (i.e., including the full auxiliary inertial state). All the states will be propagated forward with time, while the rigid body constraints and visual measurements will be used to update these states. Specifically, the state vector includes all the necessary parameters for the observability analysis as:

$$\mathbf{x} = \begin{bmatrix} \mathbf{x}_B^{\top} & \mathbf{x}_A^{\top} & \mathbf{x}_G^{\top} & \mathbf{x}_{calib}^{\top} & {}^{G}\mathbf{p}_f^{\top} \end{bmatrix}^{\top} \tag{136}$$

$$\triangleq \begin{bmatrix} \mathbf{x}_I^{\top} & \mathbf{x}_{in}^{\top} & \mathbf{x}_{I_a}^{\top} & \mathbf{x}_{A_{in}}^{\top} & \mathbf{x}_{I_g}^{\top} & \mathbf{x}_{G_{in}}^{\top} & \mathbf{x}_{Ex}^{\top} & \mathbf{x}_{C_{in}}^{\top} & {}^{G}\mathbf{p}_f^{\top} \end{bmatrix}^{\top} \tag{137}$$

Note that the auxiliary IMU and gyroscope states are:

$$\mathbf{x}_{I_a} = \begin{bmatrix} {}^{G}_{I_a}\boldsymbol{\theta}^{\top} & {}^{G}\mathbf{p}_{I_a}^{\top} & {}^{G}\mathbf{v}_{I_a}^{\top} & \mathbf{b}_{A_g}^{\top} & \mathbf{b}_{A_a}^{\top} \end{bmatrix}^{\top} \tag{138}$$

$$\mathbf{x}_{I_g} = \begin{bmatrix} {}^{G}_{I_g}\boldsymbol{\theta}^{\top} & \mathbf{b}_{G_g}^{\top} \end{bmatrix}^{\top} \tag{139}$$

After propagation, the visual measurements and rigid body constraints between inertial sensors are used to update the states with:

$$\mathbf{z} = \begin{bmatrix} \mathbf{z}_C^{\top} & \mathbf{z}_A^{\top} & \mathbf{z}_G^{\top} \end{bmatrix}^{\top} \tag{140}$$

where $\mathbf{z}_C$ denotes the visual cost [see Eq. (126)]. For simplicity, $\mathbf{z}_A$ and $\mathbf{z}_G$ represent the rigid body pose constraints between auxiliary and base inertial sensors:

$$\mathbf{z}_A = \begin{bmatrix} \log\left({}^{G}_{I_a}\mathbf{R}^{\top}\, {}^{G}_{I}\mathbf{R}\, {}^{I}_{I_a}\mathbf{R}\right) \\ {}^{G}\mathbf{p}_{I_a} - {}^{G}\mathbf{p}_I - {}^{G}_{I}\mathbf{R}\, {}^{I}\mathbf{p}_{I_a} \end{bmatrix} \tag{141}$$

$$\mathbf{z}_G = \log\left({}^{G}_{I_g}\mathbf{R}^{\top}\, {}^{G}_{I}\mathbf{R}\, {}^{I}_{I_g}\mathbf{R}\right) \tag{142}$$

### 8.2. Linearized observability analysis

The overall state transition matrix can be written as:

$$\boldsymbol{\Phi} = \mathrm{Diag}\{\boldsymbol{\Phi}_B, \boldsymbol{\Phi}_A, \boldsymbol{\Phi}_G, \boldsymbol{\Phi}_{calib}, \boldsymbol{\Phi}_F\} \tag{143}$$

The detailed derivations for $\boldsymbol{\Phi}_B$, $\boldsymbol{\Phi}_A$, $\boldsymbol{\Phi}_G$, $\boldsymbol{\Phi}_{calib}$, and $\boldsymbol{\Phi}_F$ can be found in Appendix D. The corresponding linearization Jacobians for equation (140) are:

$$\frac{\partial \tilde{\mathbf{z}}}{\partial \tilde{\mathbf{x}}} = \begin{bmatrix} \frac{\partial \tilde{\mathbf{z}}_C}{\partial \tilde{\mathbf{x}}} \\ \frac{\partial \tilde{\mathbf{z}}_A}{\partial \tilde{\mathbf{x}}} \\ \frac{\partial \tilde{\mathbf{z}}_G}{\partial \tilde{\mathbf{x}}} \end{bmatrix} = \begin{bmatrix} \mathbf{H}_{CB} & \mathbf{0} & \mathbf{0} & \mathbf{H}_{CC} & \mathbf{H}_{CF} \\ \mathbf{H}_{AB} & \mathbf{H}_{AA} & \mathbf{0} & \mathbf{H}_{AC} & \mathbf{0} \\ \mathbf{H}_{GB} & \mathbf{0} & \mathbf{H}_{GG} & \mathbf{H}_{GC} & \mathbf{0} \end{bmatrix} \tag{144}$$

where $\mathbf{H}_{ZX}$ denotes the Jacobians of measurement $Z$ regarding to state parameter $X$. These Jacobians are defined in Appendix E. The observability matrix $\mathbf{M}$ is defined as:

$$\mathbf{M} = \begin{bmatrix} \mathbf{M}_1 \\ \mathbf{M}_2 \\ \vdots \\ \mathbf{M}_k \end{bmatrix} \triangleq \begin{bmatrix} \frac{\partial \tilde{\mathbf{z}}_1}{\partial \tilde{\mathbf{x}}_1} \cdot \mathbf{\Phi}_{(1,1)} \\ \frac{\partial \tilde{\mathbf{z}}_2}{\partial \tilde{\mathbf{x}}_2} \cdot \mathbf{\Phi}_{(2,1)} \\ \vdots \\ \frac{\partial \tilde{\mathbf{z}}_k}{\partial \tilde{\mathbf{x}}_k} \cdot \mathbf{\Phi}_{(k,1)} \end{bmatrix} \tag{145}$$

where $\mathbf{z}_k$ and $\mathbf{x}_k$ denote the measurement and state at timestamp $t_k$; $\mathbf{\Phi}_{(k,1)}$ is the state transition matrix from timestamp $t_1$ to timestamp $t_k$ based on equation (143). The $k$-th block row of the observability matrix $\mathbf{M}$ can be written as:

$$\begin{aligned} \mathbf{M}_k &= \frac{\partial \tilde{\mathbf{z}}_k}{\partial \tilde{\mathbf{x}}_k} \cdot \mathbf{\Phi}_{(k,1)} \\ &= \begin{bmatrix} \mathbf{H}_{CB}\mathbf{\Phi}_B & \mathbf{0} & \mathbf{0} & \mathbf{H}_{CC}\mathbf{\Phi}_{calib} & \mathbf{H}_{CF}\mathbf{\Phi}_F \\ \mathbf{H}_{AB}\mathbf{\Phi}_B & \mathbf{H}_{AA}\mathbf{\Phi}_A & \mathbf{0} & \mathbf{H}_{AC}\mathbf{\Phi}_{calib} & \mathbf{0} \\ \mathbf{H}_{GB}\mathbf{\Phi}_B & \mathbf{0} & \mathbf{H}_{GG}\mathbf{\Phi}_G & \mathbf{H}_{GC}\mathbf{\Phi}_{calib} & \mathbf{0} \end{bmatrix} \\ &\triangleq \begin{bmatrix} \mathbf{M}_{CB} & \mathbf{0} & \mathbf{0} & \mathbf{M}_{CC} & \mathbf{M}_{CF} \\ \mathbf{M}_{AB} & \mathbf{M}_{AA} & \mathbf{0} & \mathbf{M}_{AC} & \mathbf{0} \\ \mathbf{M}_{GB} & \mathbf{0} & \mathbf{M}_{GG} & \mathbf{M}_{GC} & \mathbf{0} \end{bmatrix} \end{aligned} \tag{146}$$

where $\mathbf{M}_*$ are computed in Appendix F. If the $\mathbf{M}$ is of full column rank, the proposed system is fully observable. If we can find a null space matrix $\mathbf{N}$ s.t. $\mathbf{M} \cdot \mathbf{N} = \mathbf{0}$, the proposed system is unobservable and the $\mathbf{N}$ is denoted as unobservable directions.

By closely inspecting the observability matrix, we have the following Lemma:

**Lemma 1.** *The proposed MVIS has four unobservable directions* $\mathbf{N}$ *corresponding to the global yaw rotation and the global translation.*

$$\mathbf{N} = \begin{bmatrix} {}^{I_1}_{G}\mathbf{R}\,{}^{G}\mathbf{g} & \mathbf{0}_3 \\ -\lfloor {}^{G}\mathbf{p}_{I_1} \rfloor {}^{G}\mathbf{g} & \mathbf{I}_3 \\ -\lfloor {}^{G}\mathbf{v}_{I_1} \rfloor {}^{G}\mathbf{g} & \mathbf{0}_3 \\ \mathbf{0}_{30\times1} & \mathbf{0}_{30\times3} \\ --- & --- \\ {}^{I_{a1}}_{G}\mathbf{R}\,{}^{G}\mathbf{g} & \mathbf{0}_3 \\ -\lfloor {}^{G}\mathbf{p}_{I_{a_1}} \rfloor {}^{G}\mathbf{g} & \mathbf{I}_3 \\ -\lfloor {}^{G}\mathbf{v}_{I_{a_1}} \rfloor {}^{G}\mathbf{g} & \mathbf{0}_3 \\ \mathbf{0}_{30\times1} & \mathbf{0}_{30\times3} \\ --- & --- \\ {}^{I_{g1}}_{G}\mathbf{R}\,{}^{G}\mathbf{g} & \mathbf{0}_3 \\ \mathbf{0}_{9\times1} & \mathbf{0}_{9\times3} \\ --- & --- \\ \mathbf{0}_{27\times1} & \mathbf{0}_{27\times3} \\ --- & --- \\ -\lfloor {}^{G}\mathbf{p}_{f} \rfloor {}^{G}\mathbf{g} & \mathbf{I}_3 \end{bmatrix} \tag{147}$$

These four unobservable directions are inferred based on the 4 classical unobservable directions for a monocular VINS system (Hesch et al., 2014). Hence, we can interpret the $\mathbf{N}$ accordingly: the first column of $\mathbf{N}$ corresponds to the rotation about gravity, and the last three columns relate to the global positions ${}^{G}\mathbf{p}_{I}$, ${}^{G}\mathbf{p}_{I_a}$ and ${}^{G}\mathbf{p}_{f}$. From this lemma, we can conclude that the system observability will not be improved by simply adding more inertial sensors (IMUs or gyroscopes). It should also be pointed out that the velocity of IMU state will become unobservable if no visual measurements to static landmarks are available. Hence, without cameras, naively adding auxiliary IMUs will not significantly improve the system localization accuracy due to lack of global constraints to the base IMU velocity. It can be observed that the calibration parameters, including base IMU intrinsics $\mathbf{x}_{in}$, auxiliary IMU intrinsics $\mathbf{x}_{A_{in}}$, auxiliary gyroscope intrinsics $\mathbf{x}_{G_{in}}$ and spatiotemporal calibration $\mathbf{x}_{Ex}$, are highly related to the sensor motion. Under fully excited motions, these parameters are observable, which can be seen from our simulation results in Section 10.1.

## 9. Degenerate motion analysis

While the degenerate motions for the IMU-Camera spatiotemporal parameters, IMU intrinsics and camera intrinsics have been studied (Yang et al., 2019; 2020b; 2023b), in this paper, we for the first time, study the degenerate motions for the IMU-IMU and IMU-Gyroscope spatiotemporal calibration of MVIS.

### 9.1. *Spatiotemporal calibration of auxiliary inertial sensors*

In particular, we have identified the degenerate motions for the spatiotemporal calibration between the auxiliary inertial sensors and the base IMU, as summarized in Table 2, which will be explained in detail below. We refer interested readers to our companion technical report for the unobservable directions not reported below (Yang et al., 2023a).

*9.1.1. No rotation.* If the MVIS undergoes 3D motion but without rotation, the translation ${}^{I}\mathbf{p}_{I_a}$ between the auxiliary and base IMUs, the rotation ${}^{I}_{I_g}\mathbf{R}$ and time offset between the auxiliary gyroscope and the base IMU, will be unobservable. The unobservable directions $\mathbf{N}_{NR}$ are given by:

$$\mathbf{N}_{NR} = \begin{bmatrix} \mathbf{0}_{39\times3} & \mathbf{0}_{39\times3} & \mathbf{0}_{39\times1} \\ --- & --- & --- \\ \mathbf{0}_3 & \mathbf{0}_3 & \mathbf{0}_{3\times1} \\ \mathbf{I}_3 & \mathbf{0}_3 & \mathbf{0}_{3\times1} \\ \mathbf{0}_{33\times3} & \mathbf{0}_{33\times3} & \mathbf{0}_{33\times1} \\ --- & --- & --- \\ \mathbf{0}_3 & \mathbf{I}_3 & \mathbf{0}_{3\times1} \\ \mathbf{0}_{9\times3} & \mathbf{0}_{9\times3} & \mathbf{0}_{9\times1} \\ --- & --- & --- \\ \mathbf{0}_3 & \mathbf{0}_3 & \mathbf{0}_{3\times1} \\ {}^{I_1}_{G}\mathbf{R} & \mathbf{0}_3 & \mathbf{0}_{3\times1} \\ \mathbf{0}_{1\times3} & \mathbf{0}_{1\times3} & 0 \\ \mathbf{0}_3 & \mathbf{I}_3 & \mathbf{0}_{3\times1} \\ \mathbf{0}_{1\times3} & \mathbf{0}_{1\times3} & 1 \\ \mathbf{0}_{16\times3} & \mathbf{0}_{16\times3} & \mathbf{0}_{16\times1} \\ --- & --- & --- \\ \mathbf{0}_3 & \mathbf{0}_3 & \mathbf{0}_{3\times1} \end{bmatrix} \quad (148)$$

*9.1.2. One-axis rotation.* If the system undergoes 3D motion but with only one-axis rotation (which is common for aerial and ground vehicles), the translation ${}^{I}\mathbf{p}_{I_a}$ between the auxiliary and base IMUs, the rotation ${}^{I}_{I_g}\mathbf{R}$ between the auxiliary gyroscope and the base IMU will be unobservable, along with the rotation axis $\mathbf{k}$. Note that we, for the first time, explicitly found that one-axis rotation will cause the rotation calibration between the auxiliary gyroscope and base IMU to become unobservable. We verify this finding with simulations in Section 10.2. Specifically the additional unobservable directions are given by:

**Table 2.** Degenerate motions with related unobservable parameters for auxiliary IMU and gyroscope.

| Motion types | Auxiliary IMU | Auxiliary gyroscope |
|---|---|---|
| No rotation | ${}^{I}\mathbf{p}_{I_a}$ | ${}^{I}_{I_g}\mathbf{R}$ and $t_{d_g}$ |
| One-axis rotation | ${}^{I}\mathbf{p}_{I_a}$ along rot. axis | ${}^{I}_{I_g}\mathbf{R}$ along rot. axis |
| Constant ${}^{I}\boldsymbol{\omega}$ | ${}^{I}\mathbf{p}_{I_a}$along rot. axis | ${}^{I}_{I_g}\mathbf{R}$ along rot. axis, $t_{d_g}$ |
| Constant ${}^{I}\boldsymbol{\omega}$ and ${}^{G}\mathbf{v}_I$ | ${}^{I}\mathbf{p}_{I_a}$ along rot. axis, $t_{d_a}$ | ${}^{I}_{I_g}\mathbf{R}$ along rot. axis, $t_{d_g}$ |
| Constant ${}^{I}\boldsymbol{\omega}$ and ${}^{I}\mathbf{v}$ | ${}^{I}\mathbf{p}_{I_a}$ along rot. axis | ${}^{I}_{I_g}\mathbf{R}$ along rot. axis, $t_{d_g}$ |

$$
\mathbf{N}_{OA} = \begin{bmatrix}
\mathbf{0}_{39\times1} & \mathbf{0}_{39\times1} \\
--- & --- \\
\mathbf{0}_{3\times1} & \mathbf{0}_{3\times1} \\
{}^{G}_{I_1}\mathbf{R}\,{}^{I_1}\mathbf{k} & \mathbf{0}_{3\times1} \\
\mathbf{0}_{33\times1} & \mathbf{0}_{33\times1} \\
--- & --- \\
\mathbf{0}_{3\times1} & {}^{I_{g1}}\mathbf{k} \\
\mathbf{0}_{9\times1} & \mathbf{0}_{9\times1} \\
--- & --- \\
\mathbf{0}_{3\times1} & \mathbf{0}_{3\times1} \\
{}^{I_1}\mathbf{k} & \mathbf{0}_{3\times1} \\
0 & 0 \\
\mathbf{0}_{3\times1} & {}^{I_{g1}}\mathbf{k} \\
0 & 0 \\
\mathbf{0}_{16\times1} & \mathbf{0}_{16\times1} \\
--- & --- \\
\mathbf{0}_{3\times1} & \mathbf{0}_{3\times1}
\end{bmatrix} \tag{149}
$$

*9.1.3. Constant local angular velocity.* If the MVIS undergoes non-zero constant local angular velocity with random 3D translation for the base IMU, the translation ${}^{I}\mathbf{p}_{I_a}$ between the base and auxiliary IMUs is still unobservable along the rotation axis. In addition, the rotation ${}^{I}_{I_a}\mathbf{R}$ and the time offset $t_{d_g}$ between the base IMU and the auxiliary gyroscope become unobservable.

*9.1.4. Constant local angular and global linear velocity.* If the MVIS undergoes non-zero constant ${}^{I}\boldsymbol{\omega}$ and constant ${}^{G}\mathbf{v}_I$ for base IMU, the translation and time offsets between the base and auxiliary IMUs, the rotation and time offset between the base IMU and the auxiliary gyroscope become unobservable.

*9.1.5. Constant local angular and linear velocity.* When the ${}^{I}\boldsymbol{\omega}$ and ${}^{I}\mathbf{v}$ are constant and non-zero for the base IMU, the time offset $t_d$ between the base IMU and camera, as well as the time offset $t_{d_g}$ between the base IMU and the gyroscope are both unobservable. However, the time offset $t_{d_a}$ between the base and auxiliary IMUs is still observable (see Figure 6), which is unexpected. This is due to the fact that the local constant velocity assumption will be invalid for the auxiliary IMU if the base IMU is undergoing constant local linear and angular velocity. The local angular velocity and acceleration of the auxiliary IMU can be represented as:

$$
{}^{I_a}\boldsymbol{\omega} = {}^{I_a}_{I}\mathbf{R}\,{}^{I}\boldsymbol{\omega} \tag{150}
$$

$$
{}^{I_a}\mathbf{a}_b = {}^{I_a}_{I}\mathbf{R}\left({}^{I}\mathbf{a}_b + \lfloor {}^{I}\boldsymbol{\alpha}\rfloor {}^{I}\mathbf{p}_{I_a} + \lfloor {}^{I}\boldsymbol{\omega}\rfloor \lfloor {}^{I}\boldsymbol{\omega}\rfloor {}^{I}\mathbf{p}_{I_a}\right) \tag{151}
$$

where ${}^{I}\boldsymbol{\alpha}$ refers to the angular acceleration of the base IMU, the local sensor body acceleration ${}^{I}\mathbf{a}_b$ is defined in equation (3). If the base IMU undergoes constant local linear and angular velocity motion, the angular velocity of the auxiliary IMU ${}^{I_a}\boldsymbol{\omega}$ is also constant [see Eq. (150)]. The ${}^{I}\mathbf{a}_b$ and ${}^{I}\boldsymbol{\alpha}$ of the base IMU should be zeros. Hence, equation (151) yields:

$$
{}^{I_a}\mathbf{a}_b = {}^{I_a}_{I}\mathbf{R}\left(\lfloor {}^{I}\boldsymbol{\omega}\rfloor \lfloor {}^{I}\boldsymbol{\omega}\rfloor {}^{I}\mathbf{p}_{I_a}\right) \tag{152}
$$

If ${}^{I}\boldsymbol{\omega}$ is constant but not zero, the local linear acceleration ${}^{I_a}\mathbf{a}_b$ should not be zero. This breaks the local constant linear velocity assumption for the auxiliary IMU. Furthermore, we find that the norm of ${}^{I_a}\mathbf{v}$ is constant:

$$
{}^{I_a}\mathbf{v} = {}^{I_a}_{I}\mathbf{R}\left({}^{I}\mathbf{v} + \lfloor {}^{I}\boldsymbol{\omega}\rfloor {}^{I}\mathbf{p}_{I_a}\right) \tag{153}
$$

But the non-zero acceleration ${}^{I_a}\mathbf{a}_b$ will cause the bearing change of local velocity, which makes the time offset between the base and auxiliary IMUs observable. This is further verified through our simulation results (see Figure 6).

## 9.2. Intrinsics for auxiliary inertial sensors

In our previous work (Yang et al., 2020b; 2023b), the degenerate motions of IMU intrinsics for monocular VINS have been studied. In this work, we have found that the degenerate motion primitives in (Yang et al., 2023b) still hold for the auxiliary IMU intrinsics with our inertial model choice (see Table 3). Note that fully excited motions are needed in order to make all intrinsic parameters observable for the auxiliary IMUs/gyroscopes.

**Table 3.** Summary of basic degenerate motions for auxiliary inertial intrinsics calibration (any combinations of these unit motion primitives are degenerate).

| Motion types | Nullspace Dim. | Unobservable parameters |
|---|---|---|
| Constant ${}^{A_w}\omega_1$ | 1 | $d_{A_{w1}}$ |
| Constant ${}^{A_w}\omega_2$ | 2 | $d_{A_{w2}}, d_{A_{w3}}$ |
| Constant ${}^{A_w}\omega_3$ | 3 | $d_{A_{w4}}, d_{A_{w5}}, d_{A_{w6}}$ |
| Constant ${}^{A_a}a_1$ | 3 | $d_{A_{a1}}$, pitch and yaw of ${}^{I_a}_{A_a}\mathbf{R}$ |
| Constant ${}^{A_a}a_2$ | 3 | $d_{A_{a2}}, d_{A_{a3}}$, roll of ${}^{I_a}_{A_a}\mathbf{R}$ |
| Constant ${}^{A_a}a_3$ | 3 | $d_{A_{a4}}, d_{A_{a5}}, d_{A_{a6}}$ |
| Constant ${}^{I_a}a_1$ | 3 | $t_{A_{g1}}, t_{A_{g2}}, t_{A_{g3}}$ |
| Constant ${}^{I_a}a_2$ | 3 | $t_{A_{g4}}, t_{A_{g5}}, t_{A_{g6}}$ |
| Constant ${}^{I_a}a_3$ | 3 | $t_{A_{g7}}, t_{A_{g8}}, t_{A_{g9}}$ |
| Constant ${}^{G_w}\omega_1$ | 1 | $d_{G_{w1}}$ |
| Constant ${}^{G_w}\omega_2$ | 2 | $d_{G_{w2}}, d_{G_{w3}}$ |
| Constant ${}^{G_w}\omega_3$ | 3 | $d_{G_{w4}}, d_{G_{w5}}, d_{G_{w6}}$ |

Note that $d_{A*}$ is column-wise element from $\mathbf{D}_{A*}$, $d_{G_w}$ is column-wise element from $\mathbf{D}_{G_w}$. $t_{A_{gi}}, i = 1\ldots9$ are the elements from $\mathbf{T}_{A_g}$ for g-sensitivity.

## 10. Simulation results

The simulator, which is provided within the OpenVINS project (Geneva et al., 2020) along with the multi-IMU and RS extension from (Eckenhoff et al., 2021) and IMU intrinsic extension (Yang et al., 2023b), is leveraged to provide synthetic measurements with perfect groundtruth for verification of the proposed MVIS under different motion conditions. In the simulation, one base IMU `IMUb`, one auxiliary IMU `IMUa0`, one auxiliary gyroscope `IMUa1`, one global shutter (GS) camera `CAM0` and one rolling shutter (RS) camera `CAM1` are simulated. Note that both cameras are simulated with 10 Hz frame rates. The basic configuration of the simulator is listed in Table 4. The three trajectories used for the simulation are as follows:

- Fully excited motion (left of Figure 2): All axes of the accelerometer and gyroscope are fully excited with a general 3D handheld trajectory.
- One-axis motion (middle of Figure 2): The sensor suite moves in 3D space but with only yaw rotation. The trajectory is modified based on *tum_room1* (Schubert et al., 2018).
- Circular-planar motion (right of Figure 2): The sensor suite moves in x-y plane with constant local angular and linear velocities.

Specifically, we first build a B-spline with trajectory keyframes of the base IMU trajectory. Then, we can compute the acceleration of base IMU by calculating double derivatives for the position component of the B-spline at specified time stamp. We leverage rigid body constraints between base and auxiliary IMU to simulate the auxiliary IMU readings. The base IMU acceleration can be transferred to the auxiliary IMU frame with the groundtruth angular velocity and acceleration, which are also computed from derivatives of base IMU B-spline. The angular velocity of the auxiliary IMU can be simply computed with angular velocity from base IMU and the rigid rotation between based IMU and auxiliary IMU. Then white Gaussian noises are added to the auxiliary IMU readings based on equations (1) and (2).

To simulate RS visual bearing measurements, we follow the same logic in (Eckenhoff et al., 2021; Li and Mourikis, 2014; Yang et al., 2023b). Static environmental features are first generated along the trajectory at random depths and bearings. Then, for a given imaging time, we project each feature in view into the current image frame using the true camera intrinsic and distortion model and find the corresponding observation row. Given this projected row and image time, we can find the pose at which that RS row should have been exposed. We can then re-project this feature into the new pose and iterate until the projected row does not change (which typically requires 2–3 iterations). We now have a feature measurement which occurs at the correct pose given its RS row. This measurement is then corrupted with white noise. The imaging timestamp corresponding to the starting row is then shifted by the true IMU-Camera time offset $t_d$ to simulate cross-sensor delay. In the following simulations, the `RPNG` IMU model (see Section 3.1) is used to be aligned with the analysis.

**Table 4.** Simulation parameters and prior standard deviations that perturbations of measurements and initial states were drawn from.

| Parameter | Value | Parameter | Value |
|---|---|---|---|
| IMU Dw | 0.003 | IMU Da | 0.003 |
| Rot. atoI (rad) | 0.003 | IMU Tg | 0.001 |
| Gyro. Noise ($\mathrm{rad\,s^{-1}\sqrt{Hz^{-1}}}$) | 1.696e-04 | Gyro. Bias ($\mathrm{rad\,s^{-2}\sqrt{Hz^{-1}}}$) | 1.939e-05 |
| Accel. Noise ($\mathrm{m\,s^{-2}\sqrt{Hz^{-1}}}$) | 2.000e-3 | Accel. Bias ($\mathrm{m\,s^{-3}\sqrt{Hz^{-1}}}$) | 3.000e-3 |
| Focal Len. (px/m) | 1.0 | Cam. Center (px) | 1.0 |
| d1 and d2 | 0.002 | d3 and d4 | 0.002 |
| Rot. CtoI (Hz) | 0.004 | Pos. IinC (m) | 0.008 |
| Pixel Proj. (px) | 1 | Cam-IMU Toff (s) | 0.008 |
| IMU-IMU Toff (s) | 0.003 | Gyro-IMU Toff (s) | 0.003 |
| Rot. IatoIb (rad) | 0.003 | Pos. IainIb (m) | 0.005 |
| Cam freq. (Hz) | 10/10 | IMU freq. (Hz) | 250/300/200 |

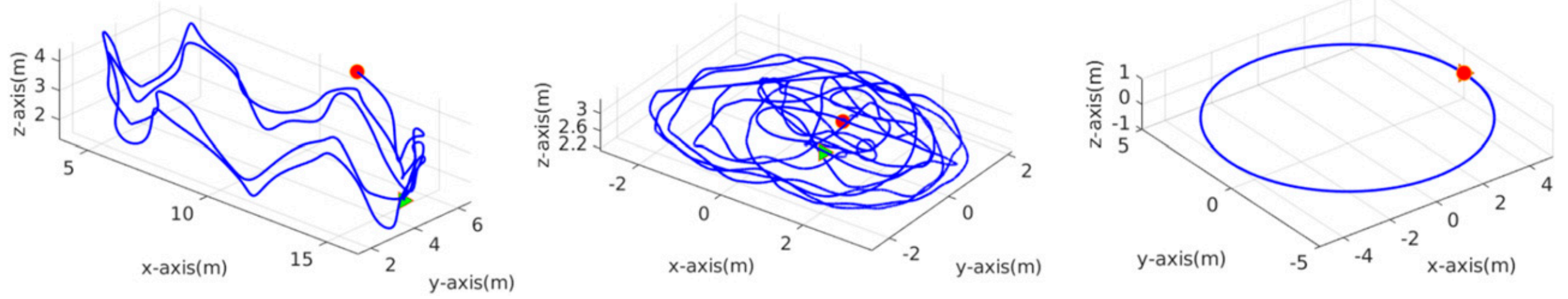

**Figure 2.** Simulated trajectories. Left: *calib_3d* with fully excited 3D motion, total length: 89.4 m; Middle: *tum_room* with one axis rotation and 3D translation, total length: 134.5 m; Right: *circle_planar* with circular planar motion (constant z and only yaw rotation), total length: 157.1 m. The green triangle and red circle denote the beginning and ending of these trajectories, respectively.

It is important to note that, in the following sections, we only present the most prominent results due to space limits, while comprehensive simulation and experimental results can be found in our companion technical report (Yang et al., 2023a).

## 10.1. Fully excited motion

We first evaluate the proposed system on a general 3D handheld trajectory, see Figure 2, which fully excites all 6 axes of the sensor platform. To save space, only selected parameters are presented, but all parameters are perturbed and estimated during our simulation runs. The camera related results are shown in Figure 3 while the IMU related results are shown in Figure 4. For each figure, there are five different runs with different initial state perturbations.

It is clear that all parameters are able to converge toward the true value within the first 20–40 s of the trajectory, which verifies our conclusion that all the calibration parameters for MVIS are observable given fully excited motions. These results also verify that the proposed MVIS indeed is able to perform calibration of all parameters for visual and inertial sensors.

## 10.2. Degenerate one-axis motion

We now perform a simulation where the trajectory only exhibits one-axis rotation about the global $z$-axis to verify our identified degenerate motion, see Figure 2. Shown in Figure 5, multiple parameters are unable to converge with either estimation errors or estimation uncertainties ($3\sigma$ bounds). This matches the parameters which we have identified as unobservable under this motion. We can see that the 3 parameters $d_{w1}$, $d_{w2}$, and $d_{w3}$ for both the base IMU `IMUb`, auxiliary IMU `IMUa0`, and auxiliary gyroscope `IMUa1` are unable to be calibrated. Additionally, the y component for the rigid position between the camera to base IMU (`CAM0-IMUb Pos.`) cannot converge at all. The z component for the position of the auxiliary IMU to base IMU (`IMUa0-IMUb Pos.`) is unable to be calibrated as expected. Note that the y component of ${}^{I}\mathbf{p}_C$ and the z component of ${}^{I}\mathbf{p}_{I_a}$ are all along the rotation axis which is degenerate.

Furthermore, it can be seen that we are unable to calibrate a portion of the relative rotation between the base IMU and auxiliary gyroscope (`IMUa1-IMUb Ori.`) is one-axis rotation, which can be calibrated nicely in the fully excited motion case. This further confirms our degenerate motion analysis summarized in Table 2.

## 10.3. Degenerate circular-planar motion

We also perform a simulation where the sensors follow a circular-planar motion shown in Figure 2. This is a typical example motion of constant angular and linear velocity. The translation and the time offset of `CAM0-IMUb` and `CAM1-IMUb`, the translation of `IMUa0-IMUb` and the orientation of `IMUa1-IMUb` are not observable. Shown in Figure 6, their $3\sigma$ bounds and estimate errors are kept as almost straight lines and do not converge at all. These results further verify our identified degenerate motions shown in Table 2.

The time offset between auxiliary gyroscope and based IMU (`IMUa1-IMUb Toff`) is unable to be calibrated, while the time offset between auxiliary IMU to base IMU (`IMUa0-IMUb Toff`) is still observable. This can be seen by the estimation errors converging in Figure 6 and thus verifies our degenerate motion analysis in Section 9.1.5. Note that the rolling shutter readout time of the `CAM1` converges quite slowly, given that the sensor motion is not fully excited.

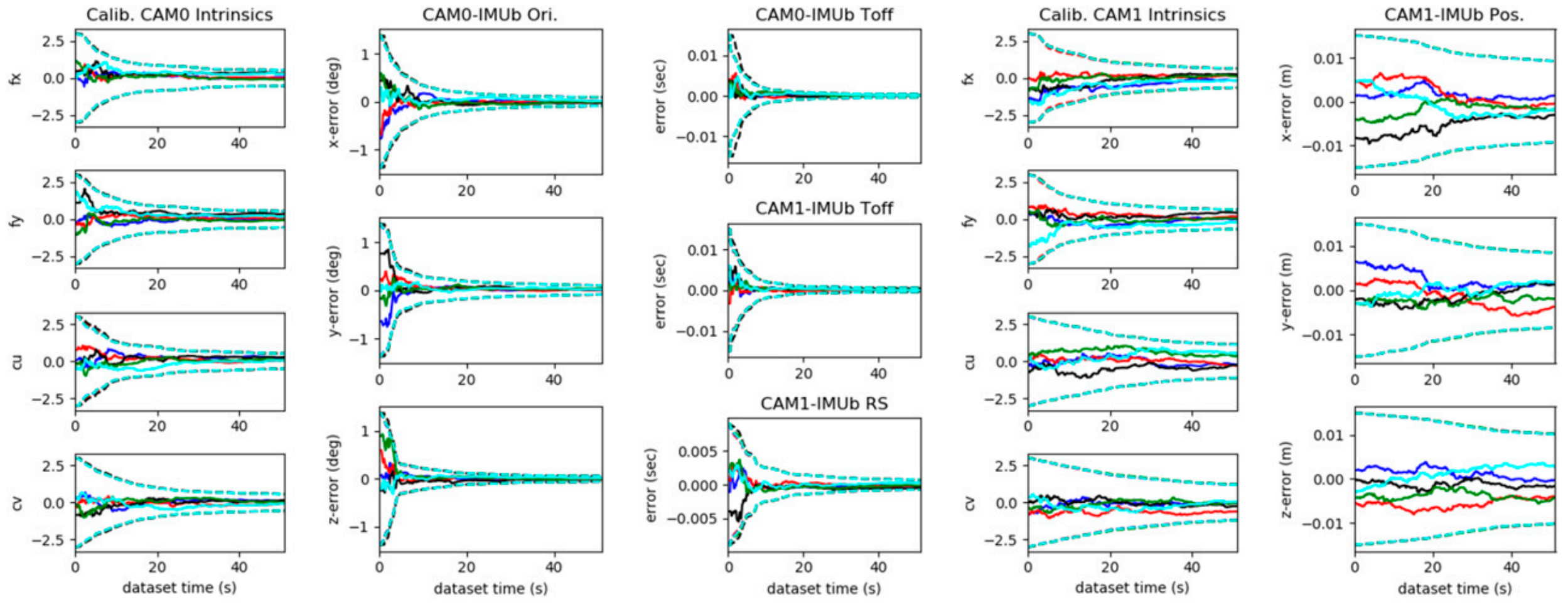

**Figure 3.** Simulation results for *fully excited motion*. All the cameras (`CAM0` and `CAM1`) related parameters converge nicely. $3\sigma$ bounds (dashed line) and estimation errors (solid line) are plotted for five different runs (shown in different colors) with different initial calibration perturbations.

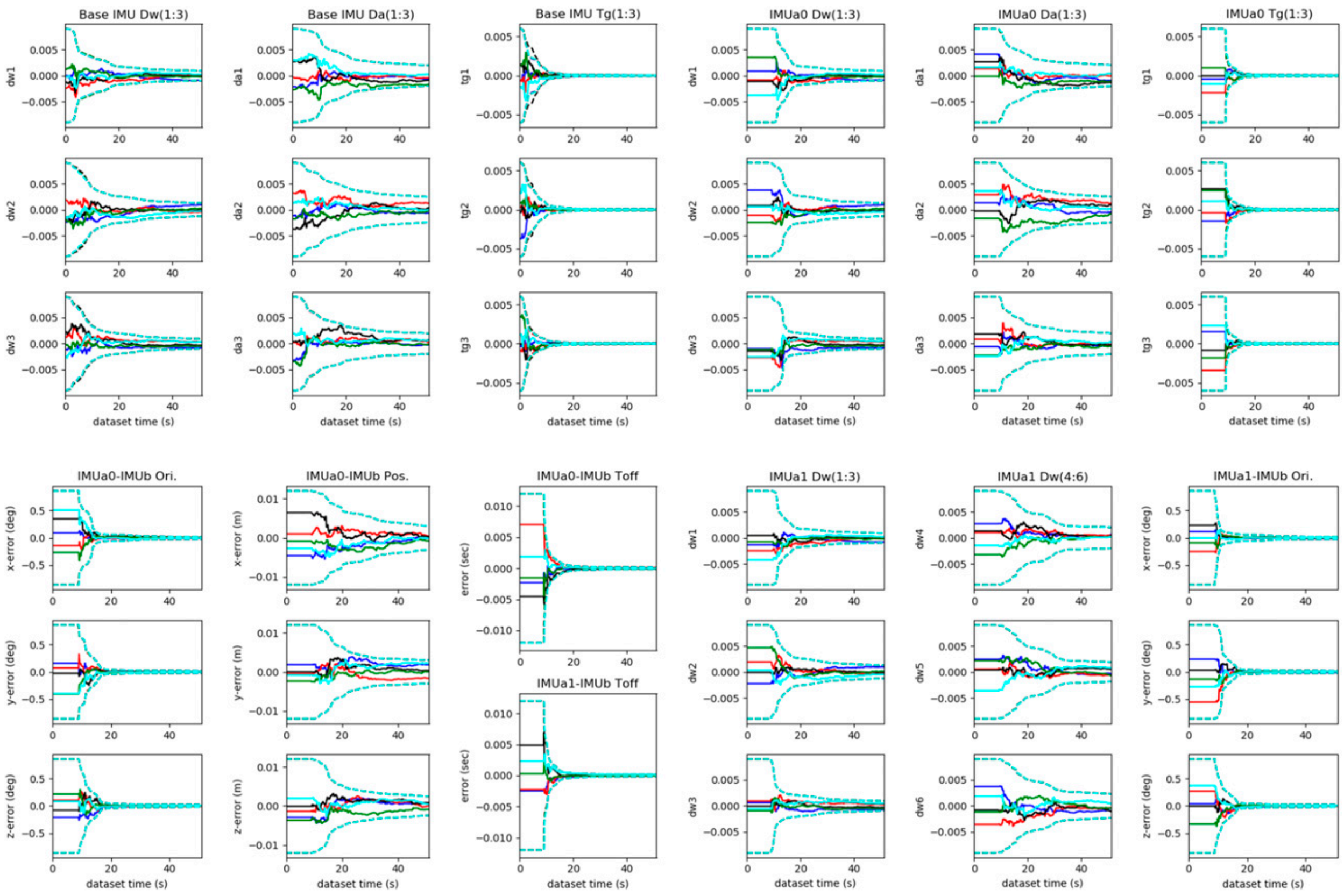

**Figure 4.** Simulation results for *fully excited motion*. All the base IMU (`IMUb`) and auxiliary IMUs (`IMUa0`, `IMUa1`) related parameters converge nicely. $3\sigma$ bounds (dashed line) and estimation errors (solid line) are plotted for five different runs (shown in different colors) with different initial calibration perturbations.

The calibration results for IMU related intrinsics are shown in Figure 7. It is clear that the gyroscope correction $\mathbf{D}_w$ and the accelerometer correction $\mathbf{D}_a$ do not converge at all. The convergence of g-sensitivity $\mathbf{T}_g$ also becomes much worse compared to fully excited motion in Figure 4 which results from fully excited motions.

## 11. Experimental results

The proposed self-calibration system is further evaluated using our own visual-inertial sensor rig (VI-Rig) as shown in Figure 8. Specifically, it contains a MS-GX-25, MS-GX-35, Xsens MTi 100, FLIR blackfly camera, RealSense T265 tracking camera (which contains an integrated BMI055 IMU along with a fisheye stereo global shutter camera), and 640x480 ELP-960P2CAM-V90-VC USB 2.0 RS-stereo camera. We perform three sets of experiments.[2]

- Fully excited motion with 4 IMUs + 3 GS Cameras.
- Fully excited motion with 4 IMUs + 2 GS Cameras + 2 RS Cameras.
- Planar motion with 4 IMUs + 2 GS cameras.

In these experiments, we evaluate the intrinsic calibration with `Kalibr` model [see Section 3.1], in order to facilitate a direct comparison to Kalibr–the calibration toolbox (Furgale et al., 2013). We also investigate if the joint calibration performance changes with different number of IMU/camera sets. In addition, planar motion, one of the most commonly seen degenerate motions, is also investigated to show its effects on calibration. The results further verify our degeneration motion analysis and have significant practical implications on practitioners performing calibration on constrained autonomous platforms (e.g. aerial or ground vehicles).

The boxplots are used to demonstrate the calibration results for the proposed MVIS and Kalibr. When drawing the boxplots for the translation part of extrinsics, the camera intrinsics and time offsets, we use the average estimates of the MVIS with all available sensors as reference value and then compute the error of each estimate from Kalibr or MVIS to this reference. When drawing the boxplots for the orientation extrinsic, we select the first estimate of MVIS with all available sensors for reference value. The middle line of each boxplot indicates the average errors while the red star +

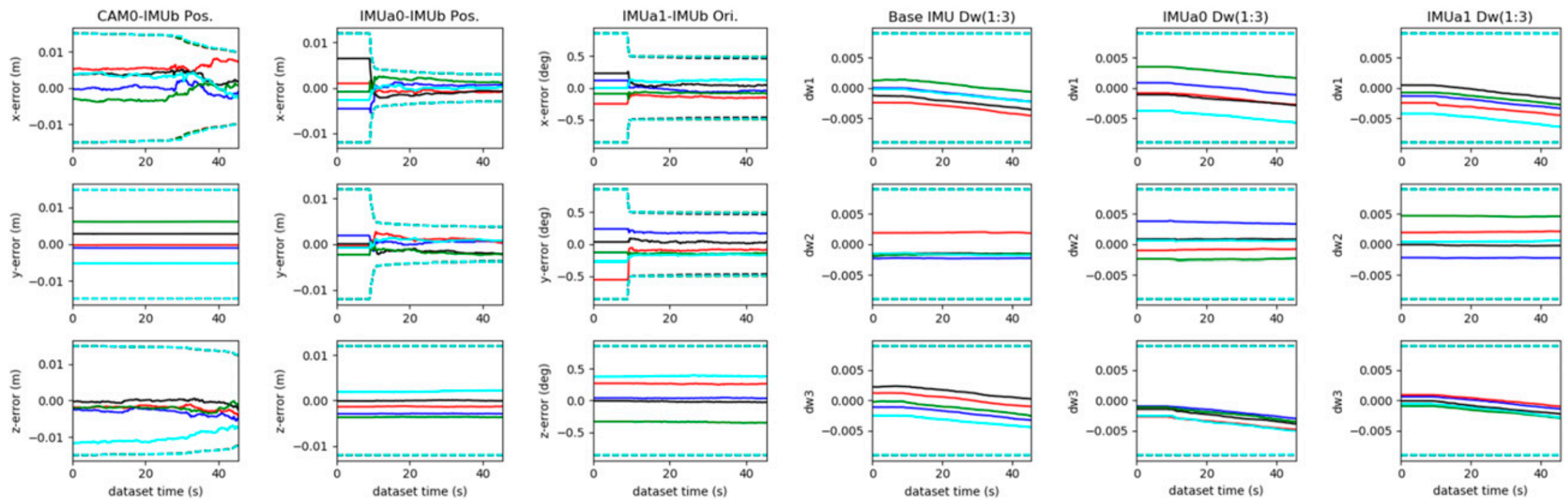

**Figure 5.** Simulation results for *one-axis motion*. The translation of `CAM0-IMUb` (y component) and `IMUa0-IMUb` (z component), the rotation of `IMUa1-IMUb` (z component), the $d_{w1}$, $d_{w2}$, and $d_{w3}$ of `IMUb`, `IMUa0`, and `IMUa1` show inability to converge (sigma bound does not decrease due to no information gain). $3\sigma$ bounds (dashed line) and estimation errors (solid line) are plotted for five different runs (shown in different colors) with different initial calibration perturbations.

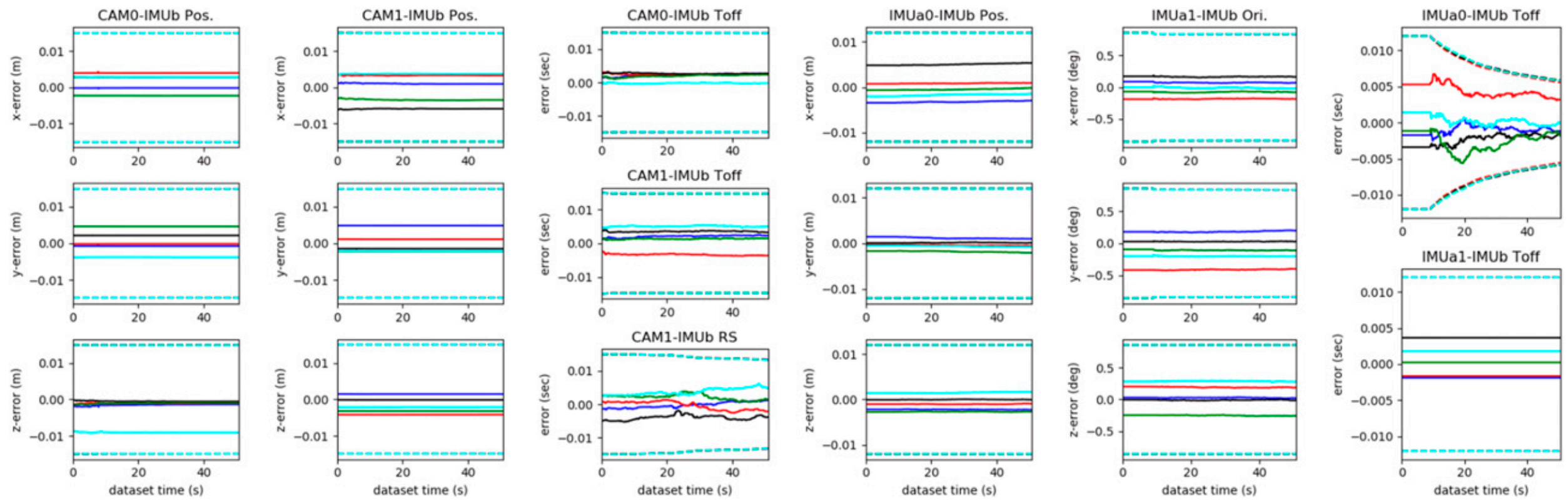

**Figure 6.** Simulation results for *circular-planar motion*. The translation of `CAM0-IMUb`, `CAM1-IMUb`, `IMUa0-IMUb`, the rotation of `IMUa1-IMUb`, the time offset of `CAM0-IMUb`, `CAM1-IMUb`, and `IMUa1-IMUb` all show inability to converge (sigma bound does not decrease due to no information gain). Note that the RS readout time of `CAM1` and the time offset of `IMUa0-IMUb` converge slower due to less motion excitation. $3\sigma$ bounds (dashed line) and estimation errors (solid line) are plotted for five different runs (shown in different colors) with different initial calibration perturbations.

indicates outliers. IMU intrinsics are computed relative to the "ideal" inertial model, with identity matrices for gyroscope correction $\mathbf{D}'_w$, accelerometer correction $\mathbf{D}'_a$ and ${}^I_w\mathbf{R}$, except for g-sensitivity, $\mathbf{T}_g$, which is set as zero matrix.

## 11.1. Four IMUs + three GS cameras

All the four IMUs, FLIR blackfly camera, and the GS stereo camera from RealSense T265 are used for this evaluation.

All cameras used in this experiments are not rolling shutter to ensure fair comparison against the baseline Kalibr (Furgale et al., 2013), which only supports IMU-Camera calibration with global shutter cameras. Total 10 datasets were collected with an AprilTag board, on which both the proposed system and the Kalibr calibration toolbox were run to evaluate the calibration accuracy and repeatability statistics on all calibration parameters.

During data collection, all 6-axis motion of the VI-Rig was excited to avoid degenerate motions for calibration parameters.

*11.1.1. Calibration with different number of cameras.* When running Kalibr, all the IMUs and cameras are used to achieve the best calibration results from Kalibr. When running our proposed MVIS, we use all the four IMUs with 1/2/3 camera, respectively. In this way, we can evaluate how the number of used cameras affects the calibration performance.

The final converged estimates of the calibration parameters from both systems on these 10 datasets can be shown in the box plots in Figure 9. The proposed MVIS was run with one (green), two (black), and three (blue) of the cameras. The baseline Kalibr (magenta) was run on all three cameras. The $x$-axis of figures in the second and third row denotes the base IMU (GX-25 `IMUb`) as `b` and auxiliary IMU (GX-35 `IMUa0`, Xsens `IMUa1`, and T265 IMU `IMUa2`) as `0`, `1`, and `2` respectively. Note that the camera intrinsics are required to be fixed for Kalibr when performing IMU-Camera calibration. Hence there is only one value for each camera intrinsics for Kalibr in Figure 9.

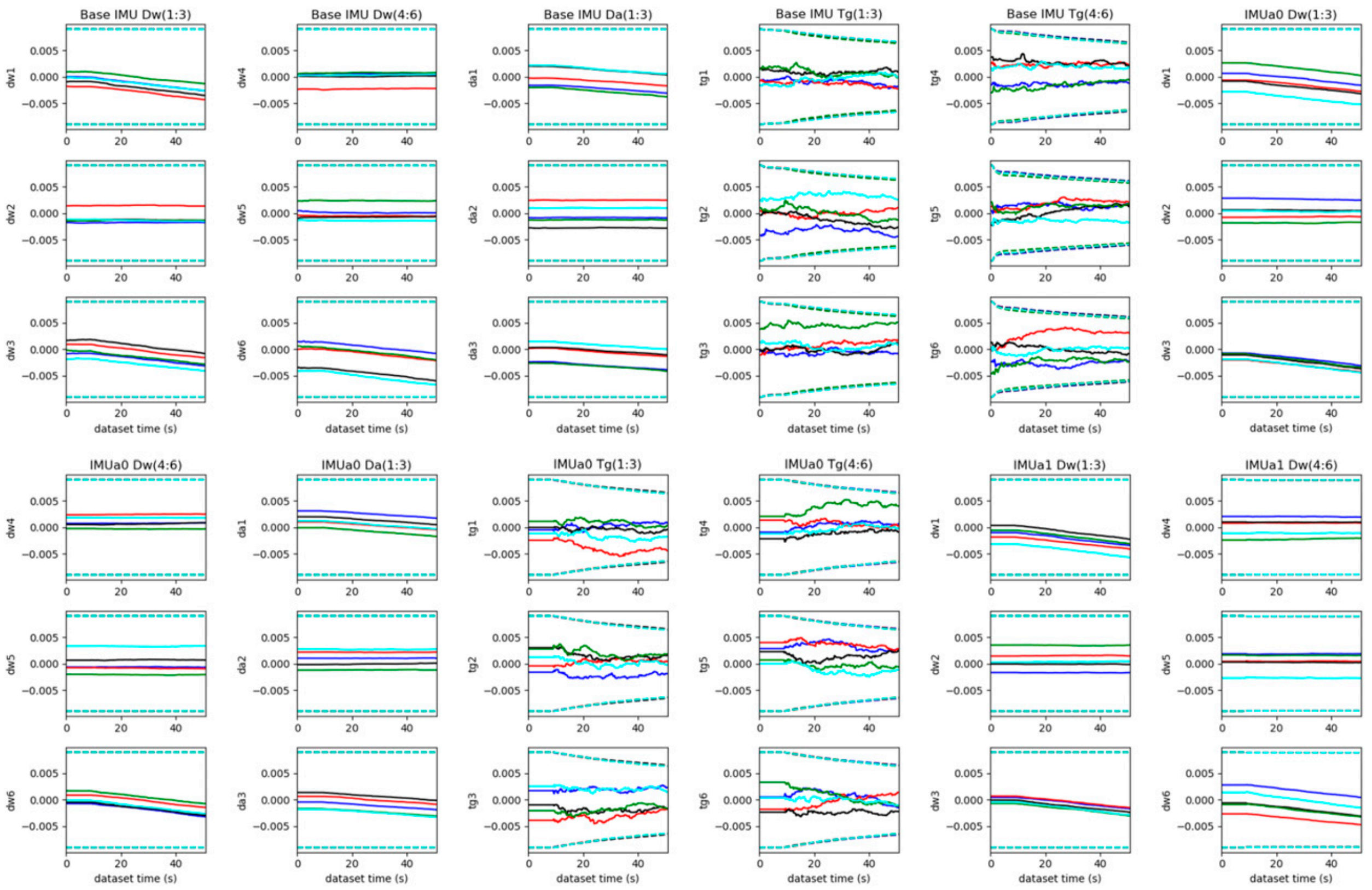

**Figure 7.** Simulation results for *circular-planar motion*. The gyroscope and acceleration related parameters (gyroscope correction $\mathbf{D}_w$, accelerometer correction $\mathbf{D}_a$, and g-sensitivity $\mathbf{T}_g$) for both base and auxiliary IMUs do not converge or converge much slower than the case of fully excited motions. $3\sigma$ bounds (dashed line) and estimation errors (solid line) are plotted for five different runs (shown in different colors) with different initial calibration perturbations.

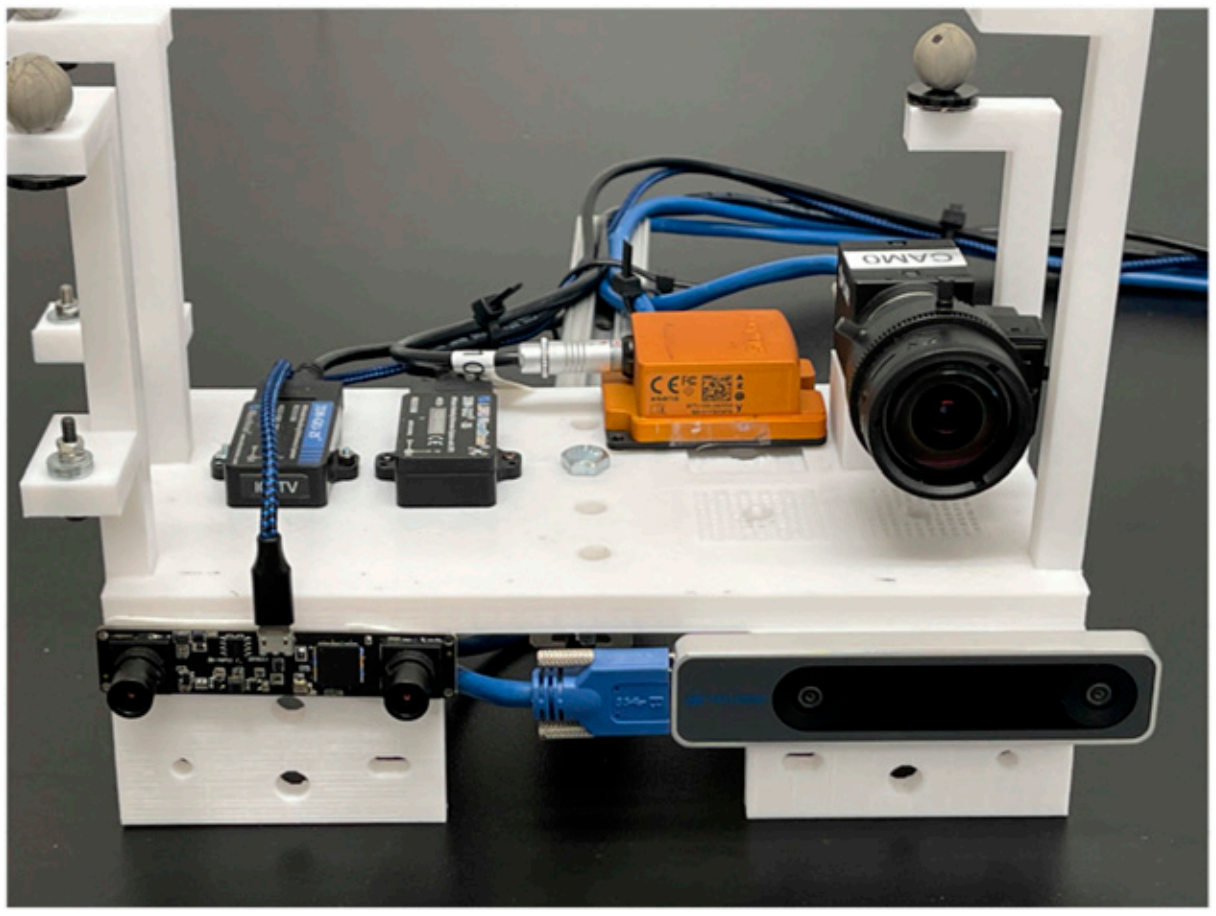

**Figure 8.** The self-assembled sensor rigs in real-world experiments, containing one Mircostrain GX-25, one MircroStrain GX-35, one MTI Xsens IMU, one BalckFly camera, one IntelRealsense T265 tracking camera (with a GS fisheye stereo camera and an BMI055 IMU inside), and one ELP stereo RS camera.

The range of the boxplot in the figure indicates the convergence repeatability of calibration parameters. The proposed MVIS needs an initial guess for the calibration parameters to start the optimization and the initial guess distribution are shown in the first row of Figure 9 for the proposed method. The initial guess for $d_2$ of `CAM0` distortion model is within ±0.5 while the final estimated values are between 0 and 0.1. The initial guess for time offset for `CAM0-IMUb` is within ±5 ms, while the final converged values from the proposed MVIS are most cases around ±0.5 ms. These results show that the calibration parameters can converge robustly with the proposed MVIS.

It can be observed from Figure 9 that the calibration estimation convergence of IMU/camera intrinsics and `CAM0-IMUb` translation are better in blue color than those in green or black colors, which indicate that more cameras can be used to improve overall calibration convergence. This is probably due to improved visual feature estimates from longer feature tracks or wider field-of-view due to multi-view constraints when more cameras are used in the experiment.

*11.1.2. Comparison with Kalibr.* One major difference between the Kalibr and our MVIS is how to represent the sensor-platform's trajectory. Kalibr leverages the camera poses, based on AprilTag-based visual odometry (VO), to build a B-spline-based continuous-time representation of the trajectory. As such, the IMU and camera measurements can be easily modeled with the poses at any time given by this B-spline. Kalibr optimizes both the calibration parameters and the B-spline parameters (knots).

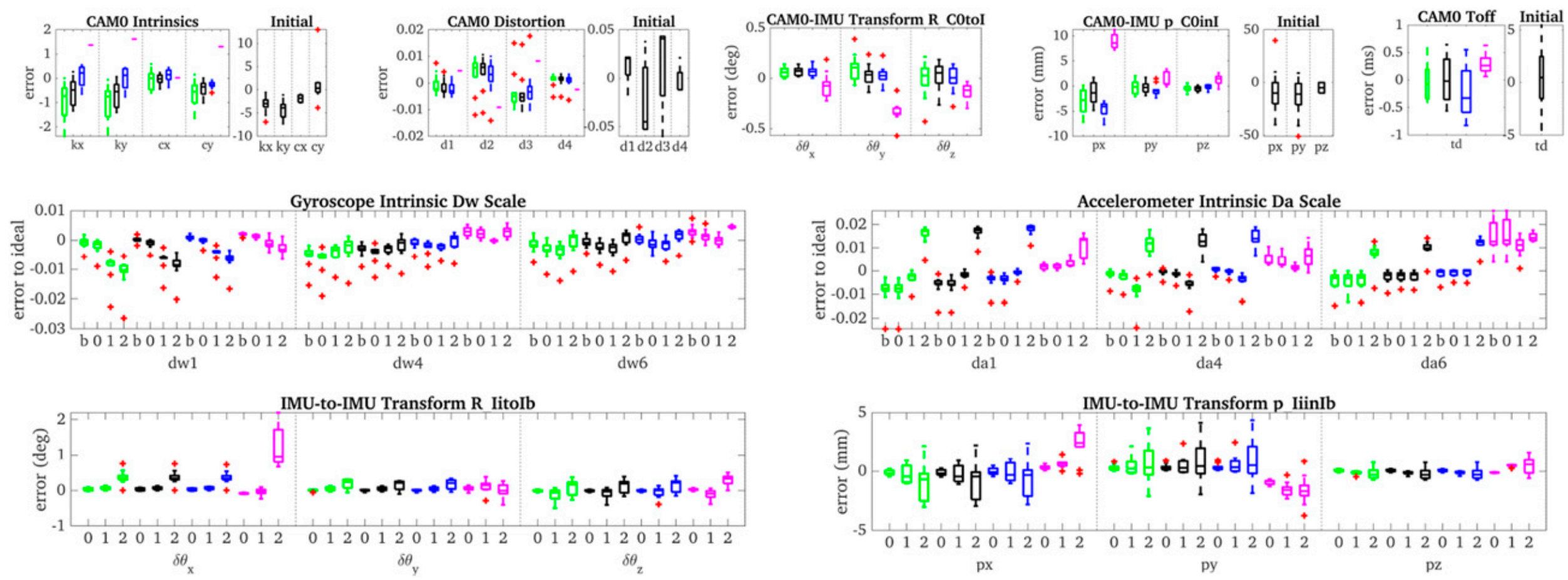

**Figure 9.** Calibration results for CAM0 and four IMUs related parameters over 10 datasets collected with Intel Realsense T265 (GS), FLIR Blackfly camera (GS), and four IMUs. The proposed MVIS was run with one (green), two (black), and three (blue) of the cameras. The baseline Kalibr (magenta) was run on all three cameras and all four IMUs. The $x$-axis of figures in the second and third row denotes the base IMU (IMUb) as b and auxiliary IMU (IMUa0-IMUa2) as 0, 1, and 2 respectively. For IMU-IMU transformation {R_IitoIb, p_IiinIb}, i=0, 1, 2. Note that the calibration convergence of camera/IMU intrinsics and camera to IMU translation are improved if more cameras are used.

In contrast, the MVIS represents the trajectory using the discrete-time IMU states at imaging times. The IMU measurements are used to build the IMU factors with pre-integration while the asynchronous visual/auxiliary inertial measurements are related to the trajectory through linear interpolation (see Sections 6 and 7). The mean values of each boxplot in Figure 9 clearly demonstrate that the proposed MVIS achieves competing calibration accuracy as the Kalibr.

*11.1.3. IMU intrinsic quality.* By evaluating the IMU calibration results across the four IMUs of IMUb, IMUa0, IMUa1, and IMUa2 (denoted as b, 0, 1, and 2 in the second and third row in Figure 9) used in the experiments, we clearly see that IMUa2, a relatively low-cost BMI055 IMU, demonstrates larger scale correction for gyroscope and accelerometer than other three high-end IMUs (GX-25 IMUb, GX-35 IMUa0, and Xsens IMUa1). This is expected as the IMUb, IMUa0, and IMUa1 are supposed to have more stable and sophisticated factory calibration than IMUa2. This result, aligned with our previous work for single IMU-Camera calibration (Yang et al., 2023b), further validates the proposed MVIS can generate reasonable and accurate calibration for IMUs.

*11.1.4. Timing evaluation.* While the B-spline-based continuous-time representation naturally enables the incorporation of asynchronous sensor measurements, multiple high-frequency IMUs can easily causes thousands of IMU cost terms to be inserted into the optimization, significantly increasing the computation. The proposed MVIS leverages ACI$^3$ to integrate all the IMU measurements from one IMU between two image frames into one single factor, which greatly simplifies the graph and saves computation. To see this, in our experiments, we also recorded the processing time of the proposed MVIS and Kalibr when running the calibration with all the IMUs and cameras on these 10 datasets.

Specifically, a ThinkPad P52 with Intel i7-8850H CPU@ 2.60GHz×12 and 32 GB RAM was used. The average processing time for MVIS is approximately 807.4 s and 2208.3 s for Kalibr. We attribute the computation savings of the proposed MVIS primarily because (1) Kalibr by default has 100 knot poses per second for constructing the B-Spline, while the MVIS creates IMU states based on camera frequency (30 Hz). (2) Kalibr needs to form IMU cost terms based on IMU frequency (more than 100 Hz), while the MVIS leverages the pre-integration and all the IMU measurements between two camera timestamps are used together, leading to a significant smaller number of IMU factors in the MVIS. Note that we have also implemented both numerical and analytical Jacobians for the proposed MVIS. The numerical and analytical Jacobians can achieve similar accuracy results but with 10%–14% running time saving when using analytical Jacobians.

## *11.2. Four IMUs + two GS cameras + two RS cameras*

All the four IMUs, GS stereo camera from RealSense T265 and ELP RS stereo camera are used in this evaluation. Both GS and RS cameras are used in this experiment to show that our proposed MVIS supports full-parameter joint calibration with GS and RS cameras, while Kalibr does not support joint calibration of IMU and RS cameras, nor GS and RS cameras. Total 15 datasets were collected with an AprilTag board, on which both the proposed MVIS and the Kalibr calibration toolbox were run to report calibration accuracy and repeatability statistics. During data collection,

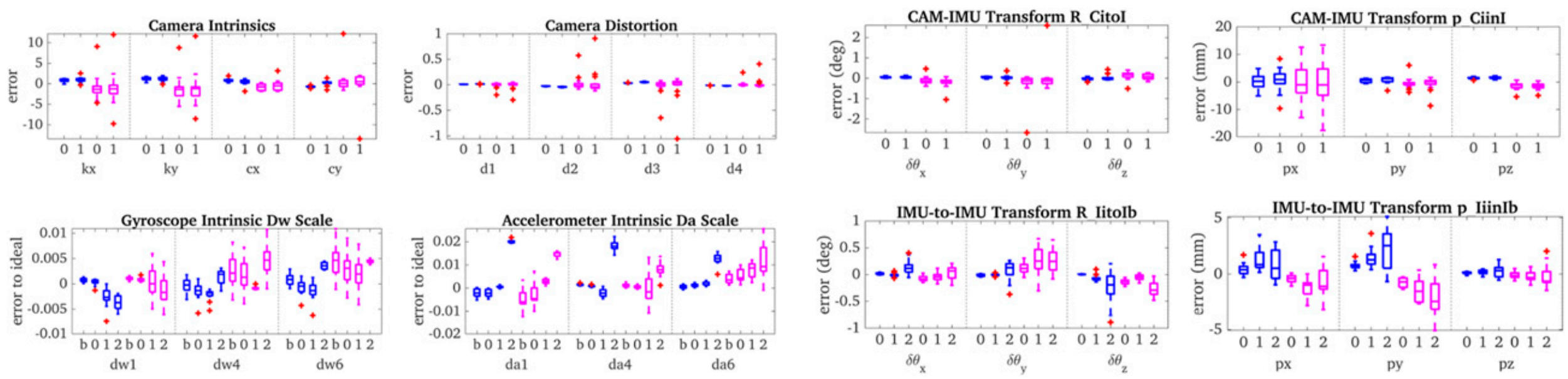

**Figure 10.** Calibration results over 15 different datasets collected with Intel Realsense T265 (GS), ELP-960P2CAM-V90-VC USB 2.0 (RS, 640x480) and four IMUs. The proposed MVIS (blue, using all the sensors) and Kalibr baseline (magenta, using only T265 cameras with all the IMUs) statistics are reported. The top $x$-axis denote the two global shutter cameras (`CAM0`, `CAM1`) as `0` and `1`, respectively; The bottom $x$-axis denotes the base IMU (`IMUb`) as `b`, and auxiliary IMUs (`IMUa0`, `IMUa1` and `IMUa2`) as `0`, `1`, and `2`, respectively. For CAM-IMU transformation {R_C`i`toI, p_C`i`inI}, `i=0, 1`. For IMU-IMU transformation {R_I`i`toIb, p_I`i`inIb}, `i=0, 1, 2`.

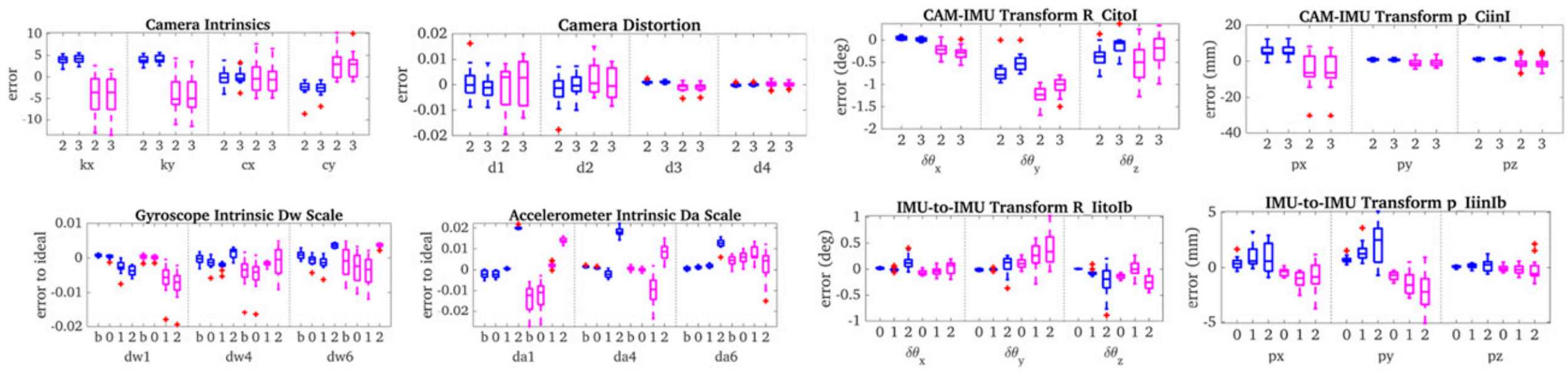

**Figure 11.** Calibration results over 15 different datasets collected with Intel Realsense T265 (GS), ELP-960P2CAM-V90-VC USB 2.0 (RS, 640x480) and four IMUs. The proposed MVIS (blue, using all the sensors) and Kalibr rolling shutter baseline (magenta, using only the RS cameras with all IMUs) statistics are reported. The top $x$-axis denote the two rolling shutter cameras (`CAM2`, `CAM3`) as `2` and `3`, respectively; The bottom $x$-axis denotes the base IMU (`IMUb`) as `b`, and auxiliary IMUs (`IMUa0`, `IMUa1`, and `IMUa2`) as `0`, `1`, and `2`. For CAM-IMU transformation {R_C`i`toI, p_C`i`inI}, `i=2, 3`. For IMU-IMU transformation {R_I`i`toIb, p_I`i`inIb}, `i=0, 1, 2`.

all 6-axis motion of the VI-Rig was excited to avoid degenerate motions for calibration parameters.

*11.2.1. Calibration for IMU and GS/RS.* During evaluation, all the GS/RS cameras and IMUs are used for the proposed MVIS. Since Kalibr does not support hybrid calibration of GS and RS cameras, we first run Kalibr with all four IMUs and only GS stereo camera from RealSense T265 (`CAM0` and `CAM1`). The results with boxplots are presented in Figure 10. Then, we run Kalibr with all four IMUs and ELP RS stereo camera (`CAM2` and `CAM3`) using a Kalibr extension (Huai et al., 2022). The results are presented in Figure 11. Note that in the evaluations, the left&right cameras from the stereo of RealSense T265 are denoted as `CAM0` and `CAM1`, while the left&right cameras from ELP RS stereo are denoted as `CAM2` and `CAM3`. In this experiment, we did camera calibration for each collected dataset with Kalibr. Therefore, we can have the statistics for the camera intrinsic estimates in Figures 10 and 11, from which, we can see that the mean estimates of both the IMU and camera related parameters are similar for both the proposed MVIS and Kalibr.

The boxplot ranges of camera and IMU related parameters from the proposed MVIS are much smaller than those of the Kalibr, which shows that MVIS is able to achieve much better estimation convergence and repeatability than Kalibr, especially for the case of using ELP RS cameras. This result verifies that the proposed MVIS can handle the joint calibration of IMU-GS/RS cameras, which is missing from Kalibr. In this experiment, MVIS used both GS/RS cameras while the Kalibr is evaluated on only GS or only RS cameras. Hence, this experiment further proves that the joint calibration of multiple sensors (i.e. cameras) does improve the calibration accuracy and repeatability.

It is interesting to see that the IMU calibration results (the scale terms for gyroscope correction $\mathbf{D}_w$ and accelerometer correction $\mathbf{D}_a$) between these two experiments (Figure 9 in Section 11.1 and Figure 10 or Figure 11 in Section 11.2) are very similar. This further validates the stability of the proposed MVIS.

*11.2.2. Evaluation of multiple gyroscopes calibration.* We further evaluate the proposed MVIS with multiple auxiliary gyroscopes. With the same 15 datasets, all cameras, base IMU (GX-25 `IMUb`) and the gyroscopes of three auxiliary IMUs (GX-35 `IMUa0`, Xsens `IMUa1`, and T265 IMU `IMUa2`) are used for evaluation with MVIS. The calibration results of MVIS with these auxiliary gyroscopes (in black), compared to the MVIS (in blue) and Kalibr (in magenta) with full auxiliary

IMUs, are shown in Figure 13. Note that the auxiliary gyroscope does not contain ${}^{I}_{w}\mathbf{R}$. Hence, we set ${}^{I}_{w}\mathbf{R} = \mathbf{I}_3$ as default.

It is clear from the results that MVIS with multiple auxiliary gyroscopes still can achieve almost the same estimates for gyroscope scales and time offsets as MVIS and Kalibr with full auxiliary IMUs (Figures 12 and 13).

At the same time, we can also see that the calibration of rotation between auxiliary gyroscope and base IMU (the second row of Figure 13) from MVIS with multiple gyroscopes is slightly worse than that of MVIS and Kalibr with full IMUs, especially for `IMUa2`. This might be due to the fact that accelerometer measurement can benefit the extrinsic calibration between IMUs. We also want to point out that the rotation calibration difference is smaller than 0.5°, which is not significant.

*11.2.3. Temporal calibration.* The temporal calibration, including time offsets and rolling shutter readout time, are presented in Figure 12, which shows the results using base IMU and all the full auxiliary IMUs. The `0`, `1`, `2`, and `3` from the `CAM Toff` and `Readout` refers to `CAM0`-`CAM3`. From Figure 12, it is clear that the time offset calibration is almost the same for the proposed MVIS and the Kalibr. The readout time calibration errors are all within 2 ms.

From the results, we can also find out that the triggering time offset of the RealSence T265 is not stable. As can be seen from the left of Figure 12, there are outliers as large as 30 ms between the base IMU `IMUb` and the right camera of T265. Similarly, the time offset of the IMU from T265 (`IMUa2`) to base IMU (`IMUb`) is also slightly unstable from the estimates of MVIS and Kalibr, as outliers (red crosses) in right of

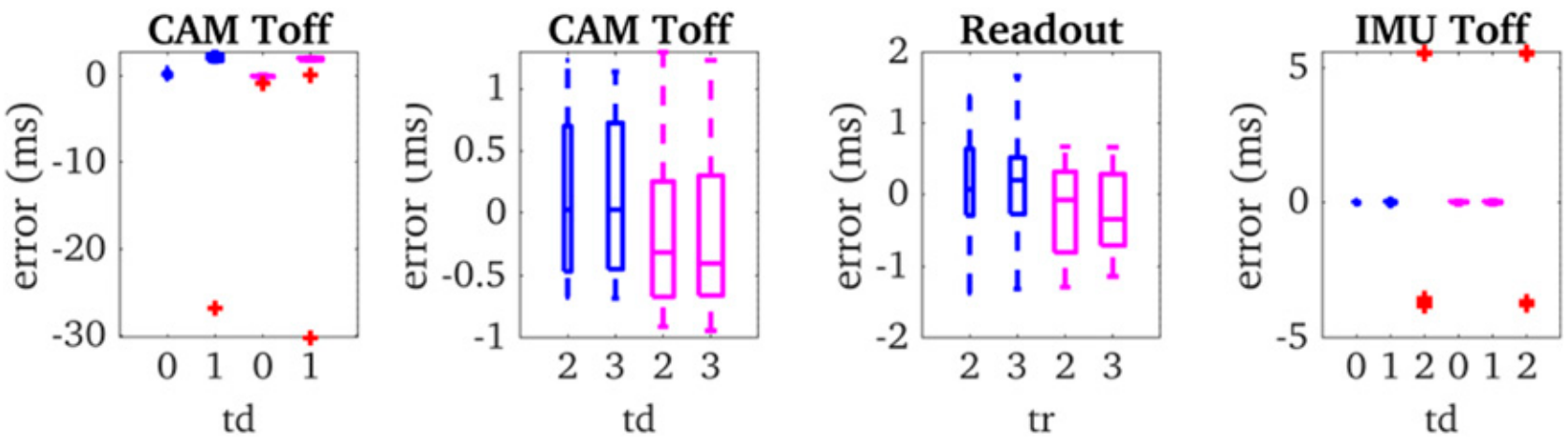

**Figure 12.** Temporal calibration results over 15 different datasets with Intel Realsense T265 (GS), ELP-960P2CAM-V90-VC USB 2.0 (RS, 640x480) and four IMUs. The proposed MVIS (blue) and Kalibr rolling shutter baseline (magenta) statistics are reported. The $x$-axis of the left three figures denotes the two global shutter camera `CAM0`, `CAM1`, two rolling shutter camera `CAM2`, `CAM3`. The $x$-axis of the right figure denotes the time offsets between the base IMU (`IMUb`), and auxiliary IMUs (`IMUa0`, `IMUa1`, and `IMUa2`).

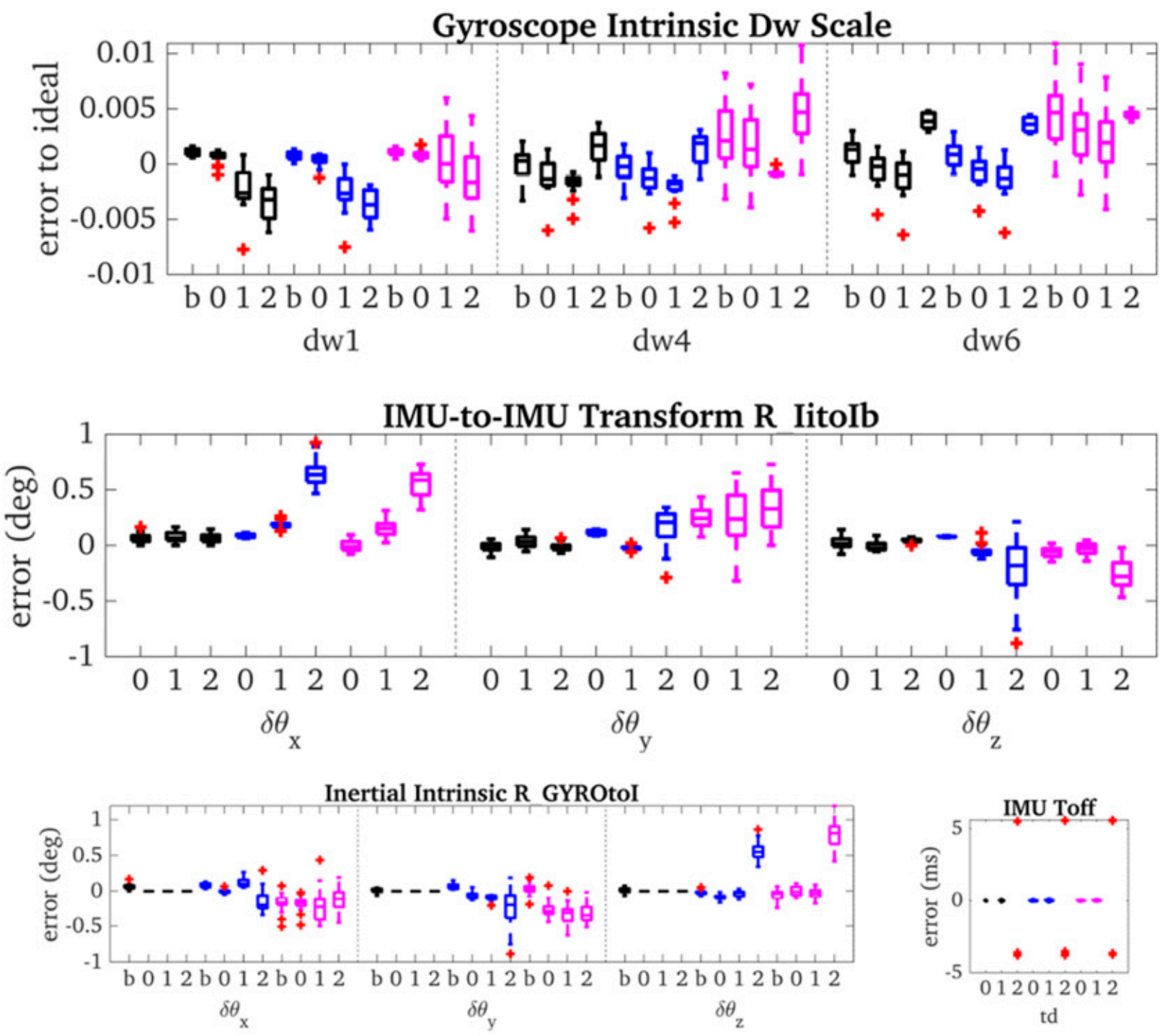

**Figure 13.** Calibration results over 15 different datasets with Intel Realsense T265 and four IMUs. The proposed MVIS with a base IMU and only gyroscopes of the 3 auxiliary IMUs (black), MVIS with a base IMU and 3 auxiliary IMUs (blue) and Kalibr baseline (magenta) statistics are reported. The $x$-axis denotes the base IMU (`IMUb`) and auxiliary IMUs (`IMUa0`, `IMUa1`, and `IMUa2`) for all algorithms. For IMU-IMU transformation {R_IitoIb}, `i=0, 1, 2`.

Figure 12 are obvious. Note that the 0, 1, and 2 in the plot of IMU Toff from right of Figure 12 denote the time offsets of IMUa0, IMUa1, and IMUa2 to IMUb, respectively.

As shown in the right of Figure 12, the time offset between based IMU IMUb and the auxiliary IMU IMUa2 (BMI055 from T265) also has 10 ms offsets (from near-5 ms to 5 ms). This is probably due to the build-in drivers of this relatively low-cost sensor (T265). This figure shows that the estimate results from Kalibr (magenta) and the proposed MVIS (blue) can identify the temporal calibration problems of T265, which validates that the proposed MVIS can be used to identify the temporal instability of T265 and provide reliable calibration results.

## *11.3. Planar motion with 4 IMUs + 2 GS cameras*

We further verify the degenerate motions with a dataset collected under planar motion. All four IMUs and the GS stereo camera from RealSense T265 are used for data collection. When collecting data, the VI-Rig is put on a chair with wheels and moved around the room in planar motion. The proposed MVIS is run on this dataset 4 times with different perturbations to the initial values of IMU-IMU translations.

Under planar motion, the rotation axis, roughly along the local $z$-axis for the base IMU, is fixed for the VI-Rig. Hence, the IMU-IMU translation along the rotation axis and the $d_{w1}$, $d_{w2}$, $d_{w3}$ from gyroscope correction $\mathbf{D}_w$ should be unobservable. The calibration results for these parameters can be clearly seen in Figure 14 and they diverge erroneously during optimization.

As a comparison, we use the same sensor rig and same perturbations to IMU-IMU translation to run the proposed MVIS under fully excited motions. As shown in Figure 15, all these calibration parameters can converge well when fully excited motions are given, as compared to Figure 14.

## *11.4. Discussion on estimation convergence*

We formulate the MVIS calibration and estimation as a nonlinear least squares (NLS) problem, which is a non-convex optimization problem and its global minimum is hard to guarantee. Also, as it is almost impossible to obtain the “true” calibration for real sensor rigs, evaluating the global optimum in real world becomes formidable. As such, we often use engineering intuitions to improve the calibration in terms of accuracy, convergence and repeatability, e.g., by fully exciting sensor motions, improving calibration priors and delaying adding auxiliary IMU factors. Although there is no theoretical guarantee, our simulation results have shown that the proposed MVIS calibration is able to converge to the true values.

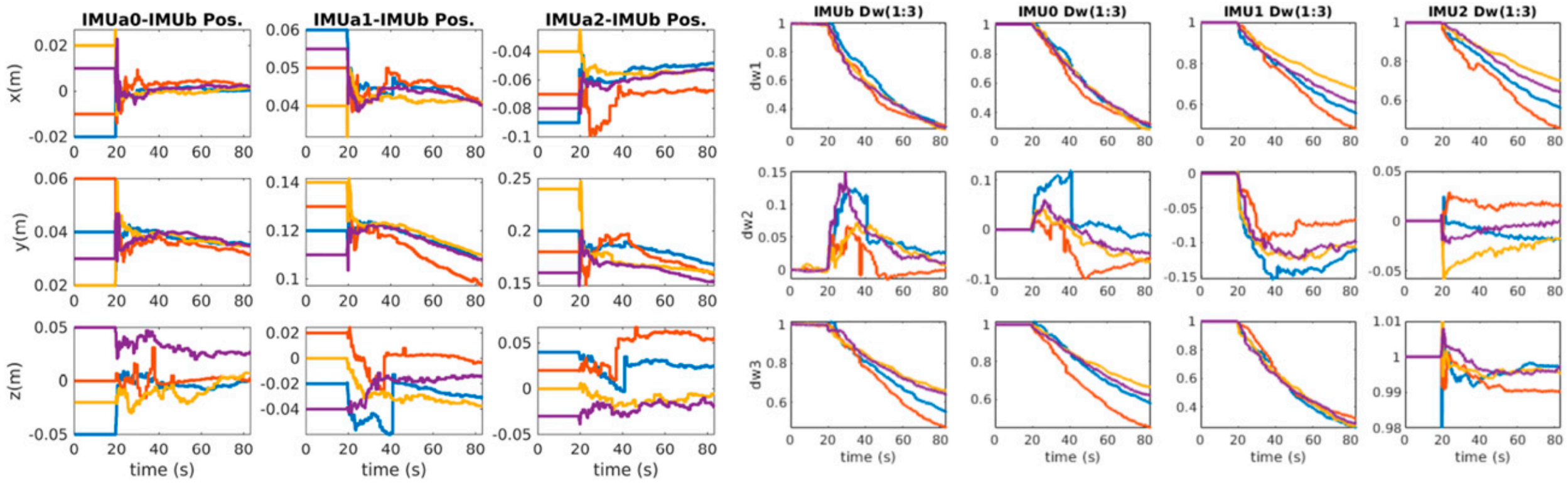

**Figure 14.** Calibration results for one planar motion dataset collected with Intel Realsense T265 and four IMUs. The translation of IMUb-IMUa0, IMUb-IMUa1, and IMUb-IMUa2, the gyroscope correction $\mathbf{D}_w$ of IMUb, IMUa0, IMUa1, and IMUa2 cannot converge under planar motions, which verifies our observability analysis. Different colors represent different initial perturbations to the IMU-IMU translations.

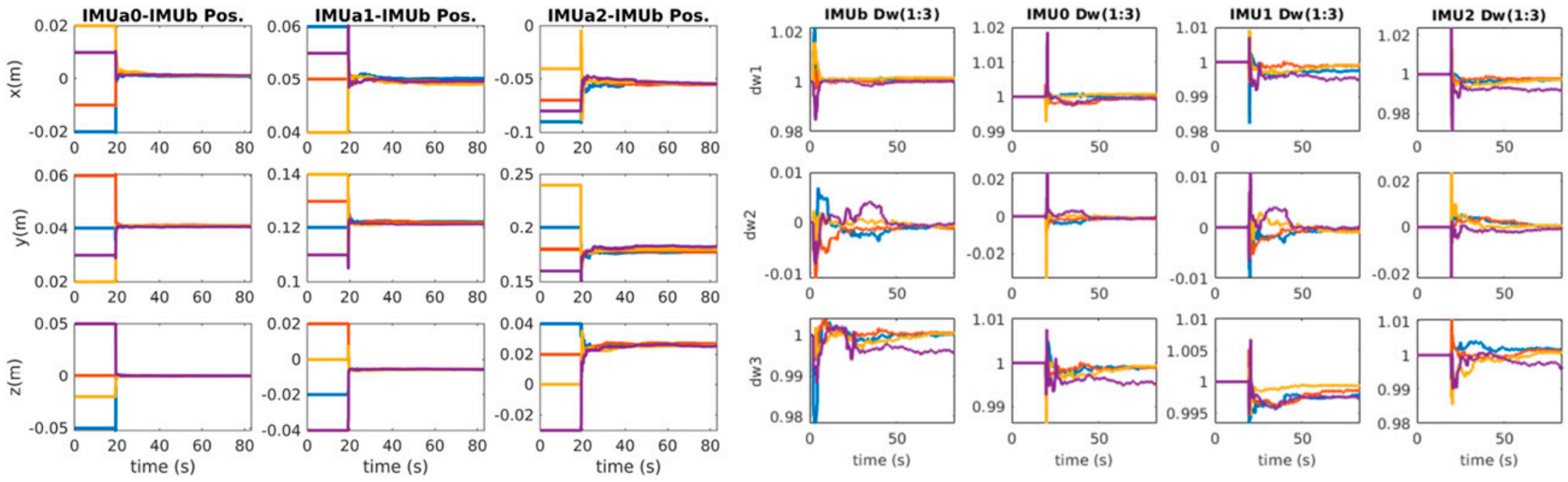

**Figure 15.** Calibration results for one fully-exicted motion dataset collected with Intel Realsense T265 and four IMUs. The translation of IMUb-IMUa0, IMUb-IMUa1, and IMUb-IMUa2, the gyroscope correction $\mathbf{D}_w$ of IMUb, IMUa0, IMUa1, and IMUa2 converge nicely compared to planar motion case. Different colors represent different initial perturbations to the IMU-IMU translations.

Based on our analysis, we need fully excited motions (3D rotation and 3D translation) for all the sensors to make sure all the related calibration parameters can converge (see Sections 10.1, 11.1, and 11.2). If the MVIS undergoes any degenerate motions listed in Section 9, some calibration parameters are unlikely to converge (see Sections 10.2, 10.3, and 11.3). From the extensive simulations and real world experiments, we find that the proposed MVIS estimation with full-parameter calibration can converge in most cases.

Given fully excited motions, the initial guess and prior information for these calibration parameters are also crucial for estimator convergence. As discussed earlier, the IMU intrinsics are in most cases not large in values, and hence initialized with “ideal” intrinsic values: identity matrices for gyroscope correction $\mathbf{D}_w$, accelerometer correction $\mathbf{D}_a$, ${}^{I}_{w}\mathbf{R}$ and zero matrix for g-sensitivity $\mathbf{T}_g$. Although the camera intrinsic and distortion parameters are usually initialized based on the camera calibration using OpenCV (OpenCV Developers Team, 2021) or Kalibr (Furgale et al., 2013), the proposed MVIS can handle inaccurate camera intrinsics as shown in Figure 9. For the IMU-IMU/camera extrinsics, the initial orientation part is decided manually while the translation part can be measured by hand. This can be improved by using trajectory alignment of visual trajectories and IMU integrated trajectory segments. The temporal related parameters are most cases initialized through orientation alignment. Through real-world experiments, see first row of Figure 9 and `IMU-IMU Pos.` in Figure 15, we find that the proposed MVIS estimation with full-parameter calibration can converge even with various perturbations to initial guesses.

The more sensors used, the more calibration parameters will be included in the state, resulting in larger NLS problems. This would potentially pose challenges to convergence of the proposed MVIS when estimating all the related calibrations at once, especially when the initial guesses for these calibrations are not of good quality. To address this issue, we add the cost terms from auxiliary sensors later than the base inertial sensor costs, after base inertial sensor related parameters converge. In our experiments, we first only optimize the base IMU and the cameras related costs until the landmark feature estimates converge. After that, the auxiliary IMUs/gyroscopes cost terms will be added to the NLS for solving their related calibration parameters. In effect, from our experiences on the data collected using VI-Rig, 5-20s of the data with fully excited motions are sufficient for the landmarks to converge.

## 12. Conclusions and future work

In this paper, we have developed a multi-visual-inertial system (MVIS) estimation algorithm which can fuse multiple IMUs, gyroscopes and GS/RS cameras, with a special focus on full-calibration of all intrinsics, extrinsics, and temporal parameters (including time offsets and readout times for RS cameras). In particular, we proposed ACI$^3$, a novel IMU preintegration which incorporates IMU intrinsic parameters. Based on ACI$^3$, we fuse multiple IMU measurements by leveraging IMU-IMU rigid body constraints with spatio-temporal and inertial intrinsic calibration. We have performed MVIS observability analysis, proving that four standard unobservable directions corresponding to global yaw and global translation remain, while the calibration parameters are observable under fully excited motion. Moreover, we have also, for the first time, identified the commonly seen degenerate motions that can cause IMU-IMU calibration parameters to become unobservable. We show that the rotation calibration between IMU and gyroscopes is unobservable given one-axis rotation, while the time offset between IMUs is observable given non-zero constant local angular and linear velocity for one of the IMUs. Extensive simulations have been performed to evaluate the proposed system and verify the degenerate motions identified for these calibration parameters. Moreover, a self-made sensor rig that consists of multiple commonly-used IMUs and GS/RS cameras were used for data collection and system evaluation. In particular, three sets of experiments were performed to fully evaluate the calibration accuracy of the proposed MVIS against the state-of-art sensor calibration framework Kalibr. A total of 25 datasets were collected with the VI-Rigs to provide detailed statistics for calibration convergence and repeatability of the proposed MVIS and Kalibr.

In the future, we will investigate vehicle dynamics (e.g., wheel odometry on ground vehicle (Lee et al., 2020), contact dynamics from legged robots (Fourmy et al., 2021), or MAV dynamics (Chen et al., 2022)) in MVIS calibration. Calibrating MVIS under degenerate motions is of particular interest along with how sensor configurations/installation affect calibration performance. We will also develop efficient marginalization to enable the proposed batch optimization-based MVIS to perform online calibration amenable for real-time performance.

### Declaration of conflicting interests

The author(s) declared no potential conflicts of interest with respect to the research, authorship, and/or publication of this article.

### Funding

The author(s) disclosed receipt of the following financial support for the research, authorship, and/or publication of this article:

This work was partially supported by the University of Delaware (UD) College of Engineering, the NSF (IIS-1924897), and Google ARCore. Yang was partially supported by the University Doctoral Fellowship and Geneva was supported by the NASA DE Space Grant Graduate Fellowship. In addition, we are grateful for the insightful and inspiring comments from anonymous reviewers.

## ORCID iDs

Yulin Yang https://orcid.org/0000-0002-9675-9147
Patrick Geneva https://orcid.org/0000-0002-2179-3447
Guoquan Huang https://orcid.org/0000-0001-9932-0685

## Notes

1. https://github.com/ethz-asl/kalibr/wiki/Multi-IMU-and-IMU-intrinsic-calibration
2. More details about the project are at: https://openmvis.com/.

# Appendix

## A. IMU readings

To simplify the derivations, we have the following:

$$\begin{aligned}\boldsymbol{\omega}_{k+i} &= {}^I_w\mathbf{R}\mathbf{D}_w\left({}^w\boldsymbol{\omega}_{m_{k+i}} - \mathbf{T}_g\mathbf{a}_{k+i} - \mathbf{b}_{g_{k+i}} - \mathbf{n}_{g_{k+i}}\right)\\ &= {}^I_w\mathbf{R}\mathbf{D}_w\left({}^w\boldsymbol{\omega}_{m_{k+i}} - \mathbf{b}_{g_{k+i}} - \mathbf{n}_{g_{k+i}} - \mathbf{T}_g\mathbf{a}_{k+i}\right)\\ &= {}^I_w\mathbf{R}\mathbf{D}_w\left({}^w\boldsymbol{\omega}_{m_{k+i}} - \Delta\mathbf{b}_{gi} - \mathbf{b}_{g_k} - \mathbf{n}_{g_{k+i}} - \mathbf{T}_g\mathbf{a}_{k+i}\right)\\ \mathbf{a}_{k+i} &= {}^I_a\mathbf{R}\mathbf{D}_a\left({}^a\mathbf{a}_{m_{k+i}} - \mathbf{b}_{a_{k+i}} - \mathbf{n}_{a_{k+i}}\right)\\ &= {}^I_a\mathbf{R}\mathbf{D}_a\left({}^a\mathbf{a}_{m_{k+i}} - \Delta\mathbf{b}_{ai} - \mathbf{b}_{a_k} - \mathbf{n}_{a_{k+i}}\right)\end{aligned}$$

Note that the $\boldsymbol{\omega}_{k+i}$ and $\mathbf{a}_{k+i}$ are actually function of constant IMU intrinsics $\mathbf{x}_{in}$ and biases $\mathbf{x}_{b_k}$. The angular velocity and linear acceleration estimates can be written as:

$$\begin{aligned}\widehat{\boldsymbol{\omega}}_{k+i} &= {}^I_w\widehat{\mathbf{R}}\widehat{\mathbf{D}}_w\left({}^w\boldsymbol{\omega}_{m_{k+i}} - \Delta\widehat{\mathbf{b}}_{gi} - \widehat{\mathbf{b}}_{g_k} - \widehat{\mathbf{T}}_g\widehat{\mathbf{a}}_{k+i}\right)\\ \widehat{\mathbf{a}}_{k+i} &= {}^I_a\widehat{\mathbf{R}}\widehat{\mathbf{D}}_a\left({}^a\mathbf{a}_{m_{k+i}} - \Delta\widehat{\mathbf{b}}_{ai} - \widehat{\mathbf{b}}_{a_k}\right)\\ &\triangleq \begin{bmatrix}{}^Ia_1 & {}^Ia_2 & {}^Ia_3\end{bmatrix}^\top\end{aligned}$$

For simplicity of derivations, we also define:

$$ {}^w\widehat{\boldsymbol{\omega}} = {}^w\boldsymbol{\omega}_{m_{k+i}} - \Delta\widehat{\mathbf{b}}_{gi} - \widehat{\mathbf{b}}_{g_k} - \widehat{\mathbf{T}}_g\widehat{\mathbf{a}}_{k+i} \tag{154}$$

$$\triangleq \begin{bmatrix}{}^w\omega_1 & {}^w\omega_2 & {}^w\omega_3\end{bmatrix}^\top \tag{155}$$

$$ {}^a\widehat{\mathbf{a}} = {}^a\mathbf{a}_{m_{k+i}} - \Delta\widehat{\mathbf{b}}_{ai} - \widehat{\mathbf{b}}_{a_k} \tag{156}$$

$$\triangleq \begin{bmatrix}{}^aa_1 & {}^aa_2 & {}^aa_3\end{bmatrix}^\top \tag{157}$$

Let's first define the error states for the $\boldsymbol{\omega}_{k+i}$ and $\mathbf{a}_{k+i}$ as:

$$\begin{aligned}\tilde{\mathbf{a}}_{k+i} &= {}^I_a\mathbf{R}\mathbf{H}_{D_a}\tilde{\mathbf{x}}_a - \lfloor\widehat{\mathbf{a}}_{k+i}\rfloor{}^I_a\mathbf{R}\delta\boldsymbol{\theta}_{I_a}\\ &\quad - {}^I_a\mathbf{R}\mathbf{D}_a\left(\tilde{\mathbf{b}}_{a_k} + \Delta\tilde{\mathbf{b}}_{ai} + \mathbf{n}_{a_{k+i}}\right)\\ \tilde{\boldsymbol{\omega}}_{k+i} &= {}^I_w\widehat{\mathbf{R}}\mathbf{H}_{D_w}\tilde{\mathbf{x}}_w - {}^I_w\mathbf{R}\mathbf{D}_w\mathbf{T}_g{}^I_a\mathbf{R}\mathbf{H}_{D_a}\tilde{\mathbf{x}}_a\\ &\quad - \lfloor\widehat{\boldsymbol{\omega}}_{k+i}\rfloor{}^I_w\mathbf{R}\delta\boldsymbol{\theta}_{I_w} + {}^I_w\mathbf{R}\mathbf{D}_w\mathbf{T}_g\lfloor\widehat{\mathbf{a}}_{k+i}\rfloor{}^I_a\mathbf{R}\delta\boldsymbol{\theta}_{I_a}\\ &\quad - {}^I_w\mathbf{R}\mathbf{D}_w\mathbf{H}_{T_g}\tilde{\mathbf{x}}_{T_g} - {}^I_w\mathbf{R}\mathbf{D}_w\\ &\quad \left(\tilde{\mathbf{b}}_{g_k} + \Delta\tilde{\mathbf{b}}_{gi} + \mathbf{n}_{g_{k+i}} - \mathbf{T}_g{}^I_a\mathbf{R}\mathbf{D}_a\left(\tilde{\mathbf{b}}_{a_k} + \Delta\tilde{\mathbf{b}}_{ai} + \mathbf{n}_{a_{k+i}}\right)\right)\end{aligned}$$

Hence, we can have:

$$\begin{bmatrix}\tilde{\boldsymbol{\omega}}_{k+i}\\ \tilde{\mathbf{a}}_{k+i}\end{bmatrix} \triangleq \mathbf{H}^{wa}_b\begin{bmatrix}\Delta\tilde{\mathbf{b}}_{gi}\\ \Delta\tilde{\mathbf{b}}_{ai}\end{bmatrix} + \mathbf{H}^{wa}_{b_k}\begin{bmatrix}\tilde{\mathbf{b}}_{g_k}\\ \tilde{\mathbf{b}}_{a_k}\end{bmatrix} + \mathbf{H}^{wa}_{in}\begin{bmatrix}\tilde{\mathbf{x}}_w\\ \tilde{\mathbf{x}}_a\\ \tilde{\mathbf{x}}_{T_g}\\ \delta\boldsymbol{\theta}_{I_w}\\ \delta\boldsymbol{\theta}_{I_a}\end{bmatrix} + \mathbf{H}^{wa}_n\begin{bmatrix}\mathbf{n}_{g_{k+i}}\\ \mathbf{n}_{a_{k+i}}\end{bmatrix} \tag{158}$$

with:

$$\mathbf{H}^{wa}_b = \begin{bmatrix}-{}^I_w\mathbf{R}\mathbf{D}_w & {}^I_w\mathbf{R}\mathbf{D}_w\mathbf{T}_g{}^I_a\mathbf{R}\mathbf{D}_a\\ \mathbf{0}_3 & -{}^I_a\mathbf{R}\mathbf{D}_a\end{bmatrix} \tag{159}$$

$$\mathbf{H}^{wa}_{b_k} = \begin{bmatrix}-{}^I_w\mathbf{R}\mathbf{D}_w & {}^I_w\mathbf{R}\mathbf{D}_w\mathbf{T}_g{}^I_a\mathbf{R}\mathbf{D}_a\\ \mathbf{0}_3 & -{}^I_a\mathbf{R}\mathbf{D}_a\end{bmatrix} \tag{160}$$

$$\mathbf{H}^{wa}_n = \begin{bmatrix}-{}^I_w\mathbf{R}\mathbf{D}_w & {}^I_w\mathbf{R}\mathbf{D}_w\mathbf{T}_g{}^I_a\mathbf{R}\mathbf{D}_a\\ \mathbf{0}_3 & -{}^I_a\mathbf{R}\mathbf{D}_a\end{bmatrix} \tag{161}$$

$$\mathbf{H}^{wa}_{in} = \begin{bmatrix}\mathbf{H}_{xw} & \mathbf{H}_{xa} & \mathbf{H}_{gs} & \mathbf{H}_{I_w} & \mathbf{H}_{I_a}\end{bmatrix} \tag{162}$$

$$\mathbf{H}_{xw} = \begin{bmatrix}{}^I_w\mathbf{R}\mathbf{H}_{D_w}\\ \mathbf{0}_3\end{bmatrix} \tag{163}$$

$$\mathbf{H}_{xa} = \begin{bmatrix}-{}^I_w\mathbf{R}\mathbf{D}_w\mathbf{T}_g{}^I_a\mathbf{R}\mathbf{H}_{D_a}\\ {}^I_a\mathbf{R}\mathbf{H}_{D_a}\end{bmatrix} \tag{164}$$

$$\mathbf{H}_{gs} = \begin{bmatrix}-{}^I_w\mathbf{R}\mathbf{D}_w\mathbf{H}_{T_g}\\ \mathbf{0}_3\end{bmatrix} \tag{165}$$

Note that if the `RPNG` model is used, $\mathbf{H}_{D_w}$, $\mathbf{H}_{D_a}$, $\mathbf{H}_{T_g}$ and $\mathbf{H}_{I_a}$ are computed with:

$$\mathbf{H}_{D_w} = \begin{bmatrix}{}^w\omega_1\mathbf{e}_1 & {}^w\omega_2\mathbf{e}_1 & {}^w\omega_2\mathbf{e}_2 & {}^w\omega_3\mathbf{I}_3\end{bmatrix} \tag{166}$$

$$\mathbf{H}_{D_a} = \begin{bmatrix}{}^aa_1\mathbf{e}_1 & {}^aa_2\mathbf{e}_1 & {}^aa_2\mathbf{e}_2 & {}^aa_3\mathbf{I}_3\end{bmatrix} \tag{167}$$

$$\mathbf{H}_{T_g} = \begin{bmatrix}{}^Ia_1\mathbf{I}_3 & {}^Ia_2\mathbf{I}_3 & {}^Ia_3\mathbf{I}_3\end{bmatrix} \tag{168}$$

$$\mathbf{H}_{I_a} = \begin{bmatrix}{}^I_w\mathbf{R}\mathbf{D}_w\mathbf{T}_g\lfloor\widehat{\mathbf{a}}_{k+i}\rfloor{}^I_a\mathbf{R}\\ -\lfloor\widehat{\mathbf{a}}_{k+i}\rfloor{}^I_a\mathbf{R}\end{bmatrix} \tag{169}$$

If the `Kalibr` model is used, $\mathbf{H}_{D_w}$, $\mathbf{H}_{D_a}$, $\mathbf{H}_{T_g}$ and $\mathbf{H}_{I_w}$ are computed with:

$$\mathbf{H}_{D_w} = \begin{bmatrix}{}^w\omega_1\mathbf{I}_3 & {}^w\omega_2\mathbf{e}_2 & {}^w\omega_2\mathbf{e}_3 & {}^w\omega_3\mathbf{e}_3\end{bmatrix} \tag{170}$$

$$\mathbf{H}_{D_a} = \begin{bmatrix}{}^aa_1\mathbf{I}_3 & {}^aa_2\mathbf{e}_2 & {}^aa_2\mathbf{e}_3 & {}^aa_3\mathbf{e}_3\end{bmatrix} \tag{171}$$

$$\mathbf{H}_{T_g} = \begin{bmatrix}{}^Ia_1\mathbf{I}_3 & {}^Ia_2\mathbf{I}_3 & {}^Ia_3\mathbf{I}_3\end{bmatrix} \tag{172}$$

$$\mathbf{H}_{I_w} = \begin{bmatrix}-\lfloor\widehat{\boldsymbol{\omega}}_{k+i}\rfloor{}^I_w\mathbf{R}\\ \mathbf{0}_3\end{bmatrix} \tag{173}$$

## B. Linearization prediction

$\mathbf{R}_{i,i+1}$ from equation (56), can be written as:

$$\mathbf{R}_{i,i+1} \simeq \exp(\boldsymbol{\omega}_{k+i}\delta t_i) \tag{174}$$

$$= \exp((\widehat{\boldsymbol{\omega}}_{k+i} + \tilde{\boldsymbol{\omega}}_{k+i})\delta t_i) \tag{175}$$

$$= \exp(\widehat{\boldsymbol{\theta}}_{i,i+1})\exp(\mathbf{J}_r(\widehat{\boldsymbol{\theta}}_{i,i+1})\tilde{\boldsymbol{\omega}}_{k+i}\delta t_i) \tag{176}$$

$\mathbf{p}_{i,i+1}$ from equation (57) can be written as:

$$
\begin{aligned}
\mathbf{p}_{i,i+1} &= \int_{t_{k+i}}^{t_{k+i+1}} \int_{t_{k+i}}^{s} {}^{I_{k+i}}_{I_\tau}\mathbf{R}^{I_\tau}\mathbf{a} d\tau ds \\
&\simeq \int_{t_{k+i}}^{t_{k+i+1}} \int_{t_{k+i}}^{s} \exp(\boldsymbol{\omega}_{k+i}\delta\tau)\mathbf{a}_{k+i} d\tau ds \\
&= \int_{t_{k+i}}^{t_{k+i+1}} \int_{t_{k+i}}^{s} \exp((\widehat{\boldsymbol{\omega}}_{k+i}+\tilde{\boldsymbol{\omega}}_{k+i})\delta\tau)(\widehat{\mathbf{a}}_{k+i}+\tilde{\mathbf{a}}_{k+i}) d\tau ds \\
&\simeq \underbrace{\int_{t_{k+i}}^{t_{k+i+1}} \int_{t_{k+i}}^{s} \exp(\widehat{\boldsymbol{\omega}}_{k+i}\delta\tau)\widehat{\mathbf{a}}_{k+i} d\tau ds}_{\widehat{\mathbf{p}}_{i,i+1}} \\
&\quad - \underbrace{\int_{t_{k+i}}^{t_{k+i+1}} \int_{t_{k+i}}^{s} \exp(\widehat{\boldsymbol{\omega}}_{k+i}\delta\tau)\lfloor\widehat{\mathbf{a}}_{k+i}\rfloor\mathbf{J}_r(\boldsymbol{\omega}_{k+i}\delta\tau)\delta\tau d\tau ds}_{\boldsymbol{\Xi}_4}\, \tilde{\boldsymbol{\omega}}_{k+i} \\
&\quad + \underbrace{\int_{t_{k+i}}^{t_{k+i+1}} \int_{t_{k+i}}^{s} \exp(\widehat{\boldsymbol{\omega}}_{k+i}\delta\tau) d\tau ds}_{\boldsymbol{\Xi}_2}\, \tilde{\mathbf{a}}_{k+i}
\end{aligned}
$$

$\mathbf{v}_{i,i+1}$ from equation (58) can be written as:

$$
\begin{aligned}
\mathbf{v}_{i,i+1} &= \int_{t_{k+i}}^{t_{k+i+1}} {}^{I_{k+i}}_{I_\tau}\mathbf{R}^{I_\tau}\mathbf{a} d\tau \\
&\simeq \int_{t_{k+i}}^{t_{k+i+1}} \exp(\boldsymbol{\omega}_{k+i}\delta\tau)\mathbf{a}_{k+i} d\tau \\
&= \int_{t_{k+i}}^{t_{k+i+1}} \exp((\widehat{\boldsymbol{\omega}}_{k+i}+\tilde{\boldsymbol{\omega}}_{k+i})\delta\tau)(\widehat{\mathbf{a}}_{k+i}+\tilde{\mathbf{a}}_{k+i}) d\tau \\
&\simeq \underbrace{\int_{t_{k+i}}^{t_{k+i+1}} \exp(\widehat{\boldsymbol{\omega}}_{k+i}\delta\tau)\widehat{\mathbf{a}}_{k+i} d\tau}_{\widehat{\mathbf{v}}_{i,i+1}} \\
&\quad - \underbrace{\int_{t_{k+i}}^{t_{k+i+1}} \exp(\widehat{\boldsymbol{\omega}}_{k+i}\delta\tau)\lfloor\widehat{\mathbf{a}}_{k+i}\rfloor\mathbf{J}_r(\widehat{\boldsymbol{\omega}}_{k+i}\delta\tau)\delta\tau d\tau}_{\boldsymbol{\Xi}_3}\, \tilde{\boldsymbol{\omega}}_{k+i} \\
&\quad + \underbrace{\int_{t_{k+i}}^{t_{k+i+1}} \exp(\widehat{\boldsymbol{\omega}}_{k+i}\delta\tau) d\tau}_{\boldsymbol{\Xi}_1}\, \tilde{\mathbf{a}}_{k+i}
\end{aligned}
$$

The block Jacobians for equation (95) are written as:

$$
\boldsymbol{\Phi}_{i+1,i} = \begin{bmatrix} \boldsymbol{\Phi}_{nn} & \mathbf{H}_{wa}\mathbf{H}_b^{wa} \\ \mathbf{0}_{6\times 9} & \mathbf{I}_6 \end{bmatrix} \tag{177}
$$

$$
\boldsymbol{\Phi}_{nn} = \begin{bmatrix} \widehat{\mathbf{R}}_{i,i+1}^{\top} & \mathbf{0}_3 & \mathbf{0}_3 \\ -\Delta\widehat{\mathbf{R}}_i\left\lfloor\widehat{\mathbf{p}}_{i,i+1}\right\rfloor & \mathbf{I}_3 & \mathbf{I}_3\delta t_i \\ -\Delta\widehat{\mathbf{R}}_i\left\lfloor\widehat{\mathbf{v}}_{i,i+1}\right\rfloor & \mathbf{0}_3 & \mathbf{I}_3 \end{bmatrix} \tag{178}
$$

$$
\mathbf{H}_{wa} = \begin{bmatrix} \mathbf{J}_r(\boldsymbol{\theta}_{i,i+1})\delta t_i & \mathbf{0}_3 \\ -\Delta\widehat{\mathbf{R}}_i\boldsymbol{\Xi}_4 & \Delta\widehat{\mathbf{R}}_i\boldsymbol{\Xi}_2 \\ -\Delta\widehat{\mathbf{R}}_i\boldsymbol{\Xi}_3 & \Delta\widehat{\mathbf{R}}_i\boldsymbol{\Xi}_1 \end{bmatrix} \tag{179}
$$

$$
\boldsymbol{\Phi}_b = \begin{bmatrix} \mathbf{H}_{wa}\mathbf{H}_{b_k}^{wa} \\ \mathbf{0} \end{bmatrix}, \boldsymbol{\Phi}_{in} = \begin{bmatrix} \mathbf{H}_{wa}\mathbf{H}_{in}^{wa} \\ \mathbf{0} \end{bmatrix} \tag{180}
$$

$$
\mathrm{G}_i = \begin{bmatrix} \mathbf{H}_{wa}\mathbf{H}_n^{wa} & \mathbf{0}_{9\times 6} \\ \mathbf{0}_{6\times 6} & \mathbf{I}_6\delta t_i \end{bmatrix} \tag{181}
$$

## C. Bias and intrinsic Jacobians

The biases or IMU intrinsic Jacobians can be recursively computed as:

$$
\frac{\partial \begin{bmatrix} \delta\Delta\boldsymbol{\theta}_{i+1} \\ \Delta\tilde{\mathbf{p}}_{i+1} \\ \Delta\tilde{\mathbf{v}}_{i+1} \end{bmatrix}}{\partial\tilde{\mathbf{x}}_*} = \boldsymbol{\Phi}_{nn}\frac{\partial \begin{bmatrix} \delta\Delta\boldsymbol{\theta}_{i} \\ \Delta\tilde{\mathbf{p}}_{i} \\ \Delta\tilde{\mathbf{v}}_{i} \end{bmatrix}}{\partial\tilde{\mathbf{x}}_*} + \mathbf{H}_{wa}\mathbf{H}_*^{wa} \tag{182}
$$

where * denotes $b_k$ or $in$. Then, these Jacobians can be rewritten as:

$$
\begin{aligned}
\frac{\partial\delta\Delta\boldsymbol{\theta}_{i+1}}{\partial\tilde{\mathbf{x}}_*} &= \widehat{\mathbf{R}}_{i,i+1}^{\top}\frac{\partial\delta\Delta\boldsymbol{\theta}_i}{\partial\tilde{\mathbf{x}}_*} + [\mathbf{J}_r(\widehat{\boldsymbol{\theta}}_{i,i+1})\delta t_i \;\; \mathbf{0}_3]\mathbf{H}_*^{wa} \\
\frac{\partial\Delta\tilde{\mathbf{p}}_{i+1}}{\partial\tilde{\mathbf{x}}_*} &= -\Delta\widehat{\mathbf{R}}_i\lfloor\widehat{\mathbf{p}}_{i,i+1}\rfloor\frac{\partial\delta\Delta\boldsymbol{\theta}_i}{\partial\tilde{\mathbf{x}}_*} + \frac{\partial\Delta\tilde{\mathbf{p}}_i}{\partial\tilde{\mathbf{x}}_*} + \frac{\partial\Delta\tilde{\mathbf{v}}_i}{\partial\tilde{\mathbf{x}}_*}\delta t_i \\
&\quad + \Delta\widehat{\mathbf{R}}_i[-\boldsymbol{\Xi}_4 \;\; \boldsymbol{\Xi}_2]\mathbf{H}_*^{wa} \\
\frac{\partial\Delta\tilde{\mathbf{v}}_{i+1}}{\partial\tilde{\mathbf{x}}_*} &= -\Delta\widehat{\mathbf{R}}_i\lfloor\widehat{\mathbf{v}}_{i,i+1}\rfloor\frac{\partial\delta\Delta\boldsymbol{\theta}_i}{\partial\tilde{\mathbf{x}}_*} + \frac{\partial\Delta\tilde{\mathbf{v}}_i}{\partial\tilde{\mathbf{x}}_*} + \Delta\widehat{\mathbf{R}}_i[-\boldsymbol{\Xi}_3 \;\; \boldsymbol{\Xi}_1]\mathbf{H}_*^{wa}
\end{aligned}
$$

## D. State transition matrix

The detailed derivations for $\boldsymbol{\Phi}_B$, $\boldsymbol{\Phi}_A$, $\boldsymbol{\Phi}_G$, $\boldsymbol{\Phi}_{calib}$ and $\boldsymbol{\Phi}_F$ can be found as:

$$
\boldsymbol{\Phi}_B = \begin{bmatrix} \boldsymbol{\Phi}_I & \boldsymbol{\Phi}_{in} \\ \mathbf{0} & \mathbf{I} \end{bmatrix}, \boldsymbol{\Phi}_A = \begin{bmatrix} \boldsymbol{\Phi}_{I_a} & \boldsymbol{\Phi}_{A_{in}} \\ \mathbf{0} & \mathbf{I} \end{bmatrix}
$$

$$
\boldsymbol{\Phi}_G = \begin{bmatrix} \boldsymbol{\Phi}_{I_g} & \boldsymbol{\Phi}_{G_{in}} \\ \mathbf{0} & \mathbf{I} \end{bmatrix}, \boldsymbol{\Phi}_{calib} = \mathbf{I}, \boldsymbol{\Phi}_F = \mathbf{I}
$$

The state transition of $\boldsymbol{\Phi}_B$ and $\boldsymbol{\Phi}_A$ have the same structure. We can grab the gyroscope part of $\boldsymbol{\Phi}_B$ to get $\boldsymbol{\Phi}_G$. Therefore, only $\boldsymbol{\Phi}_B$ is shown in this paper for clarity. The $\boldsymbol{\Phi}_I$ is:

$$
\boldsymbol{\Phi}_I = \begin{bmatrix} \boldsymbol{\Phi}_{11} & \mathbf{0}_3 & \mathbf{0}_3 & \boldsymbol{\Phi}_{14} & \boldsymbol{\Phi}_{15} \\ \boldsymbol{\Phi}_{21} & \mathbf{I}_3 & \mathbf{I}_3\delta t & \boldsymbol{\Phi}_{24} & \boldsymbol{\Phi}_{25} \\ \boldsymbol{\Phi}_{31} & \mathbf{0}_3 & \mathbf{I}_3 & \boldsymbol{\Phi}_{34} & \boldsymbol{\Phi}_{35} \\ \mathbf{0}_3 & \mathbf{0}_3 & \mathbf{0}_3 & \mathbf{I}_3 & \mathbf{0}_3 \\ \mathbf{0}_3 & \mathbf{0}_3 & \mathbf{0}_3 & \mathbf{0}_3 & \mathbf{I}_3 \end{bmatrix} \tag{183}
$$

where we have:

$$\boldsymbol{\Phi}_{11} = {}^{I_{k+1}}_{I_k}\hat{\mathbf{R}}$$

$$\boldsymbol{\Phi}_{21} = -\left\lfloor {}^G\hat{\mathbf{p}}_{I_{k+1}} - {}^G\hat{\mathbf{p}}_{I_k} - {}^G\hat{\mathbf{v}}_{I_k}\delta t_k - \frac{1}{2}\,{}^G\mathbf{g}\delta t_k^2 \right\rfloor {}^G_{I_k}\hat{\mathbf{R}}$$

$$\boldsymbol{\Phi}_{31} = -\left\lfloor {}^G\hat{\mathbf{v}}_{I_{k+1}} - {}^G\hat{\mathbf{v}}_{I_k} - {}^G\mathbf{g}\delta t_k \right\rfloor {}^G_{I_k}\hat{\mathbf{R}}$$

$$\boldsymbol{\Phi}_{14} = -\mathbf{J}_r\delta t_k\, {}^I_w\hat{\mathbf{R}}\hat{\mathbf{D}}_w$$

$$\boldsymbol{\Phi}_{24} = {}^G_{I_k}\hat{\mathbf{R}}\boldsymbol{\Xi}_4\, {}^I_w\hat{\mathbf{R}}\hat{\mathbf{D}}_w$$

$$\boldsymbol{\Phi}_{34} = {}^G_{I_k}\hat{\mathbf{R}}\boldsymbol{\Xi}_3\, {}^I_w\hat{\mathbf{R}}\hat{\mathbf{D}}_w$$

$$\boldsymbol{\Phi}_{15} = \mathbf{J}_r\delta t_k\, {}^I_w\hat{\mathbf{R}}\hat{\mathbf{D}}_w\hat{\mathbf{T}}_g\, {}^I_a\hat{\mathbf{R}}\hat{\mathbf{D}}_a$$

$$\boldsymbol{\Phi}_{25} = -{}^G_{I_k}\hat{\mathbf{R}}\left(\boldsymbol{\Xi}_4\, {}^I_w\hat{\mathbf{R}}\hat{\mathbf{D}}_w\hat{\mathbf{T}}_g + \boldsymbol{\Xi}_2\right) {}^I_a\hat{\mathbf{R}}\hat{\mathbf{D}}_a$$

$$\boldsymbol{\Phi}_{35} = -{}^G_{I_k}\hat{\mathbf{R}}\left(\boldsymbol{\Xi}_3\, {}^I_w\hat{\mathbf{R}}\hat{\mathbf{D}}_w\hat{\mathbf{T}}_g + \boldsymbol{\Xi}_1\right) {}^I_a\hat{\mathbf{R}}\hat{\mathbf{D}}_a$$

Note that $\mathbf{J}_r \triangleq \mathbf{J}_r(\hat{\boldsymbol{\theta}}_{k,k+1})$. The $\boldsymbol{\Phi}_{in}$ is:

$$\boldsymbol{\Phi}_{in} = \begin{bmatrix} \boldsymbol{\Phi}_{in11} & \boldsymbol{\Phi}_{in12} & \boldsymbol{\Phi}_{in13} & \boldsymbol{\Phi}_{in14} \\ \boldsymbol{\Phi}_{in21} & \boldsymbol{\Phi}_{in22} & \boldsymbol{\Phi}_{in23} & \boldsymbol{\Phi}_{in24} \\ \boldsymbol{\Phi}_{in31} & \boldsymbol{\Phi}_{in23} & \boldsymbol{\Phi}_{in33} & \boldsymbol{\Phi}_{in34} \\ \mathbf{0}_3 & \mathbf{0}_3 & \mathbf{0}_3 & \mathbf{0}_3 \\ \mathbf{0}_3 & \mathbf{0}_3 & \mathbf{0}_3 & \mathbf{0}_3 \end{bmatrix} \tag{184}$$

where we have:

$$\boldsymbol{\Phi}_{in11} = \mathbf{J}_r\delta t_k\, {}^I_w\hat{\mathbf{R}}\mathbf{H}_{D_w}$$

$$\boldsymbol{\Phi}_{in21} = -{}^G_{I_k}\hat{\mathbf{R}}\boldsymbol{\Xi}_4\, {}^I_w\hat{\mathbf{R}}\mathbf{H}_{D_w}$$

$$\boldsymbol{\Phi}_{in31} = -{}^G_{I_k}\hat{\mathbf{R}}\boldsymbol{\Xi}_3\, {}^I_w\hat{\mathbf{R}}\mathbf{H}_{D_w}$$

$$\boldsymbol{\Phi}_{in12} = -\mathbf{J}_r\delta t_k\, {}^I_w\hat{\mathbf{R}}\hat{\mathbf{D}}_w\hat{\mathbf{T}}_g\, {}^I_a\hat{\mathbf{R}}\mathbf{H}_{D_a}$$

$$\boldsymbol{\Phi}_{in22} = {}^G_{I_k}\hat{\mathbf{R}}\left(\boldsymbol{\Xi}_2 + \boldsymbol{\Xi}_4\, {}^I_w\hat{\mathbf{R}}\hat{\mathbf{D}}_w\hat{\mathbf{T}}_g\right) {}^I_a\hat{\mathbf{R}}\mathbf{H}_{D_a}$$

$$\boldsymbol{\Phi}_{in32} = {}^G_{I_k}\hat{\mathbf{R}}\left(\boldsymbol{\Xi}_1 + \boldsymbol{\Xi}_3\, {}^I_w\hat{\mathbf{R}}\hat{\mathbf{D}}_w\hat{\mathbf{T}}_g\right) {}^I_a\hat{\mathbf{R}}\mathbf{H}_{D_a}$$

$$\boldsymbol{\Phi}_{in13} = -\mathbf{J}_r\delta t_k\, {}^I_w\hat{\mathbf{R}}\hat{\mathbf{D}}_w\mathbf{H}_{T_g}$$

$$\boldsymbol{\Phi}_{in23} = {}^G_{I_k}\hat{\mathbf{R}}\boldsymbol{\Xi}_4\, {}^I_w\hat{\mathbf{R}}\hat{\mathbf{D}}_w\mathbf{H}_{T_g}$$

$$\boldsymbol{\Phi}_{in33} = {}^G_{I_k}\hat{\mathbf{R}}\boldsymbol{\Xi}_3\, {}^I_w\hat{\mathbf{R}}\hat{\mathbf{D}}_w\mathbf{H}_{T_g}$$

$$\boldsymbol{\Phi}_{in14} = \mathbf{J}_r\delta t_k\, {}^I_w\hat{\mathbf{R}}\hat{\mathbf{D}}_w\hat{\mathbf{T}}_g\left\lfloor {}^I\hat{\mathbf{a}}\right\rfloor {}^I_a\hat{\mathbf{R}}$$

$$\boldsymbol{\Phi}_{in24} = -{}^G_{I_k}\hat{\mathbf{R}}\left(\boldsymbol{\Xi}_2 + \boldsymbol{\Xi}_4\, {}^I_w\hat{\mathbf{R}}\hat{\mathbf{D}}_w\hat{\mathbf{T}}_g\right)\left\lfloor {}^I\hat{\mathbf{a}}\right\rfloor {}^I_a\hat{\mathbf{R}}$$

$$\boldsymbol{\Phi}_{in34} = -{}^G_{I_k}\hat{\mathbf{R}}\left(\boldsymbol{\Xi}_1 + \boldsymbol{\Xi}_3\, {}^I_w\hat{\mathbf{R}}\hat{\mathbf{D}}_w\hat{\mathbf{T}}_g\right)\left\lfloor {}^I\hat{\mathbf{a}}\right\rfloor {}^I_a\hat{\mathbf{R}}$$

## *E. Measurement Jacobians*

Jacobians of camera measurements are computed as:

$$\mathbf{H}_{CB} = \mathbf{H}_{\mathbf{p}_f}\, {}^I_C\hat{\mathbf{R}}^\top\, {}^G_I\hat{\mathbf{R}}^\top \left[\, \left\lfloor {}^G\mathbf{p}_f - {}^G\mathbf{p}_I \right\rfloor {}^G_I\hat{\mathbf{R}} \quad -\mathbf{I}_3 \quad \mathbf{0}_{3\times 33} \right]$$

$$\mathbf{H}_{CC} = \left[\mathbf{H}_{CC_{Ex}} \quad \mathbf{H}_{CC_{in}}\right]$$

$$\mathbf{H}_{CF} = \mathbf{H}_{\mathbf{p}_f}\, {}^I_C\hat{\mathbf{R}}^\top\, {}^G_I\hat{\mathbf{R}}$$

$$\mathbf{H}_{CC_{Ex}} = \mathbf{H}_{\mathbf{p}_f}\, {}^I_C\hat{\mathbf{R}}^\top\, {}^G_I\hat{\mathbf{R}}^\top \times$$
$$\left[\mathbf{H}_{CC1} \quad \mathbf{H}_{CC2} \quad \mathbf{H}_{CC3} \quad \mathbf{H}_{CC4} \quad \mathbf{H}_{CC5} \quad \mathbf{H}_{CC6}\right]$$

$$\mathbf{H}_{CC1} = \mathbf{0}_{3\times 7}$$

$$\mathbf{H}_{CC2} = \mathbf{0}_{3\times 4}$$

$$\mathbf{H}_{CC3} = \left\lfloor {}^G\hat{\mathbf{p}}_f - {}^G\hat{\mathbf{p}}_I - {}^G_I\hat{\mathbf{R}}\, {}^I\hat{\mathbf{p}}_C \right\rfloor$$

$$\mathbf{H}_{CC4} = -{}^G_I\hat{\mathbf{R}}$$

$$\mathbf{H}_{CC5} = {}^G\hat{\mathbf{v}}_I - \left\lfloor {}^G\hat{\mathbf{p}}_f - {}^G\hat{\mathbf{p}}_I \right\rfloor {}^G_I\hat{\mathbf{R}}\, {}^I\hat{\boldsymbol{\omega}}$$

$$\mathbf{H}_{CC6} = -\frac{m}{M}\mathbf{H}_{CC5}$$

$$\mathbf{H}_{CC_{in}} = \frac{\partial \tilde{\mathbf{z}}_C}{\partial \tilde{\mathbf{x}}_{C_{in}}}$$

Note that $t_I = t_C - t_d = t_C - \hat{t}_d - \tilde{t}_d$. The measurement Jacobians for auxiliary IMU constraints are computed as:

$$\mathbf{H}_{AB} = \begin{bmatrix} {}^{I_a}_I\hat{\mathbf{R}} & \mathbf{0}_3 & \mathbf{0}_3 & \mathbf{0}_3 & \mathbf{0}_3 & \mathbf{0}_{3\times 24} \\ {}^G_I\hat{\mathbf{R}}\left\lfloor {}^I\hat{\mathbf{p}}_{I_a}\right\rfloor & -\mathbf{I}_3 & \mathbf{0}_3 & \mathbf{0}_3 & \mathbf{0}_3 & \mathbf{0}_{3\times 24} \end{bmatrix}$$

$$\mathbf{H}_{AA} = \begin{bmatrix} -\mathbf{I}_3 & \mathbf{0}_3 & \mathbf{0}_3 & \mathbf{0}_3 & \mathbf{0}_3 & \mathbf{0}_{3\times 24} \\ \mathbf{0}_3 & \mathbf{I}_3 & \mathbf{0}_3 & \mathbf{0}_3 & \mathbf{0}_3 & \mathbf{0}_{3\times 24} \end{bmatrix}$$

$$\mathbf{H}_{AC} = \begin{bmatrix} \mathbf{I}_3 & \mathbf{0}_3 & -{}^{I_a}\hat{\boldsymbol{\omega}} & \mathbf{0}_{3\times 20} \\ \mathbf{0}_3 & -{}^G_I\hat{\mathbf{R}} & {}^G\mathbf{v}_I + {}^G_I\hat{\mathbf{R}}\left\lfloor {}^I\hat{\boldsymbol{\omega}}\right\rfloor {}^I\mathbf{p}_{I_a} & \mathbf{0}_{3\times 20} \end{bmatrix}$$

Note that $t_I = t_a - t_{d_a} = t_a - \hat{t}_{d_a} - \tilde{t}_{d_a}$ is used when computing the Jacobians for the $t_{d_a}$. The measurement Jacobians for auxiliary gyroscope constraints are computed as:

$$\mathbf{H}_{GB} = \left[\, {}^{I_g}_I\hat{\mathbf{R}} \quad \mathbf{0}_3 \quad \mathbf{0}_3 \quad \mathbf{0}_3 \quad \mathbf{0}_3 \quad \mathbf{0}_{3\times 24} \right]$$
$$\mathbf{H}_{GG} = \left[-\mathbf{I}_3 \quad \mathbf{0}_3 \quad \mathbf{0}_3 \quad \mathbf{0}_3\right]$$
$$\mathbf{H}_{GC} = \left[\mathbf{0}_{3\times 7} \quad \mathbf{I}_3 \quad -{}^{I_g}\hat{\boldsymbol{\omega}} \quad \mathbf{0}_{3\times 8} \quad \mathbf{0}_{3\times 8}\right]$$

Note that $t_I = t_g - t_{d_g} = t_g - \hat{t}_{d_g} - \tilde{t}_{d_g}$ is used when computing the Jacobians for the $t_{d_g}$.

## F. Observability matrix

For the component $\mathbf{M}_{CB}$, we have:

$$\mathbf{M}_{CB} = \mathbf{H}_{CB}\mathbf{\Phi}_B$$
$$= \mathbf{H}_{\mathbf{p}_f C}{}^{I}_{C}\widehat{\mathbf{R}}^{\top}{}^{G}_{I}\widehat{\mathbf{R}}^{\top} \times [\mathbf{\Gamma}_1\ \mathbf{\Gamma}_2\ \mathbf{\Gamma}_3\ \mathbf{\Gamma}_4\ \mathbf{\Gamma}_5\ \mathbf{\Gamma}_6\ \mathbf{\Gamma}_7\ \mathbf{\Gamma}_8\ \mathbf{\Gamma}_9]$$

where:

$$\mathbf{\Gamma}_1 = \lfloor {}^{G}\widehat{\mathbf{p}}_f - {}^{G}\widehat{\mathbf{p}}_{I_1} - {}^{G}\widehat{\mathbf{v}}_{I_1}\delta t - \frac{1}{2}{}^{G}\mathbf{g}\delta t^2 \rfloor {}^{G}_{I_1}\widehat{\mathbf{R}}$$
$$\mathbf{\Gamma}_2 = -\mathbf{I}_3$$
$$\mathbf{\Gamma}_3 = -\mathbf{I}_3\delta t$$
$$\mathbf{\Gamma}_4 = \lfloor {}^{G}\widehat{\mathbf{p}}_f - {}^{G}\widehat{\mathbf{p}}_{I_k} \rfloor {}^{G}_{I_k}\widehat{\mathbf{R}}\mathbf{\Phi}_{14} - \mathbf{\Phi}_{24}$$
$$\mathbf{\Gamma}_5 = \lfloor {}^{G}\widehat{\mathbf{p}}_f - {}^{G}\widehat{\mathbf{p}}_{I_k} \rfloor {}^{G}_{I_k}\widehat{\mathbf{R}}\mathbf{\Phi}_{15} - \mathbf{\Phi}_{25}$$
$$\mathbf{\Gamma}_6 = \lfloor {}^{G}\widehat{\mathbf{p}}_f - {}^{G}\widehat{\mathbf{p}}_{I_k} \rfloor {}^{G}_{I_k}\widehat{\mathbf{R}}\mathbf{\Phi}_{in11} - \mathbf{\Phi}_{in21}$$
$$\mathbf{\Gamma}_7 = \lfloor {}^{G}\widehat{\mathbf{p}}_f - {}^{G}\widehat{\mathbf{p}}_{I_k} \rfloor {}^{G}_{I_k}\widehat{\mathbf{R}}\mathbf{\Phi}_{in12} - \mathbf{\Phi}_{in22}$$
$$\mathbf{\Gamma}_8 = \lfloor {}^{G}\widehat{\mathbf{p}}_f - {}^{G}\widehat{\mathbf{p}}_{I_k} \rfloor {}^{G}_{I_k}\widehat{\mathbf{R}}\mathbf{\Phi}_{in13} - \mathbf{\Phi}_{in23}$$
$$\mathbf{\Gamma}_9 = \lfloor {}^{G}\widehat{\mathbf{p}}_f - {}^{G}\widehat{\mathbf{p}}_{I_k} \rfloor {}^{G}_{I_k}\widehat{\mathbf{R}}\mathbf{\Phi}_{in14} - \mathbf{\Phi}_{in24}$$

For the component $\mathbf{M}_{AB}$, we have:

$$\mathbf{M}_{AB} = \mathbf{H}_{AB}\mathbf{\Phi}_B$$
$$= \begin{bmatrix} {}^{I_a}_{I}\widehat{\mathbf{R}} & \mathbf{0}_3 \\ \mathbf{0}_3 & \mathbf{I}_3 \end{bmatrix} \times$$
$$\begin{bmatrix} \mathbf{\Phi}_{11} & \mathbf{0}_3 & \mathbf{0}_3 & \mathbf{\Phi}_{14} & \mathbf{\Phi}_{15} & \mathbf{\Phi}_{in11} & \mathbf{\Phi}_{in12} & \mathbf{\Phi}_{in13} & \mathbf{\Phi}_{in14} \\ \mathbf{\Gamma}_{a1} & \mathbf{\Gamma}_{a2} & \mathbf{\Gamma}_{a3} & \mathbf{\Gamma}_{a4} & \mathbf{\Gamma}_{a5} & \mathbf{\Gamma}_{a6} & \mathbf{\Gamma}_{a7} & \mathbf{\Gamma}_{a8} & \mathbf{\Gamma}_{a9} \end{bmatrix}$$

where:

$$\mathbf{\Gamma}_{a1} = \left\lfloor {}^{G}\widehat{\mathbf{p}}_{I_{a_k}} - {}^{G}\widehat{\mathbf{p}}_{I_1} - {}^{G}\widehat{\mathbf{v}}_{I_1}\delta t - \frac{1}{2}{}^{G}\mathbf{g}\delta t^2 \right\rfloor {}^{G}_{I_1}\widehat{\mathbf{R}}$$
$$\mathbf{\Gamma}_{a2} = -\mathbf{I}_3$$
$$\mathbf{\Gamma}_{a3} = -\mathbf{I}_3\delta t$$
$$\mathbf{\Gamma}_{a4} = {}^{G}_{I_k}\widehat{\mathbf{R}} \left\lfloor {}^{I}\widehat{\mathbf{p}}_{I_a} \right\rfloor \mathbf{\Phi}_{14} - \mathbf{\Phi}_{24}$$
$$\mathbf{\Gamma}_{a5} = {}^{G}_{I_k}\widehat{\mathbf{R}} \left\lfloor {}^{I}\widehat{\mathbf{p}}_{I_a} \right\rfloor \mathbf{\Phi}_{15} - \mathbf{\Phi}_{25}$$
$$\mathbf{\Gamma}_{a6} = {}^{G}_{I_k}\widehat{\mathbf{R}} \left\lfloor {}^{I}\widehat{\mathbf{p}}_{I_a} \right\rfloor \mathbf{\Phi}_{in11} - \mathbf{\Phi}_{in21}$$
$$\mathbf{\Gamma}_{a7} = {}^{G}_{I_k}\widehat{\mathbf{R}} \left\lfloor {}^{I}\widehat{\mathbf{p}}_{I_a} \right\rfloor \mathbf{\Phi}_{in12} - \mathbf{\Phi}_{in22}$$
$$\mathbf{\Gamma}_{a8} = {}^{G}_{I_k}\widehat{\mathbf{R}} \left\lfloor {}^{I}\widehat{\mathbf{p}}_{I_a} \right\rfloor \mathbf{\Phi}_{in13} - \mathbf{\Phi}_{in23}$$
$$\mathbf{\Gamma}_{a9} = {}^{G}_{I_k}\widehat{\mathbf{R}} \left\lfloor {}^{I}\widehat{\mathbf{p}}_{I_a} \right\rfloor \mathbf{\Phi}_{in14} - \mathbf{\Phi}_{in24}$$

For the component $\mathbf{M}_{AA}$, we have:

$$\mathbf{M}_{AA} = \mathbf{H}_{AA}\mathbf{\Phi}_A$$
$$= \begin{bmatrix} -\mathbf{I}_3 & \mathbf{0}_3 & \mathbf{0}_3 & \mathbf{0}_3 & \mathbf{0}_3 & \mathbf{0}_{3\times 24} \\ \mathbf{0}_3 & \mathbf{I}_3 & \mathbf{0}_3 & \mathbf{0}_3 & \mathbf{0}_3 & \mathbf{0}_{3\times 24} \end{bmatrix} \begin{bmatrix} \mathbf{\Phi}_{I_a} & \mathbf{\Phi}_{A_{in}} \\ \mathbf{0} & \mathbf{I} \end{bmatrix}$$

For the component $\mathbf{M}_{GB}$, we have:

$$\mathbf{M}_{GB} = \mathbf{H}_{GB}\mathbf{\Phi}_B$$
$$= \begin{bmatrix} {}^{I_g}_{I}\widehat{\mathbf{R}} & \mathbf{0}_3 & \mathbf{0}_3 & \mathbf{0}_3 & \mathbf{0}_3 & \mathbf{0}_{3\times 24} \end{bmatrix} \begin{bmatrix} \mathbf{\Phi}_I & \mathbf{\Phi}_{in} \\ \mathbf{0} & \mathbf{I} \end{bmatrix}$$

For the component $\mathbf{M}_{GG}$, we have:

$$\mathbf{M}_{GG} = \mathbf{H}_{GG}\mathbf{\Phi}_G$$
$$= \begin{bmatrix} -\mathbf{I}_3 & \mathbf{0}_3 & \mathbf{0}_3 & \mathbf{0}_3 \end{bmatrix} \begin{bmatrix} \mathbf{\Phi}_{I_g} & \mathbf{\Phi}_{G_{in}} \\ \mathbf{0} & \mathbf{I} \end{bmatrix}$$

Due to $\mathbf{\Phi}_{calib} = \mathbf{I}$ and $\mathbf{\Phi}_F = \mathbf{I}$, we have:

$$\mathbf{M}_{CC} = \mathbf{H}_{CC}$$
$$\mathbf{M}_{AC} = \mathbf{H}_{AC}$$
$$\mathbf{M}_{GC} = \mathbf{H}_{GC}$$
$$\mathbf{M}_{CF} = \mathbf{H}_{CF}$$